\documentclass[journal,comsoc]{IEEEtran}

\usepackage[T1]{fontenc}
\usepackage{amsmath}
\usepackage{pifont}
\usepackage[cmintegrals]{newtxmath}
\usepackage{bm}
\usepackage{tikz}
\usepackage{array}
\usepackage{multirow}
\usepackage{pdflscape}
\usepackage{xurl}
\usepackage{subcaption}
\usepackage{makecell}
\usepackage[flushleft]{threeparttable}
\usetikzlibrary{arrows,shapes,positioning,shadows,trees}

\usepackage{booktabs}
\newcolumntype{N}{>{\centering\arraybackslash}m{.5in}}
\newcolumntype{G}{>{\centering\arraybackslash}m{2in}}

\begin{document}

\title{Green Edge AI: A Contemporary Survey}

\author{Yuyi~Mao,~\IEEEmembership{Senior Member,~IEEE},
        Xianghao~Yu,~\IEEEmembership{Senior Member,~IEEE},
        Kaibin~Huang,~\IEEEmembership{Fellow,~IEEE},\\
        Ying-Jun~Angela~Zhang,~\IEEEmembership{Fellow,~IEEE},
        and~Jun~Zhang,~\IEEEmembership{Fellow,~IEEE}
\thanks{
Y. Mao is with the Department of Electrical and Electronic Engineering, The Hong Kong Polytechnic University, Hong Kong (e-mail: yuyi-eie.mao@polyu.edu.hk). 

X. Yu is with the Department of Electrical Engineering,  City University of Hong Kong, Hong Kong (e-mail: alex.yu@cityu.edu.hk).

K. Huang is with the Department of Electrical and Electronic Engineering, The University of Hong Kong, Hong Kong (e-mail: huangkb@eee.hku.hk).

Y.-J. A. Zhang is with the Department of Information Engineering, The Chinese University of Hong Kong, Hong Kong (e-mail: yjzhang@ie.cuhk.edu.hk).

J. Zhang is with the Department of Electronic and Computer Engineering, The Hong Kong University of Science and Technology, Hong Kong (e-mail: eejzhang@ust.hk). \emph{(Corresponding author: Jun Zhang.)}
}
}

\markboth{}
{Shell \MakeLowercase{\textit{et al.}}: Bare Demo of IEEEtran.cls for IEEE Communications Society Journals}

\maketitle

\begin{abstract}
Artificial intelligence (AI) technologies have emerged as pivotal enablers across a multitude of industries, including consumer electronics, healthcare, and manufacturing, largely due to their significant resurgence over the past decade. The transformative power of AI is primarily derived from the utilization of deep neural networks (DNNs), which require extensive data for training and substantial computational resources for processing. Consequently, DNN models are typically trained and deployed on resource-rich cloud servers. However, due to potential latency issues associated with cloud communications, deep learning (DL) workflows (e.g., DNN training and inference) are increasingly being transitioned to wireless edge networks in proximity to end-user devices (EUDs). This shift is designed to support latency-sensitive applications and has given rise to a new paradigm of edge AI, which will play a critical role in upcoming sixth-generation (6G) networks to support ubiquitous AI applications. Despite its considerable potential, edge AI faces substantial challenges, mostly due to the dichotomy between the resource limitations of wireless edge networks and the resource-intensive nature of DL. Specifically, the acquisition of large-scale data, as well as the training and inference processes of DNNs, can rapidly deplete the battery energy of EUDs. This necessitates an energy-conscious approach to edge AI to ensure both optimal and sustainable performance. In this paper, we present a contemporary survey on green edge AI. We commence by analyzing the principal energy consumption components of edge AI systems to identify the fundamental design principles of green edge AI. Guided by these principles, we then explore energy-efficient design methodologies for the three critical tasks in edge AI systems, including training data acquisition, edge training, and edge inference. Finally, we underscore potential future research directions to further enhance the energy efficiency of edge AI.
\end{abstract}

\begin{IEEEkeywords}
Sixth-generation (6G) wireless networks, mobile edge computing, edge artificial intelligence (AI), green AI, data acquisition, federated learning, edge inference, energy efficiency. 
\end{IEEEkeywords}

\IEEEpeerreviewmaketitle

\section{Introduction}

\subsection{Background}

Six decades after its inception at the 1956 Dartmouth Conference \cite{MoorJ_AIMAG06}, artificial intelligence (AI) demonstrated its potential to surpass human intelligence in 2016. This was evidenced when a computer program named AlphaGo \cite{SilverD_Nature16} triumphed over a top-ranked human Go player for the first time, which not only signified a landmark moment in AI history but also catalyzed a resurgence in AI interest and development.
The AI renaissance has also been fueled by advancements in computing hardware, the proliferation of big data, and the advent of deep neural networks (DNNs). As a result, AI technologies have made remarkable breakthroughs across a multitude of disciplines, including but not limited to, computer vision \cite{GuoY_NeuCom16}, natural language processing (NLP) \cite{Otter_TNNLS21}, healthcare \cite{Esteva_NatureMed19}, manufacturing \cite{Khalil_IoTJ21}, and financial technology (FinTech) \cite{Ozbayoglu_ASC20}. Consequently, AI is rapidly emerging as a restless engine of future productivity and has been recognized as a national strategic priority by major global economies, including China, the United States, and the European Union.

The success of AI is underpinned by the power of advanced data analytics, which typically entails substantial computation overhead \cite{SzeV_PIEEE17}. Traditionally, AI computation workloads (e.g., DNN training and inference) initiated by end-user devices (EUDs) are offloaded to the Cloud, which boasts virtually unlimited computational resources \cite{BIANCHINI_ACMCOMM20}. However, the prevalent cloud AI paradigm falls short in supporting latency-critical applications. This is due to the multi-hop routing and network congestion between EUDs and remote public cloud servers, which can result in round-trip time (RTT) of tens to hundreds of milliseconds \cite{MaoY_COMST17}. Furthermore, a cloud server can become inundated by a high volume of concurrent AI service requests, leading to scalability issues \cite{TongW_Book21}. Concentrating all computational resources in a few cloud data centers also exposes the system to single-point failures and security threats. Moreover, data collected from EUDs may be at risk of privacy breaches \cite{Tari_MCC15}. These significant limitations of cloud AI have necessitated the exploration of alternative solutions for delivering AI services.

Gratefully, the drawbacks of cloud AI can be complemented by the emerging mobile edge computing (MEC) technologies, which integrate cloud-like functionalities in the radio access network (RAN) near EUDs \cite{ETSI_MEC_15}. The synergy between MEC and AI enables a major paradigm shift, from cloud AI toward edge AI \cite{ZhouZ_PIEEE20,KBL_JSAC22}, where the AI computations of EUDs can be offloaded to a wirelessly connected MEC server. As such, the data communication latency can be trimmed significantly to the millisecond level \cite{ChenJ_PIEEE19}, which is essential for real-time interactive AI applications such as mixed reality, smart robots, and autonomous vehicles. Edge AI was identified by Huawei, a leading provider of telecom infrastructures and devices, as one of the six pillars in the sixth-generation (6G) wireless networks \cite{TongW_Book21}. The International Telecommunication Union Radiocommunication Sector (ITU-R) has also defined \emph{integrated AI and communication} as a key usage scenario of IMT-2030 (6G) in June 2023 \cite{ITU_R_2306}. Since 6G is expected to rollout by 2030, edge AI technologies are foreseen to receive unprecedented attention in the near future as predicted by the Gartner Hype Cycle \cite{Gartner_2022}. Apart from latency reduction, edge AI also enjoys many other benefits over cloud AI \cite{EdgeAI_WhitePaper_20,DingY_ACMCOMRev22}. On the one hand, the relatively low investment of MEC infrastructures allows their deployments at scale so that the massive AI service demands in 6G can be accommodated efficiently. On the other hand, edge AI inherits the privacy and security enhancements in MEC, since the computing infrastructures are highly distributed and can be privately owned. For instance, without relying on the cloud service providers, user data can be kept and processed inside the local domains, such as university and enterprise networks, reducing the chance of information leakage. Nevertheless, despite all these merits, there exists a significant energy challenge of edge AI, which is caused by the rapidly growing power consumption of AI workloads and the limited energy resources over the wireless edge network \cite{Desislavov_SusComput23}.

\subsection{The Energy Challenge in Edge AI}

The driving force behind AI is a variety of data analytics algorithms, with deep learning (DL) being the most prevalent largely due to the superior representation capability of DNNs. This capability was first highlighted by the Universal Approximation Theorem in the late 1980s \cite{HornikK_NN89}.
In the pursuit of state-of-the-art performance, DNN models have exhibited exponential increases in both model size (typically measured by the number of DNN parameters) and computation workload (usually quantified by the number of floating-point operations (FLOPs) or multiply-accumulate (MAC) operations). For instance, consider the ImageNet classification task \cite{DengJ_CVPR09}: The top-1$\%$ classification accuracy has been improved from $63.3\%$ (achieved by AlexNet \cite{Alex_NeurIPS12}) to $91.0\%$ (achieved by CoCa \cite{YuJ_CoCa22}) between 2012 and 2022. Concurrently, the model size increased by over 30 times, from 60 million to 2.1 billion network parameters. A 2019 analysis by a research group from Stanford University revealed that the computation demand of AI has been doubling every 3.4 months since 2012, far outstripping Moore's Law \cite{Stanford_AI19}. A significant consequence of such substantial computation overhead is the enormous energy footprint. For example, the energy consumed to train a transformer, a top-tier DNN model for NLP, can exceed the energy consumption of a car over its entire lifespan \cite{Strubell_ACL19}. The energy consumption of using trained DNNs for inference (prediction) is even more substantial. A case study in \cite{WangH_GLOBECOM19} demonstrated that for a DNN-based object tracking application running on smartphones, the inference process accounts for over $20\%$ of the total energy consumption. It was also estimated that each ChatGPT query consumes the same amount of energy as powering a 5 Watt light-emitting diode (LED) bulb for 1.3 hours \cite{ChatGPTenergy_2023}. The escalating carbon footprint of DL applications has also placed significant environmental stress \cite{Ligozat_Sustainability22}. Consequently, researchers have recently begun to emphasize the importance of AI \emph{efficiency}, in addition to the long-standing golden standard of \emph{effectiveness}. This shift in focus has spawned the field of green AI \cite{Schwartz_ACMCOM19,Verdecchia_WILEY23}, which is more concerned about the computational cost and energy efficiency (EE) of AI developments.{\footnote{Although ``greenness'' can have a broader context (e.g., reducing waste and pollution generation~\cite{UN_Green-economy}), we narrow down our discussion to ``high EE'', which is consistent with the definitions in extensive literature~\cite{ChenY_MCOM11,Sharma_Internet17,Albreem_Access21}.}}

As complex DL processing tasks move away from the Cloud \cite{MastelicT_CC15}, edge AI is emerging as a more sustainable solution for intelligent mobile applications. To a great extent, this is due to the use of low-power edge computing devices, which saves the high energy costs associated with operating cloud data centers. However, the energy challenge remains a significant issue in edge AI, and in fact, becomes even more pronounced \cite{EdgeAI_WhitePaper_20}. Firstly, the number of EUDs requiring AI services, such as smartphones, personal computers, and vehicles, is increasing at an astonishing rate. A market analysis report by Tractica predicted that the annual worldwide shipment of AI-enabled EUDs will reach 2.6 billion units by 2025, representing a 16-fold increase compared to 2018 \cite{Tractia_18}. The widespread adoption of intelligent mobile applications will undoubtedly exacerbate the energy crisis of edge AI. Secondly, EUDs are primarily powered by batteries, the development of which significantly lags behind AI algorithmic innovations. Consequently, if compute-intensive DL algorithms are executed on EUDs, they can quickly deplete the battery energy. This not only hinders the provision of stable AI services but may also disrupt other regular system operations. While such an issue can be mitigated with MEC infrastructures, it is challenging to provide sufficient resources to meet the heavy AI computation workloads offloaded from a large number of EUDs. Thirdly, the collaboration between EUDs and MEC servers involves frequent two-way wireless communications \cite{ShiY_COMST20}. As a result, the energy consumption of communication contributes significantly to the overall energy consumption of edge AI systems, in addition to the computation energy consumption in the mobile AI paradigm, where EUDs handle all the computation workloads.

\begin{figure*}[ht!]
	\includegraphics[width=\linewidth]{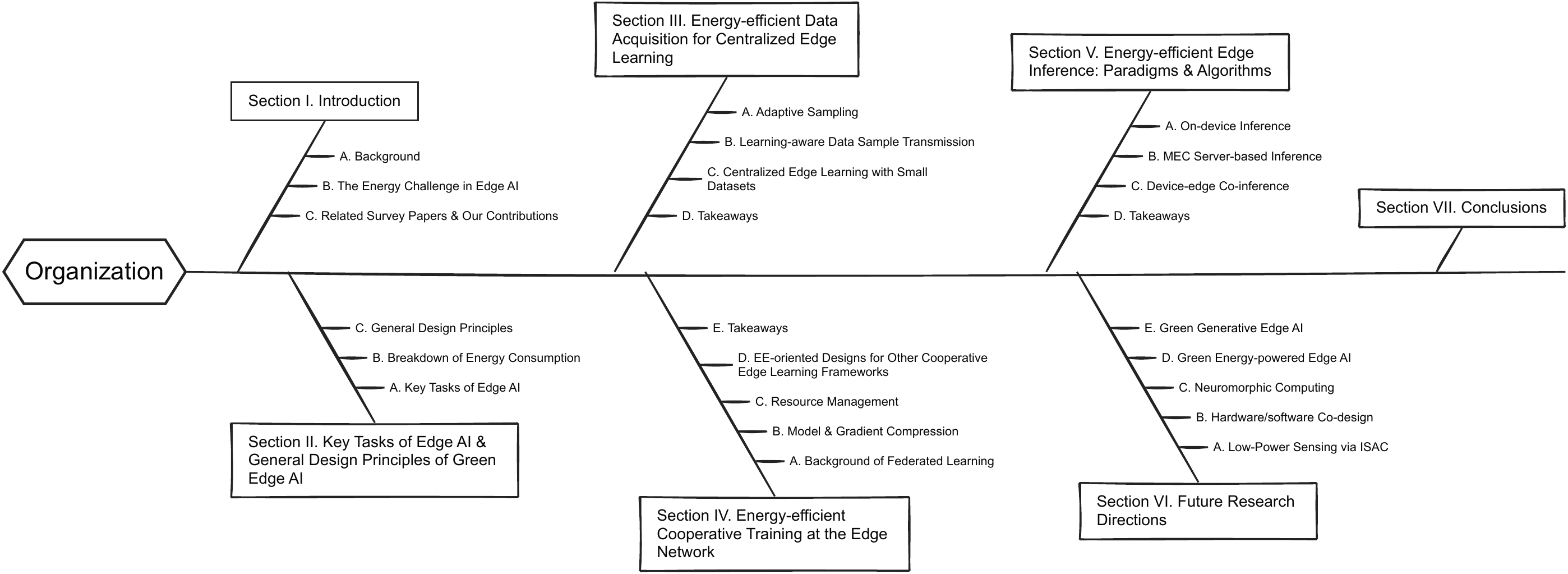}
	\caption{Paper organization.}
	\label{fig:organization}
\end{figure*}

To provide high-quality, low-latency, and sustainable AI services to EUDs, it is crucial to develop dedicated solutions to address the energy challenge in edge AI. This forms the primary motivation for this survey on \emph{green edge AI}. As the first survey of its kind, our focus will be on reducing energy consumption at EUDs, given that MEC infrastructures are likely to have stable power supplies. The improvements of EE shall naturally relieve the environmental impacts of edge AI due to the reduced demand of conventional non-renewable energy.

\begin{table*}[ht]
	\centering
	\caption{Summary of Recent Survey Papers on Edge AI}
	\begin{tabular}{m{1.5em}|m{3.5em}m{3.5em}m{3.5em}|m{4em}|m{4em}|m{32em}}
		\hline
		\multirow{3}{*}{Ref.}       & \multicolumn{3}{c|}{\scriptsize{AI Tasks}}                                                      & \multirow{3}{*}{\makecell[l]{\scriptsize{\makecell[l]{Energy \\Efficiency-\\oriented}}}}      & \multirow{3}{*}{\scriptsize{\makecell[l]{Device-\\edge\\Cooperation}}}      & \multirow{3}{*}{Overview}      \\ \cline{2-4}
		& \multicolumn{1}{m{3.5em}|}{\scriptsize{\makecell[l]{Dataset\\Acqui.}}}  & \multicolumn{1}{m{3.5em}|}{\scriptsize{\makecell[l]{Model\\ Training}}}  & \multicolumn{1}{m{3.5em}|}{\scriptsize{\makecell[l]{DNN\\ Inference}}}              &                         &                         &                         \\ \hline\hline
		\multicolumn{1}{l|}{\cite{SzeV_PIEEE17}}                       & \multicolumn{1}{c|}{{\color{red}$\bm{\times}$}} & \multicolumn{1}{c|}{{\color{red}$\bm{\times}$}} & \multicolumn{1}{c|}{{\color{teal}$\bm{\checkmark}$}}                      & \multicolumn{1}{c|}{{\color{teal}$\bm{\checkmark}$}}                      & \multicolumn{1}{c|}{{\color{red}$\bm{\times}$}}                      & \multicolumn{1}{l}{\makecell[l]{Hardware platforms and architectures that support DNN processing, and trends \\of hardware and hardware/software co-designs to reduce the computational cost}}                      \\ \hline
		\multicolumn{1}{c|}{\cite{ZhouZ_PIEEE20}} & \multicolumn{1}{c|}{{\color{red}$\bm{\times}$}} & \multicolumn{1}{c|}{{\color{teal}$\bm{\checkmark}$}} & \multicolumn{1}{c|}{{\color{teal}$\bm{\checkmark}$}} & \multicolumn{1}{c|}{{\color{red}$\bm{\times}$}} & \multicolumn{1}{c|}{{\color{teal}$\bm{\checkmark}$}} & \multicolumn{1}{l}{\makecell[l]{Architectures, enabling techniques, systems and frameworks for model training\\ and inference at the network edge}} \\ \hline
		\multicolumn{1}{c|}{\cite{KBL_JSAC22}}                       & \multicolumn{1}{c|}{{\color{red}$\bm{\times}$}} & \multicolumn{1}{c|}{{\color{teal}$\bm{\checkmark}$}} & \multicolumn{1}{c|}{{\color{teal}$\bm{\checkmark}$}}                     & \multicolumn{1}{c|}{{\color{red}$\bm{\times}$}}                   & \multicolumn{1}{c|}{{\color{teal}$\bm{\checkmark}$}}                      & \multicolumn{1}{l}{\makecell[l]{Wireless techniques, resource management approaches, network architectures\\  for edge AI. Standardizations, platforms, and applications are also discussed}}                      \\ \hline
		\multicolumn{1}{l|}{\cite{ChenJ_PIEEE19}}	  & \multicolumn{1}{c|}{{\color{red}$\bm{\times}$}} 				  & \multicolumn{1}{c|}{{\color{teal}$\bm{\checkmark}$}} 			  & \multicolumn{1}{c|}{{\color{teal}$\bm{\checkmark}$}}             & \multicolumn{1}{c|}{{\color{red}$\bm{\times}$}}                 & \multicolumn{1}{c|}{{\color{teal}$\bm{\checkmark}$}}             & \multicolumn{1}{l}{\makecell[l]{Applications of edge AI, fast DNN inference across the device-edge-cloud \\synergy, and model training methods on EUDs}}                      \\ \hline
		\multicolumn{1}{c|}{\cite{ShiY_COMST20}} & \multicolumn{1}{c|}{{\color{red}$\bm{\times}$}} & \multicolumn{1}{c|}{{\color{teal}$\bm{\checkmark}$}} & \multicolumn{1}{c|}{{\color{teal}$\bm{\checkmark}$}} & \multicolumn{1}{c|}{{\color{red}$\bm{\times}$}} & \multicolumn{1}{c|}{{\color{teal}$\bm{\checkmark}$}} & \multicolumn{1}{l}{\makecell[l]{Techniques to overcome the communication challenges in edge AI, from both\\algorithmic and system perspectives for the training and inference tasks}} \\ \hline
		\multicolumn{1}{c|}{\cite{XuW_JSTSP23}}                       & \multicolumn{1}{c|}{{\color{red}$\bm{\times}$}} & \multicolumn{1}{c|}{{\color{teal}$\bm{\checkmark}$}} & \multicolumn{1}{c|}{{\color{teal}$\bm{\checkmark}$}} & \multicolumn{1}{c|}{{\color{red}$\bm{\times}$}} & \multicolumn{1}{c|}{{\color{teal}$\bm{\checkmark}$}} & \multicolumn{1}{l}{\makecell[l]{{An overview on distributed edge learning techniques and their interplay with }\\{the designs of beyond 5G networks}}}       
		\\ \hline
		\multicolumn{1}{c|}{\cite{Baccour_COMST22}}                       & \multicolumn{1}{c|}{{\color{teal}$\bm{\checkmark}$}} & \multicolumn{1}{c|}{{\color{teal}$\bm{\checkmark}$}} & \multicolumn{1}{c|}{{\color{teal}$\bm{\checkmark}$}} & \multicolumn{1}{c|}{{\color{red}$\bm{\times}$}}                      & \multicolumn{1}{c|}{{\color{teal}$\bm{\checkmark}$}}                      &\multicolumn{1}{l}{\makecell[l]{Models and resource-efficient techniques for pervasive training and inference\\ across IoT devices, edge infrastructures, and the cloud}}                   \\ \hline
		\multicolumn{1}{c|}{\cite{XuDianlei_PIEEE21}}                     & \multicolumn{1}{c|}{{\color{teal}$\bm{\checkmark}$}} & \multicolumn{1}{c|}{{\color{teal}$\bm{\checkmark}$}} & \multicolumn{1}{c|}{{\color{teal}$\bm{\checkmark}$}}                      &\multicolumn{1}{c|}{{\color{red}$\bm{\times}$}}                     & \multicolumn{1}{c|}{{\color{teal}$\bm{\checkmark}$}}                      & \multicolumn{1}{l}{\makecell[l]{Literature review on edge caching, training, inference, and offloading, from the \\ perspectives of adopted techniques, objectives, and performance}}                      \\ \hline
		\multicolumn{1}{c|}{\cite{LimW_COMST20}}                        & \multicolumn{1}{c|}{{\color{red}$\bm{\times}$}} & \multicolumn{1}{c|}{{\color{teal}$\bm{\checkmark}$}} & \multicolumn{1}{c|}{{\color{red}$\bm{\times}$}}                      & \multicolumn{1}{c|}{{\color{red}$\bm{\times}$}}                      & \multicolumn{1}{c|}{{\color{teal}$\bm{\checkmark}$}}                      & \multicolumn{1}{l}{\makecell[l]{Introduction to the challenges of communication costs, resource allocation, \\ and privacy/security of federated learning, and review the solutions}}                      \\ \hline
		\multicolumn{1}{l|}{\cite{Mazumder_JESTCS21}}                        & \multicolumn{1}{c|}{{\color{red}$\bm{\times}$}} & \multicolumn{1}{c|}{{\color{red}$\bm{\times}$}} & \multicolumn{1}{c|}{{\color{teal}$\bm{\checkmark}$}}                     & \multicolumn{1}{c|}{{\color{teal}$\bm{\checkmark}$}}                       & \multicolumn{1}{c|}{{\color{red}$\bm{\times}$}}                       & \multicolumn{1}{l}{\makecell[l]{Recent developments of energy-efficient deployments on micro-AI platforms, \\considering both micro-AI model optimization and hardware optimization}}         \\ \hline
		\multicolumn{1}{c|}{\cite{LiuD_Neurocomputing22}}                       & \multicolumn{1}{c|}{{\color{red}$\bm{\times}$}} & \multicolumn{1}{c|}{{\color{red}$\bm{\times}$}} & \multicolumn{1}{c|}{{\color{teal}$\bm{\checkmark}$}}                      & \multicolumn{1}{c|}{{\color{red}$\bm{\times}$}}                      & \multicolumn{1}{c|}{{\color{red}$\bm{\times}$}}                      & \multicolumn{1}{l}{\makecell[l]{Survey on deep learning techniques for efficient deployment of DNN models \\on edge computing systems}}                      \\ \hline
		\multicolumn{1}{c|}{\cite{RenQ_MIR23}}                       & \multicolumn{1}{c|}{{\color{red}$\bm{\times}$}} & \multicolumn{1}{c|}{{\color{red}$\bm{\times}$}} & \multicolumn{1}{c|}{{\color{teal}$\bm{\checkmark}$}}                      & \multicolumn{1}{c|}{{\color{red}$\bm{\times}$}}                      & \multicolumn{1}{c|}{{\color{teal}$\bm{\checkmark}$}}                      & \multicolumn{1}{l}{\makecell[l]{A survey on collaborative edge inference, including the cloud-device, edge\\-device, cloud-edge-device, and device-device collaborative inference paradigms}}                      \\ \hline
		\multicolumn{1}{c|}{\cite{Shuvo_PIEEE23}}                       & \multicolumn{1}{c|}{{\color{red}$\bm{\times}$}} & \multicolumn{1}{c|}{{\color{red}$\bm{\times}$}} & \multicolumn{1}{c|}{{\color{teal}$\bm{\checkmark}$}} & \multicolumn{1}{c|}{{\color{red}$\bm{\times}$}} & \multicolumn{1}{c|}{{\color{teal}$\bm{\checkmark}$}} & \multicolumn{1}{l}{\makecell[l]{{Emerging trends, tools, and techniques for efficient DNN inference on EUDs,}\\{including designs of architectures, algorithms, hardware, and accelerators}}}       
		\\ \hline
		\multicolumn{1}{c|}{\cite{JJXu_arXiv21}}                       & \multicolumn{1}{c|}{{\color{teal}$\bm{\checkmark}$}} & \multicolumn{1}{c|}{{\color{teal}$\bm{\checkmark}$}} & \multicolumn{1}{c|}{{\color{teal}$\bm{\checkmark}$}}                      & \multicolumn{1}{c|}{{\color{teal}$\bm{\checkmark}$}}                      & \multicolumn{1}{c|}{{\color{red}$\bm{\times}$}}                      & \multicolumn{1}{l}{\makecell[l]{A systematic overview on green deep learning techniques, including energy-\\efficient training,  energy-efficient inference, and efficient data usage strategies}}                       \\ \hline
		\multicolumn{1}{c|}{\cite{LanQ_JCIN21}}                       & \multicolumn{1}{c|}{{\color{teal}$\bm{\checkmark}$}} & \multicolumn{1}{c|}{{\color{teal}$\bm{\checkmark}$}} & \multicolumn{1}{c|}{{\color{teal}$\bm{\checkmark}$}}                      & \multicolumn{1}{c|}{{\color{red}$\bm{\times}$}}                       &  \multicolumn{1}{c|}{{\color{teal}$\bm{\checkmark}$}}                     & \multicolumn{1}{l}{\makecell[l]{Semantic communication principles and techniques for human-to-human, \\human-to-machine, and machine-to-machine applications in edge AI}}                          \\ \hline
		\multicolumn{1}{c|}{\cite{ZhuG_SciChina23}}                       & \multicolumn{1}{c|}{{\color{teal}$\bm{\checkmark}$}} & \multicolumn{1}{c|}{{\color{teal}$\bm{\checkmark}$}} & \multicolumn{1}{c|}{{\color{teal}$\bm{\checkmark}$}}                    & \multicolumn{1}{c|}{{\color{red}$\bm{\times}$}}                     & \multicolumn{1}{c|}{{\color{teal}$\bm{\checkmark}$}}                      & \multicolumn{1}{l}{\makecell[l]{Integrated sensing, communication, and computation for edge intelligence,\\ covering centralized edge learning, federated edge learning, and edge inference}}                       \\ \hline
		\multicolumn{1}{c|}{Ours}                       & \multicolumn{1}{c|}{{\color{teal}$\bm{\checkmark}$}} & \multicolumn{1}{c|}{{\color{teal}$\bm{\checkmark}$}} & \multicolumn{1}{c|}{{\color{teal}$\bm{\checkmark}$}} & \multicolumn{1}{c|}{{\color{teal}$\bm{\checkmark}$}} & \multicolumn{1}{c|}{{\color{teal}$\bm{\checkmark}$}} & \multicolumn{1}{l}{\makecell[l]{{Comprehensive literature review on EE-oriented design approaches for edge AI,}\\{including training data acquisition, distributed edge training, and edge inference}}}       
		\\ \hline
	\end{tabular}
	\label{table:comparisonSurveyPapers}
\end{table*}

\subsection{Related Survey Papers and Our Contributions}

There are several recent survey papers on edge AI \cite{ZhouZ_PIEEE20,KBL_JSAC22,ChenJ_PIEEE19,ShiY_COMST20,XuW_JSTSP23,Baccour_COMST22,XuDianlei_PIEEE21}. Specifically, a survey on edge AI was first presented in \cite{ZhouZ_PIEEE20}, which provides an overview of the key architectures and technologies for training and inference at the mobile edge network. Applications and open challenges of the synergy between DL and MEC were discussed in \cite{ChenJ_PIEEE19}. The communication challenges of edge AI were analyzed in \cite{ShiY_COMST20}, for which, numerous communication-efficient techniques were introduced. Similarly, practical edge AI techniques and their interplays with advanced wireless communication systems were elaborated in \cite{XuW_JSTSP23}. The challenges of edge AI were also examined from the general perspective of resource scarcity in \cite{Baccour_COMST22}. Scalable and trustworthy edge AI systems, built upon the latest innovations in 6G and distributed machine learning (ML) technologies, were discussed in \cite{KBL_JSAC22}. Additionally, the authors of \cite{XuDianlei_PIEEE21} identified the fundamental components of edge AI as edge caching, edge training, edge inference, and edge offloading, and summarized the respective research results.

Some other survey papers focus on specific aspects of edge AI \cite{SzeV_PIEEE17,LimW_COMST20,Mazumder_JESTCS21,LiuD_Neurocomputing22,RenQ_MIR23,Shuvo_PIEEE23}. For instance, the fundamental challenges of model training over the mobile edge network and the corresponding solutions were reviewed in \cite{LimW_COMST20}. Resource-efficient on-device and collaborative edge inference strategies were elaborated in \cite{LiuD_Neurocomputing22} and \cite{RenQ_MIR23}, respectively. Moreover, recent developments of efficient DNN accelerators for edge AI were summarized in \cite{SzeV_PIEEE17,Mazumder_JESTCS21,Shuvo_PIEEE23}. However, there is still a lack of a holistic overview on energy-efficient design approaches for different key tasks in emerging edge AI systems, including data acquisition, edge learning, and edge inference, which motivates the investigation in this paper.
 
This paper presents a timely and comprehensive literature review on EE-oriented design approaches for edge AI. The novel contributions of this study, compared with existing survey papers on edge AI, are summarized in Table \ref{table:comparisonSurveyPapers}. Specifically, we first introduce the three key tasks of edge AI, namely 1) data acquisition for centralized edge learning, 2) distributed edge model training, and 3) edge model inference. Next, we identify general design principles of green edge AI according to the main energy consumption components of practical edge AI systems. These principles provide clear threads to discuss and classify the existing design methodologies for achieving energy-efficient edge AI. In addition, we elaborate on several potential research directions and opportunities that could further improve the EE of edge AI. It is worth noting that a closely related survey on green and energy-efficient AI was presented in \cite{JJXu_arXiv21}. However, it focused on cloud AI and did not address the distinctive challenges in MEC environments. Also, while the discussions in \cite{Baccour_COMST22,XuDianlei_PIEEE21,LanQ_JCIN21,ZhuG_SciChina23} cover all three key tasks of edge AI to certain extents, the EE-oriented design approaches fall outside their primary scopes.

\subsection{Paper Organization}

The rest of this paper is organized as follows. In Section \ref{sec:TaskGeneralPrinciple}, we first introduce the three key tasks of edge AI, followed by a summary of the general design principles of green edge AI. In Sections \ref{DataAcquisitionSec}, \ref{CooperativeTrainingSec}, and \ref{EdgeInferenceSec}, we respectively review the EE-oriented design approaches for the three edge AI tasks, namely data acquisition for centralized edge learning, distributed edge model training, and edge model inference. Section \ref{FutureSec} highlights the potential future research directions of green edge AI, and Section \ref{ConclusionsSec} concludes this paper. Fig.~\ref{fig:organization} summarizes the paper organization.

\section{Key Tasks of Edge AI and General Design Principles of Green Edge AI}~\label{sec:TaskGeneralPrinciple}

In this section, we first introduce the three key tasks in edge AI systems and analyze their respective sources of energy consumption. We then decompose the energy consumption of edge AI systems into sensing{\footnote{Sensing refers to the process of data sampling on EUDs in this paper.}}, communication, and computation energy consumption, based on the configurations of realistic sensing modules, communication technologies, and edge AI platforms. From this analysis, we identify the general design principles for green edge AI.

\subsection{Key Tasks of Edge AI}\label{sec:KeyTasks}

Similar to cloud AI, model training and inference at the mobile edge network are the core tasks of edge AI. However, unlike cloud AI, which leverages existing datasets (e.g., ImageNet and GLUE \cite{WangAlex_ICLR19}) for model training, EUDs in edge AI systems are also responsible for dataset collection to facilitate centralized model training at MEC servers. Additionally, some compute-capable EUDs may perform on-device model training, which enables distributed ML to make the most of the pervasive big data resources. Therefore, \emph{data acquisition for centralized edge model training}, \emph{distributed edge model training}, and \emph{edge model inference} define the three key tasks of edge AI, as elaborated below.

\begin{figure*}[ht!]
	\includegraphics[width=\linewidth]{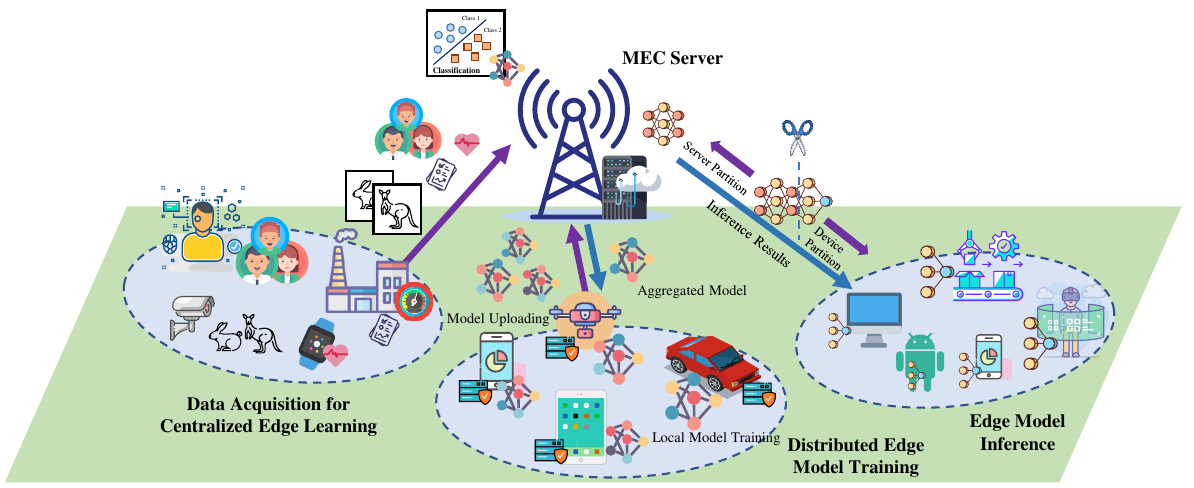}
	\caption{Illustration of the three key tasks in edge AI systems, including: (Left) data acquisition for centralized edge learning, (Middle) distributed edge model training, (Right) edge model inference.}
	\label{fig:sysmodel}
\end{figure*}

\textbf{Data Acquisition for Centralized Edge Model Training:} The objective of centralized edge model training is to train DNN models by leveraging the computational resources at MEC servers. Since raw data are generated at or collected by EUDs, they need to be transmitted to an MEC server prior to model training \cite{LiuD_TCCN21}, as illustrated in Fig. \ref{fig:sysmodel}. This task is essential in scenarios where EUDs have both sensing and communication capabilities, but limited computation power. Typical applications include training student learning activity prediction models using physiology data collected by smart watches \cite{ZhouZ_ICTAI19}, and continuously updating AI models for Internet of Things (IoT) applications with the newly available data \cite{JiaL_IOTJ23}. Notably, the model performance depends on both quality and quantity of the acquired training data at the MEC server, which are restricted by the limited battery energy at EUDs.

\textbf{Distributed Edge Model Training:} With the widespread adoption of smart devices, data is generated anywhere and at any time in mobile edge networks. Although a large volume of data is critical for model training, it is generally challenging to send many data samples from EUDs to the MEC server due to limited bandwidth and energy resources. For some applications, outsourcing data is even prohibited due to privacy concerns. Therefore, distributed model training at the network edge becomes a promising alternative. This approach aims to cooperatively train DNNs across EUDs without gathering their local data at the MEC server. A significant difference between distributed and centralized edge model training is that EUDs perform both communication and local model training. This is becoming feasible as some high-end mobile processors are integrated with graphics and neural processing units. To achieve consensus among EUDs, model updates need to be exchanged. Federated learning (FL) \cite{LiT_MSP20}, as depicted in Fig. \ref{fig:sysmodel}, is a representative distributed edge model training framework.

\textbf{Edge Model Inference:} Model inference refers to the deployment phase of DNN models. For each inference request, such as recognizing the species of an animal in an image, forward propagation is performed to calculate the intermediate variables of a DNN model from the input to the output layer. In cloud AI, a trained DNN is deployed at a cloud server that processes inference data offloaded from EUDs \cite{ZhouZ_PIEEE20}. However, the real-time requirements of many intelligent mobile applications, such as 10 to 100 ms for extended reality and autonomous vehicles, cannot be satisfied due to excessive communication latency. Therefore, model inference is also being migrated toward the network edge, termed \emph{edge inference}, for which, harnessing the computational resources at both EUDs and MEC servers is essential. A simple yet effective approach is to partition a DNN model between an EUD and an MEC server as shown in Fig.~\ref{fig:sysmodel}, which balances the communication overhead and on-device computation workload \cite{ShaoJ_MCOM20}.

The three key tasks of edge AI involve heterogeneous sources of on-device energy consumption, with different operations contributing to varying degrees. Specifically, the energy consumption of data acquisition at EUDs for centralized edge model training primarily includes sensing and communication energy, although some energy may also be consumed for data preprocessing. In the cases of distributed edge model training and edge model inference, communication and computation energy consumption dominate, although the energy cost of training and inference data acquisition at EUDs cannot be disregarded in practical systems. This survey paper focuses on techniques aiming at reducing on-device energy consumption, driven by the limited battery capacities of EUDs. To provide a clear and comprehensive overview, we summarize the main energy consumption components of different edge AI tasks in TABLE \ref{energyAItasks}.

\begin{table*}[ht!]
	\caption{Main Energy Consumption Components at EUDs of Different Edge AI Tasks}
	\centering
	\begin{tabular}{m{6.5cm}|m{1.8cm}|m{1.8cm}|m{1.8cm}  }
		\hline
		Edge AI Task & Sensing & Communication & Computation\\
		\hline
		\hline
		Data Acquistion for Centralized Edge Model Training   &  {\color{teal}$\bm{\checkmark}$}    &  {\color{teal}$\bm{\checkmark}$} &  {\color{red}$\bm{\times}$} \\
		\hline
		Distributed Edge Model Training &  {\color{red}$\bm{\times}$}  & {\color{teal}$\bm{\checkmark}$}   & {\color{teal}$\bm{\checkmark}$} \\
		\hline
		Edge Model Inference & {\color{red}$\bm{\times}$} & {\color{teal}$\bm{\checkmark}$} &  {\color{teal}$\bm{\checkmark}$} \\
		\hline
	\end{tabular}
	\label{energyAItasks}
\end{table*}

\subsection{Breakdown of Energy Consumption}
\label{sec:BreakdownEnergy}

In order to enhance the understanding of how the sensing, computation, and communication operations impact energy consumption in edge AI systems, this subsection aims to analyze the configurations of different sensing modules, communication technologies, and edge AI platforms. Consequently, the energy consumption will be dissected into three distinct categories, as elaborated below.

\textbf{Sensing Energy:} In TABLE~\ref{Sensingenergy}, the sensing power consumption of four sensing modules is presented, including a 5 megapixel (MP) imaging sensor \cite{OV5675}, a LiDAR camera \cite{Intel_CameraL515}, a radar sensor \cite{TI_AWRL1432}, and an optical sensor for heart rate monitoring \cite{ROHM_BH1790GLC}. It can be concluded from TABLE~\ref{Sensingenergy} that the energy consumed during sensing operations is primarily influenced by the sensing mechanism, sampling rate, and sampling quality. For instance, an imaging sensor can consume much higher power (96 mW) compared to an optical sensor for heart rate monitoring (0.72 mW). Besides, the power consumption of a sensing module in active and standby modes are in a sharp difference, e.g., 96 mW and 165 $\mu$W respectively for the imaging sensor, implying the significance of sampling frequency. Furthermore, configurating the LiDAR camera with different depth resolutions (i.e., VGA and XGA) results in varying power consumption. This fact is also exemplified by the case of the radar sensor, again confirming the critical impact of sampling quality on sensing energy consumption.

\textbf{Communication Energy:}
Efficient and reliable communication between MEC servers and EUDs is crucial for their collaborative functioning in all three key tasks of edge AI. The power consumption associated with communication depends on many factors, such as carrier frequency, modulation scheme, transmission data rate, and antenna deployment. For instance, an EUD can consume up to 15 dBm ($\sim$32 mW) in a 2.4 GHz WiFi network \cite{Wifienergy}. More recently, fifth-generation (5G) new radio (NR) offers two frequency modes in sub-6 GHz and millimeter-wave bands, allowing EUDs transmit at the maximum power of 23 dBm ($\sim$199 mW) and 26 dBm ($\sim$398 mW) respectively \cite{3GPP_TS38101}. While numerous communication technologies have been developed, Bluetooth, WiFi, and cellular communication are particularly suitable for edge AI applications, considering the requirements for data rate and transmission range. To achieve energy-efficient edge AI, it is crucial to select the appropriate communication technology that meets the application's needs while minimizing power consumption. Moreover, rather than perceiving the communication channel as a mere conduit for data transmission, it is necessary to adapt existing communication protocols to the specific characteristics of edge AI tasks.

\textbf{Computation Energy:} The majority of computations performed by EUDs in edge AI systems are dedicated to distributed model training and model inference. As depicted in TABLE \ref{Comptenergy}, the power consumption associated with computation varies across different edge AI platforms, typically ranging from a few to a few tens of Watts. While the exact power consumption is affected by numerous factors, there is generally a positive correlation between processing speed and power consumption. Additionally, the types of AI workloads directly impact the power consumption. For instance, training a ResNet-110 \cite{HeKaiMing_CVPR16} and VGG-16 \cite{SimonyanK_ICLR15} model on the NVIDIA Jeston TX2 platform requires approximately $8\times 10^5$ and $3.8\times10^5$ Joules of energy, respectively \cite{WuY_TCAD20}. These energy requirements correspond to approximately 17.5 and 8.3 times the battery capacity of an iPhone 14 (i.e., 3279 mAh operating at 3.87 V). Similarly, utilizing different DNN models for inference results in varying levels of computation energy consumption. Furthermore, it has been observed that specialized AI accelerators, such as tensor processing units (TPUs), are more energy-efficient compared to general-purpose graphics processing units (GPUs).

In conclusion, the energy consumption in edge AI systems is influenced by different factors for each of the three sources discussed. Moreover, for each edge AI task, there is a distinctive interplay among different components of energy consumption as will be detailed in the sequel. Consequently, aiming at minimizing energy consumption associated with sensing, communication, and computation, customized approaches need to be developed based on the overarching design principles for energy-efficient edge AI, which will be explored in the next subsection.

\begin{table*}[h]
	\caption{Power Consumption of Different Sensing Modules on EUDs}
	\centering
	\begin{tabular}{ m{2cm}|m{3.0cm}|m{3.0cm}|m{3.0cm}|m{3.0cm}}
		\hline
		\multicolumn{1}{c|}{\makecell[l]{Model}} & \multicolumn{1}{c|}{\makecell[l]{OMNIVISION\\OV5675}} & \multicolumn{1}{c|}{\makecell[l]{Intel RealSense\\ LiDAR Camera L515}}
		& \multicolumn{1}{c|}{\makecell[l]{ROHM Optical\\Sensor BH1790GLC}} & \multicolumn{1}{c}{\makecell[l]{Automotive Radar\\Sensor AWRL1432}} \\
		\hline\hline
		\multicolumn{1}{c|}{\makecell[l]{Functionality}}   & \multicolumn{1}{c|}{\makecell[l]{5MP imaging sensor\\for smartphones}} &  \multicolumn{1}{c|}{\makecell[l]{High-resolution LiDAR\\ depth camera for indoor\\ applications}}
		&  \multicolumn{1}{c|}{\makecell[l]{Heart rate monitor \\for wearable devices}} & \multicolumn{1}{c}{\makecell[l]{Millimeter wave radar \\sensor operating in the\\ 76-81 GHz band}}\\
		\hline
		\multicolumn{1}{c|}{\makecell[l]{Power\\ consumption}} & 
		\multicolumn{1}{c|}{\makecell[l]{Active: 96~mW \\
				Standby: 165~$\mu$W \\
				Xshutdown: 1~$\mu$W}}
		& \multicolumn{1}{c|}{\makecell[l]{Idle: 0.8~W\\
				Depth (VGA): 3.0~W\\
				Depth (VGA) + RGB\\(1080p, 30FPS): 3.2~W\\
				Depth (XGA): 3.1~W\\
				Depth (XGA) + RGB\\(1080p, 30FPS): 3.3~W}}
		&  \multicolumn{1}{c|}{\makecell[l]{Active: 0.5$\sim$0.72~mW \\
				Standby: 2$\sim$2.88~$\mu$W}} & \multicolumn{1}{c}{\makecell[l]{Sampling rate: 12.5 MSps \\
				RF front-end: \\
				(2Tx, 3Rx): 1.358 W \\
				(2Tx, 2Rx): 1.267 W \\
				(1Tx, 2Rx): 1.013 W \\
				(1Tx, 1Rx): 0.945 W \\
				Processing: 159 mW \\
				Idle: 13.80 mW \\	
				Deep sleep: 1.38 mW}}
		\\
		\hline
	\end{tabular}
	\centering
	\label{Sensingenergy}
\end{table*}

\begin{table*}[ht]
	\caption{Computational Resources, Processing Speed, and Computation Power Consumption of Different Edge AI Platforms$^{\dagger}$}
	\centering
	\begin{tabular}{ m{2.55cm}|m{2.55cm}|m{2.55cm}|m{2.55cm}|m{2.55cm} }
		\hline
		\multicolumn{1}{c|}{\makecell[l]{Model}}  & \multicolumn{1}{c|}{\makecell[l]{RPi-4B}} & \multicolumn{1}{c|}{\makecell[l]{Jetson Nano}} & \multicolumn{1}{c|}{\makecell[l]{Jetson TX2}}  & \multicolumn{1}{c}{\makecell[l]{Coral Dev Board}} \\
		\hline\hline
		\multicolumn{1}{c|}{\makecell[l]{CPU}}    & \multicolumn{1}{c|}{\makecell[l]{Broadcom BCM2711\\ (Quad-core \\ Cortex-A72)}}  & \multicolumn{1}{c|}{\makecell[l]{Quad-core ARM \\ Cortex-A57 MPCore \\ processor}}  & \multicolumn{1}{c|}{\makecell[l]{Dual-core NVIDIA\\ Denver 2 64-bit CPU \\ and Quad-core ARM \\ Cortex-A57 MPCore \\ processor }}   & \multicolumn{1}{c}{\makecell[l]{NXP i.MX 8M SoC \\ (Quad Cortex-A53, \\Cortex-M4F)}}   \\
		\hline
		\multicolumn{1}{c|}{\makecell[l]{GPU}}   & \multicolumn{1}{c|}{\makecell[l]{N.A.}}  &  \multicolumn{1}{c|}{\makecell[l]{NVIDIA Maxwell\\architecture GPU\\ with 128 CUDA cores}}  & \multicolumn{1}{c|}{\makecell[l]{NVIDIA Pascal\\architecture GPU\\with 256 CUDA cores}}    &  \multicolumn{1}{c}{\makecell[l]{Integrated GC7000\\Lite Graphics}}  \\
		\hline
		\multicolumn{1}{c|}{\makecell[l]{TPU}}   & \multicolumn{1}{c|}{\makecell[l]{N.A.}}  &  \multicolumn{1}{c|}{\makecell[l]{N.A.}}  & \multicolumn{1}{c|}{\makecell[l]{N.A.}}  & \multicolumn{1}{c}{\makecell[l]{Google Edge TPU}} \\
		\hline
		\multicolumn{1}{c|}{\makecell[l]{Memory}}  & \multicolumn{1}{c|}{\makecell[l]{1/2/4/8GB LPDDR4\\-3200 SDRAM}}   &  \multicolumn{1}{c|}{\makecell[l]{4 GB 64-bit LPDDR4}}  & \multicolumn{1}{c|}{\makecell[l]{8GB 128-bit LPDDR4}}  & \multicolumn{1}{c}{\makecell[l]{1/2/4 GB LPDDR4}} \\
		\hline
		\multicolumn{1}{c|}{\makecell[l]{Processing speed}}  & \multicolumn{1}{c|}{\makecell[l]{9.69 GFLOPS}} &  \multicolumn{1}{c|}{\makecell[l]{472 GFLOPS}}  & \multicolumn{1}{c|}{\makecell[l]{1.33 TFLOPS}} & \multicolumn{1}{c}{\makecell[l]{4 TFOPS}} \\
		\hline
		\multicolumn{1}{c|}{\makecell[l]{Power consumption}} &  \multicolumn{1}{c|}{\makecell[l]{Idle: $2.8$ W \\ Average: $7.2$ W \\Maximum: $8.2$ W}} &  \multicolumn{1}{c|}{\makecell[l]{5$\sim$10 W}} 
		& \multicolumn{1}{c|}{\makecell[l]{7.5$\sim$15 W}} & \multicolumn{1}{c}{\makecell[l]{4 W}} \\
		\hline
	\end{tabular}
	
	\centering
	\begin{tablenotes}
		\item $^{\dagger}$The processing speed and power consumption of RPi-4B are measurement results of the 4 GB memory model available at:  \url{https://web.eece.maine.edu/~vweaver/group/green_machines.html}. The processing speed and power consumption of Jeston Nano and Jetson TX2 are available at the official website of NVIDIA: \url{https://developer.nvidia.com/embedded/jetson-modules}. The processing speed and power consumption of the Cora Dev Board are based on the specification of the Google Edge TPU available at: \url{https://coral.ai/static/files/Coral-Dev-Board-datasheet.pdf}. 
	\end{tablenotes}
	\label{Comptenergy}
\end{table*}

\subsection{General Design Principles}

To effectively reduce energy consumption at EUDs for edge AI tasks, a wide range of approaches can be developed. In order to provide a cohesive structure for the literature review in the upcoming sections, we will first establish the general design principles for energy-efficient edge AI. These principles are derived from the energy breakdown analysis of edge AI systems presented in Section \ref{sec:BreakdownEnergy}.

\textbf{Energy Efficiency-oriented Optimization:} Shifting the design objective of edge AI systems from solely maximizing intelligence performance (e.g., classification accuracy) to maximizing EE is crucial. Specifically, EE of AI systems can be defined as the amount of intelligence obtained with per Joule of energy consumption \cite{MWelling_ICML18}. Optimizing edge AI systems for EE necessitates a holistic consideration on all the three critical energy components discussed in Section \ref{sec:BreakdownEnergy}. Given that most edge AI applications involve integrated processes of multiple operations, it is essential to investigate the intricate interplay among sensing, computation, and communication in order to strike a decent balance between achievable intelligence and energy consumption, which vary drastically for three edge AI tasks. Therefore, achieving the vision of green and energy-efficient edge AI requires not only significant algorithmic innovations with EE-oriented design objectives, but also fundamental re-architecting of existing mobile networks to enable the all-round cooperation between EUDs and MEC servers to best utilize their computational resources.

\textbf{Adaptation to System Dynamics:} Exploiting the dynamics of edge AI systems can enhance EE through adaptive sensing, computation, and communication. For instance, in the case of a moving smart vehicle, the surroundings constantly change, necessitating timely adjustments in the sampling frequency and resolution of a vehicular LiDAR camera based on the scene's informativeness. Regarding information exchange between EUDs and MEC servers, since wireless channels exhibit time-varying signal attenuation, adapting transmission schemes, such as power control, modulation, and coding, to the instantaneous channel conditions can improve the EE of uplink communication \cite{Hasan_COMST11}. Furthermore, the unique characteristics of edge AI computations, including workload, latency, and accuracy requirements, vary significantly. Therefore, optimizing processing strategies to achieve maximum energy savings at EUDs is also essential \cite{MaoY_COMST17}. Another dimension of system dynamics in edge AI that should be considered is the \emph{task state}, which can refer to the progress of a learning task or the sample difficulty of an inference task. In addition, it is important to monitor the arrivals/departures of EUDs and their requested edge AI services to facilitate energy-efficient dynamic resource allocation.

\textbf{Trading Intelligence for Greenness \& Energy Efficiency:} As AI algorithms, such as DNN models, approach their performance limits, the energy consumption required to achieve further intelligence improvement increases super-linearly \cite{Desislavov_SusComput23,Sevilla_IJCNN22}. Therefore, it is crucial to develop application-specific solutions that can provide \emph{just-enough intelligence}, rather than blindly adopting state-of-the-art approaches. Besides, the fault-tolerance property of DNN models \cite{HUITZIL_ACCESS17,RuospoA_Comput23} enables to control the trade-off between intelligence and reduced energy consumption. This can be achieved by identifying \emph{what to compute}, in addition to optimizing \emph{where and how to compute}, as typically considered for generic computation tasks. In other words, computations that have minimal impact on intelligence performance should be avoided. Such a principle can also be applied to the sensing and communication operations in edge AI systems. For instance, the data scaling law for model training~\cite{Rosenfeld_ICLR20,Johnson_ACL20} indicates that increasing the training dataset size beyond a limit has saturating benefit on accuracy. Thus, sensing and communication energy consumption in centralized edge model training can be reduced by collecting just enough amount of data samples for a target model accuracy. However, due to the lack of a universal and tractable characterization of the achievable intelligence and the energy consumption of sensing, computation, and communication, the optimization of their trade-off needs to be task-oriented with prior information obtained through offline profiling or online learning.

\begin{figure*}
	\includegraphics[width=\linewidth]{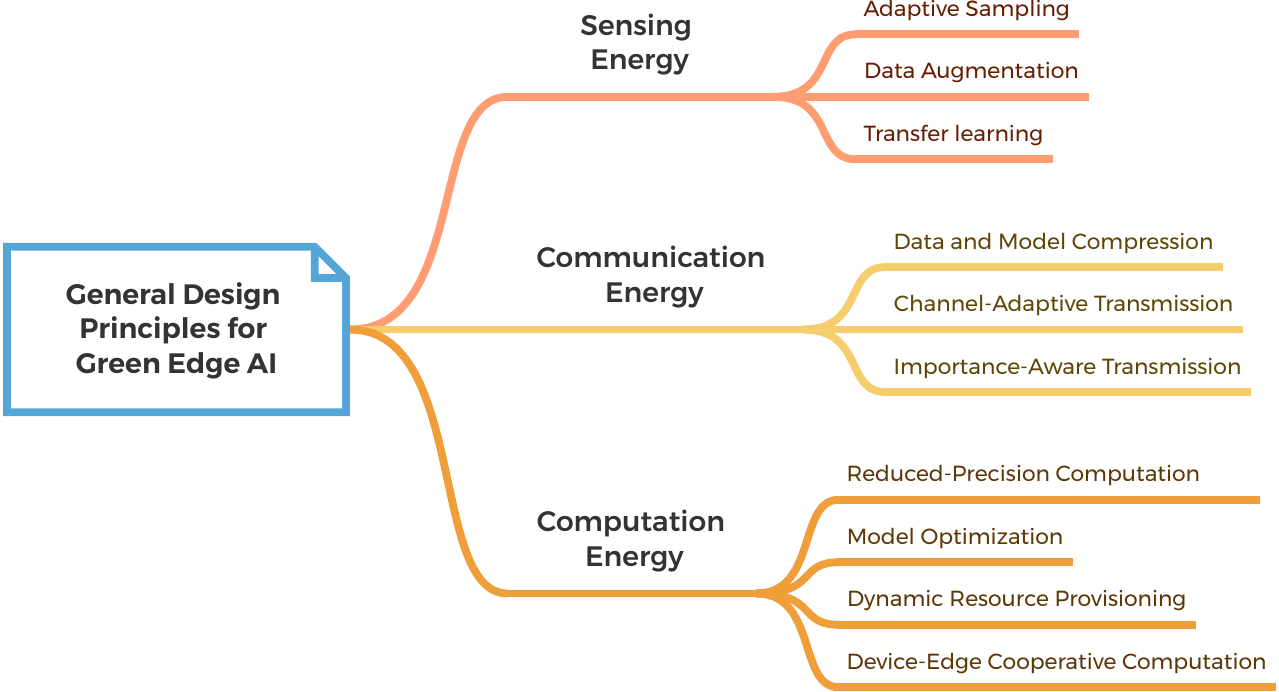}
	\caption{Energy-saving techniques for sensing, communication, and computation operations in edge AI systems.}
	\label{fig:principles}
\end{figure*}

Based on the aforementioned general design principles, we summarize the energy-saving techniques for sensing, communication, and computation operations in Fig.~\ref{fig:principles}, which will be discussed for different edge AI tasks in the following sections.

\section{Energy-Efficient Data Acquisition for Centralized Edge Learning}\label{DataAcquisitionSec}

EUDs are able to collect numerous data on their surroundings. In order to distill insights from these data, AI models need to be trained. However, because of the cost and form factor constraints, many EUDs are just equipped with limited processing power and a small-size battery. Hence, their model training tasks need to be outsourced to an MEC server, which is referred to as \emph{centralized edge model training} (a.k.a. \emph{centralized edge learning}~\cite{ZhuG_MCOM20}). 

Although the computations of model training are offloaded in centralized edge learning, sensing the surroundings (i.e., data sampling) and transmitting the sensed data to the MEC server contribute to most of the on-device energy consumption~\cite{CarlosJ_TCSVT17,Likamwa_MPRV21}, which increases with the required amount of training data samples. In this section, we investigate energy-efficient data acquisition approaches for centralized edge learning, which answer the following three questions: 1) How to perform energy-efficient data sampling (Section~\ref{acquisition:sampling}); 2) How to transmit these data to the MEC server with minimal energy consumption (Section~\ref{acquisition:transmission}); and 3) How to improve the learning performance when only a small training dataset can be acquired by the MEC server (Section~\ref{acquisition:smalldata}).

\subsection{Adaptive Sampling}\label{acquisition:sampling}
As model training tasks are data-intensive, energy consumed by the sensing modules of EUDs to collect a vast amount of environmental measurements is significant. Wireless sensor networks (WSNs) are typical venues where data are periodically sampled by the distributed sensor nodes. According to the Nyquist Theorem~\cite{Giouroukis_DEBS20}, the data sampling rate should be sufficiently high to ensure artifact-free sensing. However, constant-rate sampling fails to exploit the time-varying property of the sensed data and easily results in redundant data samples. In contrast, adaptive sampling, which reacts to behaviors of the sensed data by adjusting the sampling rate in runtime, is effective in reducing the sampling energy cost~\cite{Likamwa_MPRV21,UTKUDEMIRELB_ACMJCOMPHEALTH22}. For centralized edge learning, it also saves the communication energy since less amount of data needs to be transmitted. 

There are lots of discussions on adaptive sampling strategies for WSNs~\cite{Giouroukis_DEBS20}. For instance, a two-stage adaptive sampling scheme was proposed in~\cite{KarakiA_IWCMC19}. It first eliminates the redundant data for transmission at each sensor according to their similarity. Then, a sink node determines the new sampling rates for sensors in the next aggregation round by exploiting the spatio-temporal correlation of received data. Compared with a naive approach, adaptive sensing reduces up to $60\%$ of the sensing energy, and the similarity search further saves $93\%$ of the transmission energy, while incurring $<7\%$ data accuracy loss. The principle of adaptive sampling was also utilized in MEC systems~\cite{LouP_Sensors20,Sushmita_TGCN22}. In~\cite{LouP_Sensors20}, a data-driven method that uses the latest collected data by EUDs to adjust the sampling frequency was developed using linear fitting. It was observed that the energy consumption of different sensors at an edge node is reduced by $12.86$$\sim$$35.83\%$ compared with a constant-rate sampling strategy. Besides, an edge intelligence-based priority-aware sensing and transmission framework was presented in~\cite{Sushmita_TGCN22}, where in every measurement cycle, a subset of sensors are selected based on both the spatial and temporal correlation of their sensing signals, and the sampling interval of each selected sensor is dynamically adapted based on the temporal prediction error. This framework saves up to $41\%$ of the sensing energy compared to several baseline approaches.

Although most research efforts on adaptive sampling focused on data reconstruction quality (e.g.~\cite{LouP_Sensors20,Sushmita_TGCN22}), a few recent studies turned their attention to model training for emerging edge AI applications~\cite{ChengW_AAAI18,Siddique_JTRC19}. For human activity recognition, a dynamic sampling policy for built-in sensors of smartphones and wearables named datum-wise frequency selection was developed in~\cite{ChengW_AAAI18} to balance the recognition accuracy and EE. Specifically, an optimization problem on the classification model and sampling frequencies was formulated to minimize the weighted sum of recognition error and sensing energy cost, which was solved via a continuous-state Markov decision process (MDP) formulation. A self-adaptive sampling method was proposed in~\cite{Siddique_JTRC19} to collect the Global Positioning System-based vehicular trajectory data. A hidden Markov model based classifier was adopted to estimate the vehicle flow state from its trajectory and a support vector machine was implemented to identify the ``stop'' and ``go'' segments in the trajectory, based on which, the sampling frequency is adjusted. This method reduces the amount of sampled data by $\sim$$70\%$ without losing the most critical data points on a queue location estimation application.

\subsection{Learning-aware Data Sample Transmission}\label{acquisition:transmission}

Albeit some redundant data can be eliminated using adaptive sampling strategies, the remaining ones exhibit different importance to centralized edge learning tasks, bringing ample opportunities for further transmission energy reduction. First, data samples can be prioritized and transmitted so the less important ones can be discarded~\cite{LiuD_TCCN21,LiuD_TWC21}. Second, data samples can be transmitted in reduced quality via lightweight pre-processing such as lossy compression~\cite{ZhouS_ICC20}. For both methods, balancing the communication cost and learning performance is crucial. 

Data sample importance typically changes at different stages of the learning process. To obtain such information at EUDs, the MEC server needs to feedback the newest learned model, denoted as $\mathbf{w}$, for data importance evaluation~\cite{LiuD_TCCN21,LiuD_TWC21,ZhouS_ICC20}. In~\cite{ZhouS_ICC20}, the importance of a sample $\mathbf{x}$ is determined by the training loss $l\left(\mathbf{x};\mathbf{w}\right)$, of which, a larger value implies higher significance. However, re-calculating the training loss for a large number of data samples whenever the model is updated is both time- and energy-consuming. One solution is to restrict the importance evaluation on a small subset of training data. It was validated on the MNIST dataset that, only $2.5\%$ of the data samples need to be evaluated to maintain similar learning accuracy. Data importance evaluation is also closely related to the burgeoning area of \emph{active learning}~\cite{SettlesB_2012}, where prediction uncertainty of a data sample under the newest learned model is commonly adopted~\cite{LiuD_TCCN21,LiuD_TWC21}. Besides, based on the heuristics that training with highly disparate data samples may produce better models, the Euclidean distance between a data sample of interest and centroid of all available samples at the MEC server, was considered as the data importance metric in~\cite{ZengZ_WCL21}. The concept of data sample importance was also leveraged for edge robotic systems in \cite{ChinchaliS_arXiv20}, where a lightweight HARVESTNET was developed to determine whether to cache and transfer a training sample to a remote server for model re-training.

In contrast to conventional wireless communication systems, energy-efficient transmission schemes for centralized edge learning should consider the data sample importance. In~\cite{Khong_IECON20}, a multi-criteria training data subset (MCTS) algorithm for prioritizing data transmission in a predictive maintenance application was developed, where the high-quality data samples are protected against wireless signal attenuation with more resources. Besides, data compression before transmission was proposed for centralized edge learning, where the compression ratio is determined according to the training loss in~\cite{ZhouS_ICC20} and via an error-bounded compression scheme in~\cite{AzarJ_FGCS19}. To avoid the complex baseband processing at EUDs, analog modulation was exploited in~\cite{LiuD_TCCN21,LiuD_TWC21,ZengZ_WCL21} for wireless transmission, where data samples are directly mapped to channel symbols and thus vulnerable to noise corruption. As a result, a retransmission mechanism was developed in~\cite{LiuD_TWC21} to eliminate negative effects of noise, where a data sample is retransmitted until the effective signal-to-noise ratio (SNR) is above a data importance-adaptive threshold. With multiple EUDs, it is also necessary to schedule the most appropriate ones to transmit, which minimizes the communication overhead for a target learning performance~\cite{LiuD_TCCN21,ZengZ_WCL21}.

Apart from sample-level importance, the amount of data used for model training also has a great impact on the learning performance. Although the exact relationship between data volume and training loss is hard to find, data-driven methods are able to establish power-law models for approximate characterizations~\cite{Johnson_ACL20}. By adopting a nonlinear function mapping the data volume to the training loss, a learning-centric power allocation problem for centralized edge learning was solved in~\cite{WangShuai_TWC20} to decide the amount of data samples used for the training of multiple classification models. This study was extended in~\cite{ZhouLiangKai_TVT21} and \cite{XieH_TWC23}, respectively incorporating wireless resource allocation and multi-device scheduling. More realistic scenarios accounting for continuous data arrivals and limitations of MEC servers were investigated in~\cite{LiXY_TWC22,Merluzzi_ACCESS21}.

\subsection{Centralized Edge Learning with Small Datasets}\label{acquisition:smalldata}
Beyond the techniques discussed in Sections~\ref{acquisition:sampling} and~\ref{acquisition:transmission} on energy-efficient data acquisition for centralized edge learning, alternatives to improve the learning performance given a small dataset at the MEC server are explored in this subsection, including data augmentation and knowledge transfer, of which, the basic principles and benefits are elaborated.

\emph{1) Data Augmentation:} Data augmentation (DA) refers to the process of enhancing the size, quality, and diversity of a training dataset~\cite{Shorten_IOTJ21}, hoping to obtain a more accurate model than using the original dataset. For instance, basic spatial operations such as rotation and noising, can be applied to images for DA. In addition to handcrafted methods, generative adversarial network (GAN) based image synthesis~\cite{HuangH_arXiv18} becomes a new favorite to increase the robustness and generalizability of DA~\cite{Maayan_ISBI18,Sandfort_SciReport19}. DA methods for other data types are less investigated, despite with recent advances on speech recognition~\cite{ParkD_INTERSPEECH19}, video analytics~\cite{YunS_arXiv20}, and NLP~\cite{FengY_IJCNLP21}.

\begin{figure*}[ht!]
	\includegraphics[width=18cm]{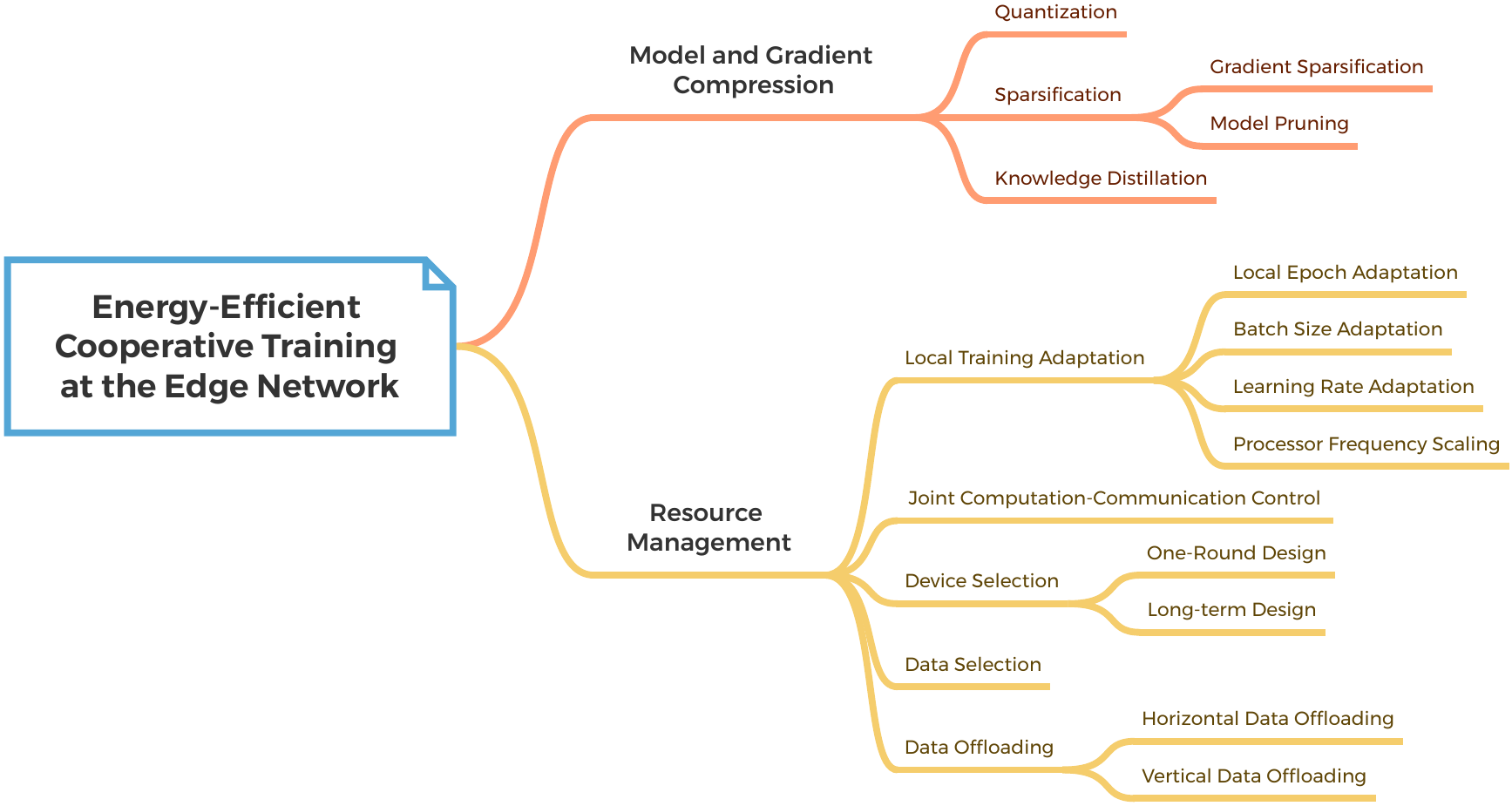}
	\caption{Energy-efficient design approaches for cooperative training at the edge network.}
	\label{fig:FLtree}
\end{figure*}

DA has found its adoptions in a variety of edge AI applications. In~\cite{MaximilianE_ICASSP20}, the feasibility of generating sequential IoT data via GAN for classification was validated, even with a small volume of training data. The hybrid smart medical architecture for electrocardiogram diagnostics developed in~\cite{HeY_ApplSci20} adopts a similar approach to overcome the class imbalance issue in the training dataset. For WiFi signal-based human activity recognition, non-learning based data synthesis was utilized in~\cite{ZhangJ_IoTJ21} to mitigate the negative impacts caused by human movement and subject-specific bias. We are aware that while~\cite{MaximilianE_ICASSP20,HeY_ApplSci20,ZhangJ_IoTJ21} did not directly address the energy consumption of EUDs, DA improves the learned model performance in the presence of a small and possibly low-quality training dataset. In other words, the amount of data transmitted from EUDs to the MEC server and their quality can be reduced to achieve the target learning performance with DA, thereby saving the sampling and communication energy.

\emph{2) Transfer Learning:} The objective of transfer learning (TL) is to reuse the learned knowledge for a new but related model training task~\cite{PanJ_TKDE10}. An attractive benefit of knowledge transfer techniques for centralized edge learning is the reduced requirement of labeled training data~\cite{ZhuangF_ProcIEEE21}, therefore minimizing the cost of training data acquisition, including the sensing and communication energy. Besides, since MEC environments and data distributions may experience regular drifts, AI models need periodical fine-tuning~\cite{SufianA_CISS21,YangJ_IoTJ20}. This can also be assisted with TL to avoid re-training models entirely from scratch on large amounts of labeled data. 

The developments of industrial IoT applications have been actively utilizing TL techniques. The use of TL for industrial IoT component recognition was investigated in~\cite{LiuX_IoTJ21}, where a pre-trained VGG-16 network on ImageNet was transferred to a new scenario with a small target domain T-Less dataset of texture-less objects. Compared with training a convolutional neural network (CNN) model from scratch, knowledge transfer reduces both the required training data volume and training time, while attaining better training accuracy. For edge camera image recognition in unmanned stores, the experimental results in~\cite{LuCH_IOTJ21} showed that with an elite-instance-based matching approach that appropriately selects the images to be used for TL, up to $70\%$ of source image samples can be eliminated from transmission to the target cameras. TL was also utilized for device-free crowd counting based on the WiFi channel state information (CSI)~\cite{Khan_IoTJ22}. In situations where data acquisition is extremely costly and energy-expensive, some data categories may only have a few samples for centralized model training at the MEC server, for which, few-shot learning approaches~\cite{WangY_ACMComputSurveys20} can be leveraged. Sample edge AI applications include object detection~\cite{ZengY_TSUSC20}, text sentiment analysis~\cite{YangL_IOTJ21}, and IoT traffic classification~\cite{ZhaoZ_CL22}.

\subsection{Takeaways}

This section summarizes energy-efficient data acquisition methods for centralized edge learning, which were developed following either of the two approaches: 1) To improve the EE of data sampling and transmission; 2) To improve the learning performance with only a small training dataset available at the MEC server. For the former approach, tractable models relating various decision variables, e.g., data sampling frequency, resolution, and context, with the learning performance are essential, which can be obtained via offline data-fitting or online learning. Although their effectiveness has been observed, existing works are mainly based on heuristics without a strong theoretical support. Besides, accurate energy models of sensing modules on EUDs are necessary to formulate valid optimization problems. For cameras, the energy model developed for visual computing systems in~\cite{CarlosJ_TCSVT17,Likamwa_MPRV21} can be used. Also, a power model suitable for three-dimensional depth sensors (e.g., LiDAR cameras) is available in \cite{TilmonB_CVPR23}. Moreover, in-depth studies to unveil the impacts of deploying more EUDs for sensing on EE should be conducted. The second approach stems from the fields of transfer learning and few-shot learning, while other techniques, such as meta-learning-based methods, may also be applied~\cite{PanJ_TKDE10,WangY_ACMComputSurveys20}. To further reduce the communication energy consumption, it is important to understand the minimum amount of training data that needs to be transmitted from EUDs to the MEC server with the joint aid of DA and TL techniques. Since the three classes of techniques reviewed respectively in Section \ref{acquisition:sampling}, \ref{acquisition:transmission}, and \ref{acquisition:smalldata} were examined separately, their composite effects on achieving energy-efficient data acquisition for centralized edge learning deserve further investigations. 

\section{Energy-Efficient Cooperative Training at the Edge Network}\label{CooperativeTrainingSec}

Centralizing the model training process at an MEC server necessitates substantial communication bandwidth to transfer large volumes of training data, a requirement that is not always feasible in wireless edge networks. This approach may also infringe upon data privacy regulations \cite{Rubaie_IEEESP19} as local data may be private and confidential. However, advancements in embedded hardware technologies have empowered a growing population of EUDs to perform on-device model training \cite{TensorFlow_onDeviceTraining,DasA_arXiv22}. As such, cooperative model training at the edge network {(a.k.a. cooperative edge learning)}, which distributes a model training task among multiple EUDs, has been extensively explored to leverage both the ubiquity of big data and computational resources. In this section, we focus on federated learning \cite{YangQ_ACMTIST19}, a \emph{de facto} standard for privacy-preserving cooperative training, and summarize the energy-efficient design approaches in Fig. \ref{fig:FLtree}. We remark that these approaches for FL are also applicable to other cooperative training paradigms at the edge network, such as decentralized edge learning (DEEL)~\cite{LiuW_TSIPN22} and hierarchical federated edge learning (H-FEEL)~\cite{LiuL_TWC23}, which are outlined in Section~\ref{otherCoLearnFrameworks}. 

\subsection{Background of Federated Learning}

\begin{figure*}[t!]
	\includegraphics[width=18cm]{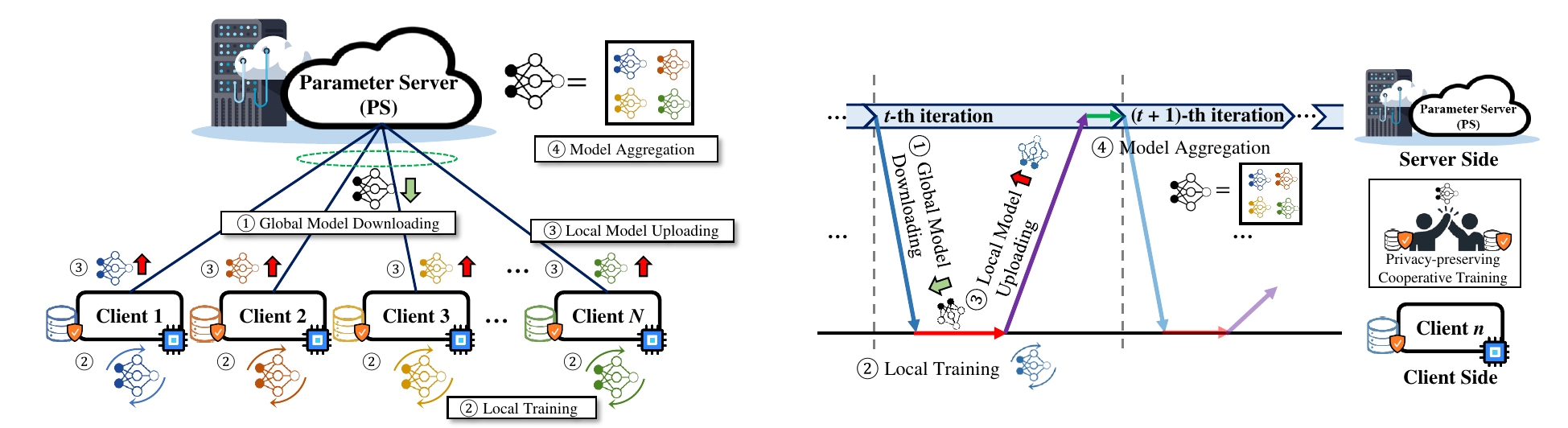}
	\caption{A typical federated learning system (Left) and its training procedures (Right).}
	\label{fig:FLmodel}
\end{figure*}

FL was first introduced by Google in 2016 \cite{McMahanB_AISTATS17}. In this framework, a set of client nodes (e.g., EUDs) collaborate to train an ML model under the coordination of a parameter server (PS), as illustrated in Fig. \ref{fig:FLmodel}. During each training iteration, clients initially download the global model from the PS. Subsequently, each client independently updates the downloaded model using its local data, typically through methods such as stochastic gradient descent (SGD). Following this, the local model updates are uploaded to the PS for global aggregation. As the training data remains local to the clients, data privacy is significantly better protected compared to the centralized edge learning paradigm discussed in Section \ref{DataAcquisitionSec}.

Although initial FL systems utilized cloud-based PSs, there has been a growing interest in deploying FL at the resource-constrained mobile edge network, a concept known as \emph{federated edge learning} (FEEL). This approach holds potential for facilitating rapid cooperative training \cite{LimW_COMST20,ChenM_JSAC21} with an MEC server being designated as the PS. Given that FEEL includes multiple iterations of local training (e.g., backpropagation) and model uploading, the learning performance is predominantly associated with the energy consumption of both the computation and communication operations. Therefore, techniques to maximize the EE of FEEL mostly optimize either or both of the local training and model transmission processes without causing significant degradation to learning performance. In the following two sections, we introduce two categories of techniques as potential pathways to realize energy-efficient FEEL, including model and gradient compression (Section \ref{FL_Compression}), and resource management (Section \ref{FL_RM}). We discuss these techniques from an EE perspective, although some of them are also essential for communication-efficient FEEL \cite{ShiY_COMST20}.

\subsection{Model and Gradient Compression}\label{FL_Compression}

In each training iteration, the model updates to be uploaded by EUDs take the form of model weights or model gradients. Compressing these model updates can reduce communication energy consumption, as fewer bits need to be transmitted. In addition, locally training compressed models can effectively minimize on-device computation energy consumption, as it can avoid substantial floating-point/MAC and memory access operations~\cite{Justus_BigData18,Kim_TWC22}. Typical compression approaches such as quantization, sparsification, and knowledge distillation are often employed to achieve energy-efficient FEEL.

\subsubsection{Quantization}
In FEEL, quantization employs low bitwidths to represent DNN model parameters on-device. This technique can also be applied to model gradients prior to transmission. Given that quantization is an irreversible process, it is crucial to balance achieving maximum EE with maintaining satisfactory learning performance.

As an extreme case of model quantization, the cooperative training of binary neural networks (BNNs) \cite{Hubara_NIPS2016} is particularly appealing. Since the model weights are binarized, BNNs have a small memory footprint and can be trained with high EE. Typically, a set of auxiliary real-valued parameters, which carry a sizable communication overhead, need to be uploaded in each iteration. This issue was addressed by a novel parameter updating scheme with guaranteed convergence in \cite{YangY_JSAC21}. A green quantized FEEL framework that accommodates general DNN precision levels was developed in \cite{Kim_TWC22}, achieving up to 70\% energy savings compared to full-precision training without deteriorating the convergence rate. Specifically, the precision levels were determined based on the energy models for local training and transmission with quantization.

Quantized SGD (QSGD) randomly rounds each element of the model gradients to discrete values before communication in FEEL \cite{AlistarhD_NeurIPS17}. Although the reduced precision of model gradients extends the training time, QSGD decreases the communication energy consumption by $5.7\times$ compared to the standard SGD. FedPAQ \cite{Reisizadeh_AISTATS20} further integrates QSGD with periodic averaging and partial participation. Drawing inspiration from SignSGD \cite{Bernstein_ICML18}, a recent study \cite{JinR_TWC22} proposed exchanging only the signs of model gradients. With the joint optimization of local processing and communication parameters, this approach achieves $18$$\sim$$40\%$ energy consumption reduction and notable accuracy improvements compared to SignSGD and FedAvg~\cite{McMahanB_AISTATS17}. In \cite{Jhunjhunwala_ICASSP21}, an adaptive model update quantization scheme named AdaQuantFL was developed based on a convergence bound of FL. This scheme gradually increases the number of quantization levels throughout the training process. Compared to fixed-level quantization, AdaQuantFL converges with significantly less communication energy consumption and minor accuracy loss. Additionally, universal vector quantization was utilized in \cite{Shlezinger_TSP21} to reduce the gradient quantization error. 

\subsubsection{Sparsification} 
Sparsification is another indispensable approach for energy-efficient FEEL, as it also eliminates unnecessary computations and communications. In \cite{LinY_ICLR18}, a gradient sparsification framework was developed for distributed SGD. EUDs in this framework only transmit gradient elements with a magnitude greater than a threshold, while the remaining elements are accumulated locally and transmitted when they become sufficiently large. This approach achieves hundreds of times reduction in communication energy without compromising accuracy. Notably, the sparsification level is a critical parameter to be determined. Instead of relying on trial-and-error, a fairness-aware gradient sparsification method was developed in \cite{HanP_ICDCS20} via an online learning framework. This method improves the model accuracy by up to $40\%$ for a given training period, even in unfavorable scenarios with statistical data heterogeneity across EUDs. Moreover, a coarse-to-fine compression-based FEEL scheme that combines update quantization and sparsification was proposed in \cite{LiP_TWC23}, achieving $54.7$$\sim$$79.5\%$ energy savings compared to various baselines on the CIFAR-10 classification task.

While model update sparsification reduces communication overhead, model pruning (a.k.a. model sparsification) eliminates unimportant DNN elements, such as neurons and filters, reducing both computation and communication energy consumption in FEEL. By analyzing the impact of model pruning on learning performance, the static pruning ratios and spectrum allocation at EUDs were determined by solving a non-convex problem for FEEL in \cite{LiuS_WCL21}. The authors of \cite{WenD_WCL19} proposed generating random subnets independent of the global model at different dropout rates for local training. These subnets are adaptive to the communication bandwidth and computation capacity at EUDs, which not only reduces energy consumption, but also mitigates the overfitting problem. In \cite{JiangY_TNNLS22}, a model size adaptation method named PruneFL was developed. This method includes initial pruning at a selected EUD and further pruning during the FL process. In each training iteration, the model size can either grow or shrink to maximize the empirical risk reduction divided by the round duration. Compared to FedAvg, PruneFL saves $35.2\%$ of the FLOPs and $71.2\%$ of the training time, and achieves $80\%$ accuracy on the FEMNIST dataset. Additionally, a top-$k$-based gradient compression control algorithm was developed in \cite{LiLiang_INFOCOM21} according to a convergence bound of FEEL. This algorithm consumes $1.5$$\sim$$100\times$ less energy than several baselines for a target accuracy. Furthermore, joint quantization and model pruning was applied in \cite{PrakashP_IoTJ22} to simultaneously improve the time efficiency, EE, and model accuracy of FEEL.

\subsubsection{Knowledge Distillation} Many existing frameworks for FEEL follow the FedAvg algorithm, where all EUDs share the same model architecture, resulting in identical computation and communication overhead. However, this approach does not align well with the heterogeneous resources available at EUDs, necessitating the adoption of different model architectures. Knowledge distillation (KD), which involves transferring knowledge from a teacher network to a student network \cite{Hinton_arXiv15}, plays a crucial role in eliminating such limitations and reducing the on-device energy consumption of FEEL.

A well-known KD-based FL algorithm called FedMD \cite{LiD_NISPwkshop19} addresses this issue by allowing EUDs to have heterogeneous model architectures based on their resource constraints. To facilitate cooperative training, logits computed from a public dataset using local models are uploaded to and aggregated at the MEC server. These aggregated logits are then redistributed to EUDs for local training using TL and KD. Since logits are of low dimension, the communication overhead of FedMD is significantly reduced compared to FedAvg. Empirical results in \cite{LiuL_ICC22} provide further insights into key parameters of KD-based FEEL, such as the size and distribution of the public dataset, as well as the logit aggregation strategy. Another approach, FedKD \cite{wu2021fedkd}, proposes maintaining a teacher and a student model at each EUD that can learn from each other, with only the approximated student model gradients, which have much smaller sizes, being uploaded. Complexity analysis demonstrates that FedKD significantly reduces communication costs with a marginal increase in local computation energy consumption. Besides, a selective knowledge sharing mechanism was developed for KD-based FL in~\cite{ShaoJ_NatCom24}, which identifies accurate and precise knowledge respectively from local and ensemble predictions over the learning process. KD can also be implemented at the MEC server to simplify computation at EUDs. For example, in Mix2FLD \cite{OhS_CL20}, each EUD uploads a set of mixed-up data samples along with the logits. The logit-to-model conversion is performed by running a KD algorithm at the MEC server, after which the updated global model is downloaded by the EUDs. Mix2FLD demonstrates faster convergence and higher model accuracy, implying much lower energy consumption at the EUDs.

\subsection{Resource Management}\label{FL_RM}

In tandem with model and gradient compression, energy-efficient resource management strategies can be adopted in FEEL systems. Although resource management has always been a crucial aspect of communication and computing systems, the unique requirement of joint EE and learning awareness in FEEL poses new challenges. In the following, we will review recent advancements of energy-efficient resource management strategies for FEEL from various perspectives. These perspectives include local training adaptation, joint computation-communication control, device selection, and data selection and offloading.

\subsubsection{Local Training Adaptation} Since most computations in FEEL are disseminated to EUDs, the approach to local training in each iteration has direct impacts on learning performance, as well as time and energy efficiency. Intuitively, better training of local models in each iteration is beneficial of reducing the required number of iterations to achieve a learning performance target \cite{McMahanB_AISTATS17,MillsJ_arXiv23,Dinh_TON21}, but it may increase the on-device processing and computation energy consumption per iteration. However, a smaller number of iterations is advantageous for saving communication energy throughout the training process. Therefore, local training in FEEL should be meticulously adapted to find the optimal balance.

In \cite{WangS_JSAC19}, local training adaptation was first studied for FEEL, where a convergence bound was derived considering arbitrary local training epochs in each global iteration. This bound was utilized to determine the number of local epochs that minimizes the loss function while adhering to resource constraints, such as energy and monetary costs. The batch size of local SGD also plays a significant role, as a larger batch size reduces gradient variance but increases computation overhead \cite{Balles_UAI17}. To conserve battery energy by minimizing device idling, an efficient algorithm was developed in \cite{MaZ_TMC22} to dynamically determine batch sizes and learning rates for heterogeneous EUDs. Additionally, scaling the on-device processors to appropriate speeds can improve the EE of local training. For instance, in \cite{ZhanY_IPDPS20}, the total system cost of FEEL defined as a weighted sum of training time and energy consumption was minimized using an experience-driven central processing unit (CPU)-cycle frequency control algorithm. Compared to using a static CPU-cycle frequency, this approach achieves over $4\%$ energy savings with a substantial reduction in training time.

\subsubsection{Joint Computation-Communication Control}
In addition to local training, the periodic model exchanges between  EUDs and the MEC server require novel transmission schemes to achieve a high EE. As a result, there have been extensive efforts on joint computation-communication control for FEEL \cite{Dinh_TON21, YangZ_TWC21, MoX_JCIN21}. Latency-constrained energy consumption minimization for time division multiple access (TDMA) and frequency division multiple access (FDMA)-based FEEL systems were investigated in \cite{Dinh_TON21} and \cite{YangZ_TWC21} respectively, where the required number of training iterations for a target model accuracy is related to the local training quality. In \cite{YangZ_TWC21}, a low-complexity iterative algorithm was developed to jointly optimize the local training (i.e., CPU-cycle frequencies and local training accuracy) and model uploading (i.e., transmit power and bandwidth allocation) schemes. Numerical results demonstrated its superiority in reducing energy consumption compared to baseline methods with equal bandwidth allocation and sub-optimal local training policy. Additionally, assuming a given number of training iterations, a similar investigation for non-orthogonal multiple access (NOMA)-enabled FEEL systems was conducted in \cite{MoX_JCIN21}.

Recent studies have also incorporated state-of-the-art computing architectures \cite{ZengQ_TWC21} and air interface technologies \cite{ZhangT_TGCN22}. In \cite{ZengQ_TWC21}, each EUD was assumed to be equipped with a CPU-GPU heterogeneous computing platform. Consequently, the joint optimization of communication resource management, CPU-GPU workload partitioning, and processor frequency scaling was performed to minimize total energy consumption. Experimental results demonstrated the effectiveness of joint resource management in improving EE compared to separate management of heterogeneous resources. Attention was also given to the intelligent reflecting surface (IRS) technology, which utilizes passive elements such as diodes and phase shifters to customize wireless environments in an energy-efficient manner. In \cite{ZhangT_TGCN22}, an IRS was proposed to be deployed in FEEL systems to provide favorable wireless channels, with the phase shifts at the IRS and the local CPU speed being judiciously designed. Simulation results showcased speedy convergence and significant energy savings, especially with a large-scale IRS.

\subsubsection{Device Selection}

FEEL systems often face limitations in accommodating a large number of EUDs due to limited wireless bandwidth \cite{Nishio_ICC19}. Therefore, selecting an appropriate subset of EUDs to participate in the training process is crucial for both learning performance and device energy consumption. The simplest method for device selection in FEEL is to assign a deadline for each training iteration, allowing exclusion of straggling EUDs. Building upon this method, a hierarchical online pace control framework for FEEL was proposed in \cite{LiL_RTSS19}, which determines the iteration deadline based on predicted connectivity and energy budget. In conjunction with the CPU speed control, it achieves a $32.8\%$ reduction in device training energy consumption without any loss in accuracy.

In \cite{HuY_WCSP20}, a device selection problem was tackled by a hybrid branch-and-bound algorithm in each training iteration to minimize the energy consumption of participating EUDs in a TDMA-based FEEL system. Additionally, the trade-off between maximizing the number of selected EUDs and minimizing their total energy consumption was investigated in \cite{YuL_IoTJ22}, leading to the development of an energy- and latency-aware device selection algorithm. Both \cite{HuY_WCSP20} and \cite{YuL_IoTJ22} demonstrate the effectiveness of device selection in achieving energy-efficient FEEL. To ensure unbiased model aggregation with a subset of selected devices, probabilistic device scheduling frameworks were developed in~\cite{RenJ_TWC20,SunY_TWC23}.

The aforementioned studies focus on optimizing device selection independently in each training iteration. However, all decisions made throughout the training process can have a significant impact on the learning outcome. This motivates the design of device selection strategies from a long-term perspective \cite{XuJ_TWC2021, GuoK_JSTSP22, BattiloroC_TGCN22,KimYG_MICRO21}. In particular, recognizing that the number of participating EUDs in later training rounds has a more profound impact on learning performance, device selection and bandwidth allocation were jointly optimized in \cite{XuJ_TWC2021} using Lyapunov optimization, subject to energy constraints at each EUD. Experimental results demonstrate the benefits of this approach over methods that myopically minimize energy consumption in each round. This work was further extended in \cite{GuoK_JSTSP22} to address the device heterogeneity issue by exploiting sequential transmission of local model updates, where early available local model updates can be uploaded to the MEC server using more bandwidth. Unlike \cite{XuJ_TWC2021} and \cite{GuoK_JSTSP22}, which use the average number of participating EUDs as a surrogate for learning performance, \cite{BattiloroC_TGCN22} developed an online data-driven method to estimate learning performance using a validation dataset. Based on this estimation, the selected devices and resource management were jointly optimized to minimize the system energy consumption while respecting latency and accuracy requirements. In~\cite{KimYG_MICRO21}, a lightweight AutoFL framework was developed based on reinforcement learning to select the best EUDs in each training round, achieving $5.2\times$ EE improvement compared with random device selection. 
 
\subsubsection{Data Selection} 

The efficiency of model training can be improved by identifying important data samples in large datasets \cite{Katharopoulos_ICML18}. This presents two opportunities for energy savings in FEEL. First, removing irrelevant and potentially adversarial data samples accelerates model convergence, reducing both the computation energy for local training and the communication energy for model exchanging. Second, using only the most informative data samples in local training reduces the computation overhead in each training iteration. However, evaluating the utility of data samples is challenging in FEEL, as data are kept locally at EUDs.

In \cite{XiaoY_WCSP20}, an FEEL framework that allows EUDs to determine suitable amounts of data samples for local training was developed. Based on a mapping function that quantifies impacts of the dataset and local batch size on training performance, an energy cost minimization problem was formulated and solved via the distributed alternative direction method of multipliers (ADMM). Sample-level importance was evaluated in \cite{HeY_TVT20, Albaseer_TNSE21, LiA_INFOCOM21} to design energy-efficient FEEL mechanisms. In \cite{HeY_TVT20}, the gradient norm, which contributes positively to loss decay, was adopted as a data importance indicator. Joint data selection and communication resource allocation was performed to maximize learning efficiency in each training iteration. The authors of \cite{Albaseer_TNSE21} proposed measuring sample-level importance using the maximum categorical probability computed via the updated local model in the first training iteration, and less important samples can hence be excluded from subsequent local training. By optimizing communication and computational resource allocation, the energy consumption can be reduced by over $70\%$ on the MNIST and CIFAR-10 datasets. Furthermore, a systematic investigation on sample-level importance for FEEL was conducted in \cite{LiA_INFOCOM21}, considering both statistical homogeneity across EUDs and content diversity at each EUD. To reduce the data selection overhead, EUDs with less relevant data are eliminated before training, while other EUDs and their data samples are dynamically selected during training. Numerical evaluations demonstrate that such a hierarchical data selection mechanism not only significantly saves computation and communication costs but also notably improves model accuracy and convergence rate.
 
\subsubsection{Data Offloading}

In FEEL applications where EUDs and MEC servers are managed by the same authority, the data privacy concerns can be much relieved, allowing data offloading to reduce on-device training overhead. The MEC server, with its more capable computational resources, is well-suited to collect data offloaded from EUDs (i.e., vertical data offloading) and perform model training using the collected data \cite{YangW_WCSP20,JiZ_TVT21,Wutyee_ACCESS21}. The data offloading problem in FEEL was formulated as an MDP problem in \cite{YangW_WCSP20}, where each EUD determines whether to upload a data sample to the MEC server or use it for local training based on channel quality and battery energy state. In \cite{JiZ_TVT21} and \cite{Wutyee_ACCESS21}, it was proposed to offload data from EUDs to the MEC server prior to model training. The offloaded data volume and resource management were jointly optimized to minimize training latency while satisfying energy consumption constraints. In addition to vertical data offloading to the MEC server, there is also growing interest in horizontal data offloading to nearby EUDs by leveraging the device-to-device (D2D) communication links \cite{CaiX_CJIoT19,WangS_INFOCOM21,Shullary_MWSCAS22}. However, for applications with stringent data privacy requirements, offloading raw data may not be admissible. In such cases, the framework of coded FL can be adopted, where privacy-preserving coding schemes are applied to the local dataset before data offloading \cite{Prakash_JSAC21,SunY_ISIT22,ShaoJ_NeurIPS22}.

\subsection{EE-oriented Designs for Other Cooperative Edge Learning Frameworks}\label{otherCoLearnFrameworks}
In this subsection, we introduce two alternative cooperative edge learning frameworks, including DEEL and H-FEEL, and elaborate several of their EE-oriented designs. 

\subsubsection{DEEL} Contrary to FEEL systems that employ an MEC server as the PS, EUDs in DEEL systems share local model updates with their one-hop neighboring EUDs in each training iteration, which is more resilient to node failures. Model aggregation is performed at individual EUDs. Typically, an EUD generates a new local model by mixing its own model and those from neighboring EUDs with appropriate weighting factors~\cite{LiuW_TSIPN22}. Despite the training algorithm of DEEL is different from that of FEEL, most design principles of energy-efficient FEEL are applicable for achieving energy-efficient DEEL. For example, a DEEL mechanism where EUDs exchange quantized model parameters was developed in~\cite{MaT_IoTJ23}. Experimental results on a beat signal dataset show that the model accuracy just slightly degrades even with 1-bit quantization. Since communications in DEEL take place between neighboring EUDs, similar as FEEL, wireless links in unfavorable conditions may prolong the learning latency and result in waste of transmission energy. Therefore, link selection strategies should be optimized apart from communication and computational resource management~\cite{ZhangX_ICASSP24,LiuS_TWC23}. To reduce both the communication and computation energy consumption, joint model pruning and network topology construction for DEEL was investigated in~\cite{JiangZ_TPDS23}, where each EUD trains a differently pruned sub-model and the network topology is adapted according to data heterogeneity and link speeds.

\begin{figure*}[t!]
	\includegraphics[width=18cm]{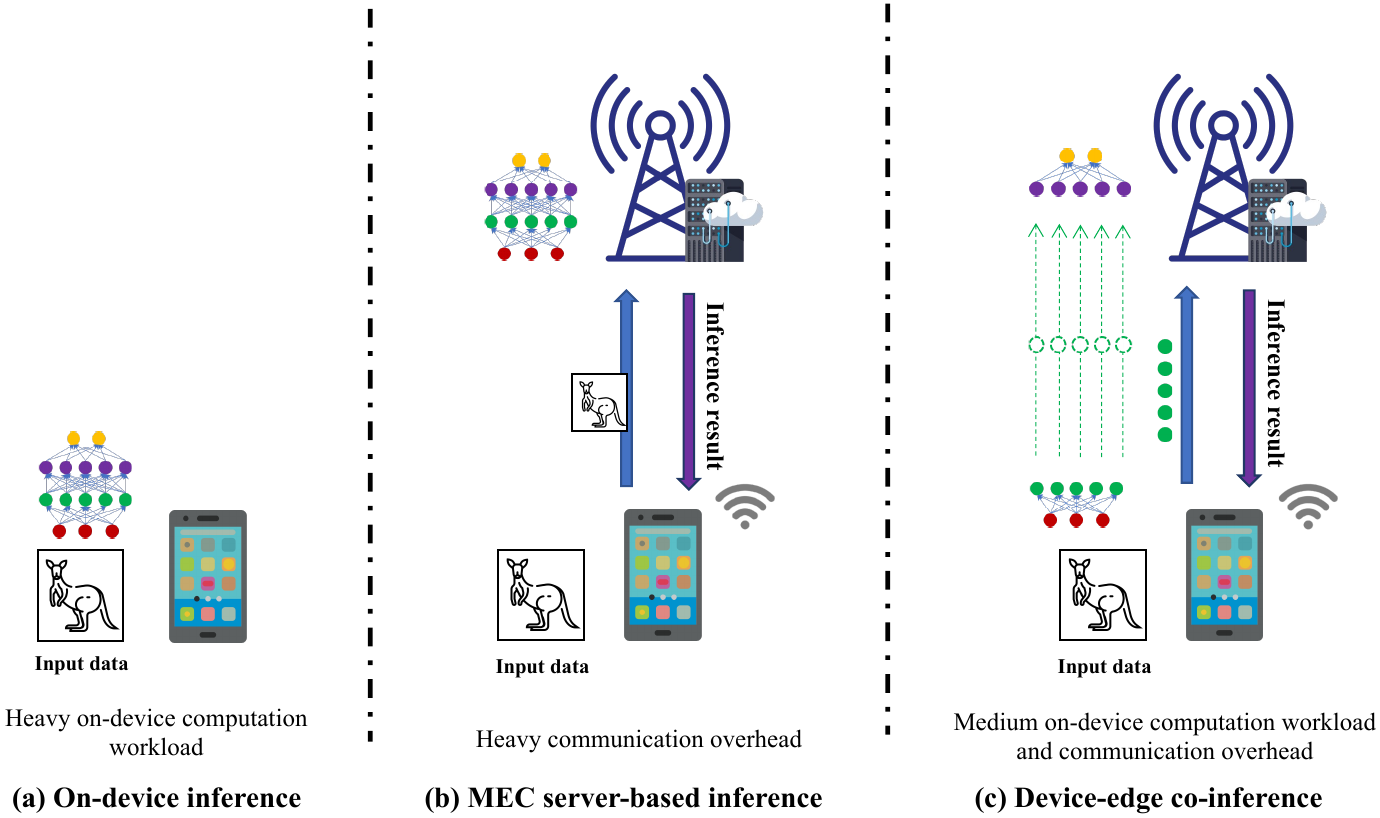}
	\caption{Three typical edge inference paradigms, including: (a) on-device inference; (b) MEC server-based inference; and (c) device-edge cooperative inference (``co-inference'').}
	\label{fig:inferenceParadigms}
\end{figure*}

\subsubsection{H-FEEL} In H-FEEL systems, EUDs are partitioned into multiple clusters and EUDs in a cluster are associated with an MEC server. The MEC servers are coordinated by a cloud server~\cite{LiuL_TWC23}. Hence, model aggregation of H-FEEL consists of two levels, namely the edge- and cloud-level aggregation. Such a cooperative edge model training paradigm extends the network coverage to engage many EUDs, thus improving the learning performance. As the operations within each cluster of H-FEEL are largely the same as those of FEEL, it can be expected that the design approaches discussed in Section \ref{FL_Compression} and \ref{FL_RM} remain effective~\cite{SunH_ICASSP24,LiuX_TWC24}. For instance, in~\cite{SunH_ICASSP24}, convergence performance of H-FEEL with quantized gradient transmission and imperfect CSI was analyzed, which unveils the relationship between quantization accuracy and transmission error. Adaptive model pruning was adopted for H-FEEL in~\cite{LiuX_TWC24} to minimize the communication and computation overhead. To determine the prune ratios of EUDs and bandwidth allocation in each cluster, a minimization problem on a convergence bound was optimally solved. Compared to FEEL, device-edge association schemes in H-FEEL have profound impact on the learning performance and cost, which thus become a new frontier to be studied~\cite{LuoS_TWC20,Mhaisen_TNSE22}. Specifically, communication-computation resource allocation and device-edge association was jointly optimized for H-FEEL to minimize the global energy and latency cost. Noticing that the imbalanced data distribution across clients is a main cause of slow convergence and degraded accuracy, device-edge association was obtained by minimizing the divergence between the federated and hypothetical central models in~\cite{Mhaisen_TNSE22}. As such, the communication and computation energy consumption can be reduced with less required training rounds.

\subsection{Takeaways}
This section introduces energy-efficient cooperative model training for edge AI, with a specific focus on FEEL as an illustrative framework. Green FEEL systems are typically realized through two techniques: compression and resource management. Compression techniques aim to reduce the amount of data communicated between the MEC server and EUDs, and the number of FLOPs required for local model training. These objectives can be achieved through methods such as quantization, sparsification, and knowledge distillation. While various approaches have been summarized in this regard, it is important to delve deeper into compression methodologies that can adapt to the inherent heterogeneity at the device, model, and data levels. This nuanced investigation is essential to align with practical usage scenarios. Moreover, privacy-enhancing techniques such as local differential privacy (LDP), can be jointly applied with compression~\cite{LangN_TSP23}. Resource management techniques focus on efficiently utilizing the limited communication and computational resources during the training process of FEEL. This involves optimizing training parameters, such as the number of local training epochs per iteration, batch sizes, and CPU-cycle frequencies, to maximize the EE while maintaining good learning performance. Furthermore, judiciously selecting the data and devices that participate in the training procedures contributes to enhancing the EE. A highly important aspect in this area is the robust allocation of resources to handle device failures that might occasionally happen due to energy shortage~\cite{SunY_TMC23} within the FEEL paradigm. Besides, EE analysis for large-scale FEEL systems is instructive for real-world deployments, for which, stochastic geometry theory can provide a powerful framework~\cite{LinZhen_TWC22}. These practical concerns warrant comprehensive investigations. Moreover, although the EE-oriented design approaches of FEEL can be leveraged for other cooperative edge learning frameworks, attentions on their unique operations should be given.

\section{Energy-Efficient Edge Inference: Paradigms and Algorithms}\label{EdgeInferenceSec}

\begin{figure*}[ht!]
	\includegraphics[width=18cm]{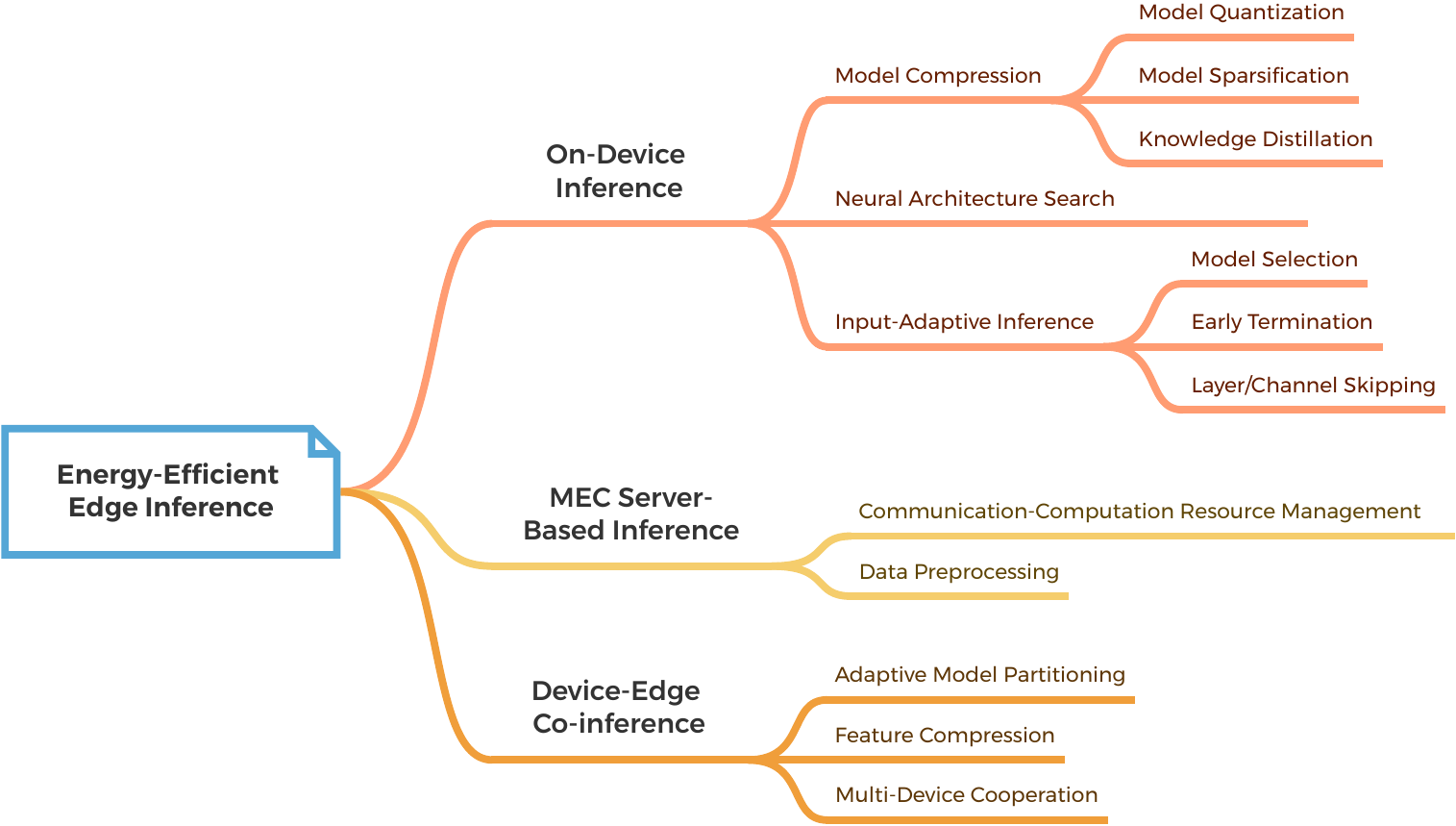}
	\caption{Energy-efficient design approaches for on-device inference, MEC server-based inference, and device-edge co-inference.}
	\label{fig:inferenceTechniques}
\end{figure*}

Despite model training is extremely energy-demanding, most energy of DNN processing is actually consumed at the inference stage that repeats for every new observation. This section introduces three representative edge inference paradigms~\cite{ZhouZ_PIEEE20} and the corresponding EE-oriented design methodologies. Specifically, we start from the primitive paradigms of on-device and MEC server-based inference, which respectively present high on-device computation and communication overhead. To eliminate these bottlenecks, the emerging paradigm of device-edge cooperative inference (``co-inference'') is then investigated. Fig.~\ref{fig:inferenceParadigms} illustrates the three typical edge inference paradigms and Fig.~\ref{fig:inferenceTechniques} summarizes the corresponding energy-efficient design approaches.

\subsection{On-device Inference}\label{sec:on-device_inference}

On-device inference makes predictions for local observations by deploying pre-trained DNN models at EUDs. Such a paradigm fits well for EUDs with a reasonable amount of computational resources and privacy-sensitive applications that prohibit data offloading. However, processing DNNs fully on-device is very challenging due to the computational resource, battery energy, and thermal constraints~\cite{LeeJ_abs-1907-01989}. Previous efforts mainly target at improving the real-time inference performance~\cite{Howard_MobileNet17,NiuW_IJCAI20,GuoP_NSDI21} without scrutinizing the energy footprint. Next, we present three categories of techniques that can effectively trim the on-device inference energy consumption.

\emph{1) Model Compression:} 
Energy consumption of on-device inference mainly comes from accessing the off-chip memory for model parameters and performing forward passes on inference data. Similar to cooperative edge learning, shrinking the sizes of DNN models saves both the computation and memory access energy. Also, reduced-precision arithmetic is beneficial for low-power processing~\cite{Hashemi_DATE17}. Hence, model compression techniques are promising to improve the EE of on-device inference, among which, model quantization, model sparsification, and knowledge distillation, will be discussed. 

\textbf{Model Quantization:} To reduce the on-device computation energy consumption, it is desired to quantize DNNs to the lowest possible bit-widths while retaining their inference accuracy~\cite{Gholami_abs-2103-13630}, which is challenging via handcrafted quantization methods. Hence, a quantized DNN and its quantizer were jointly trained in~\cite{ZhangD_ECCV18}, showing remarkable accuracy with just 4-bit representations of weights and activations. To the extreme case, BNNs~\cite{Hubara_NIPS2016} as mentioned in Section~\ref{FL_Compression}, enable ultra low-power inference as most floating-point MACs can be replaced with low-cost bit-wise operations. However, dedicated training methods are required. Post-training quantization was investigated in~\cite{Banner_NIPS19}, which avoids model retraining but normally with lower accuracy. Besides, an energy-constrained mixed precision quantization framework named HAQ was developed based on deep reinforcement learning (DRL) in~\cite{WangK_CVPR19}, where a hardware simulator was employed to inform the DRL agent on the energy consumption. Compared with an 8-bit MobileNet-V1 \cite{Howard_MobileNet17}, $50\%$ of the inference energy can be saved with $<0.5\%$ of top-1 accuracy drop on ImageNet. 

\textbf{Model Sparsification:}
As discussed in Section \ref{FL_Compression}, model sparsification removes individual weights or entire filters in DNN models, which can be jointly applied with model quantization methods~\cite{LiangT_NeuCom21}. It reduces the required number of arithmetic operations and memory access for on-device inference, thus saving computation energy. In~\cite{YangT_CVPR17}, an energy-aware layer-by-layer pruning scheme was proposed for CNNs, which uses the energy consumption parameters extrapolated from actual hardware measurements to guide the pruning process and achieves $3.7\times$ ($1.6\times$) energy consumption reduction for AlexNet (GoogleNet) with $<1\%$ of top-5 accuracy loss on ImageNet. Assuming a systolic-array-based hardware architecture~\cite{Kung_Computer82} and a three-level memory hierarchy, energy models for DNN computation and data access were developed in~\cite{YangH_ICLR19}, which depend on the input and weight sparsity of each DNN layer. Accordingly, DNNs were trained by alternatively optimizing the model parameters and input masks subject to the energy budget and input sparsity constraints. This solution needs only $26$$\sim$$68\%$ energy budget of the dense models with $0.5$$\sim$$1.0\%$ top-5 accuracy drop on classifying ImageNet. Observing that the model sparsification procedures incur extra overhead, a novel method that enables simultaneous pruning and training was developed in~\cite{ZhiX_arXiv23}, which can remove $\sim$$50\%$ of the FLOPs with minor accuracy reduction via a lightweight binary gating module and a polarizing regularizer.

\textbf{Knowledge Distillation:} KD is able to generate compressed DNN models in customized architectures according to energy budgets of EUDs. The use of different knowledge on DNN models results in different KD techniques, including response-based, feature-based, and relation-based KD, as summarized in~\cite{GouJ_IJCV21}. Specifically, response-based KD utilizes logits of one or more teacher DNNs to train a student model, which shows promising results on image classification and automatic speech recognition~\cite{Hinton_arXiv15}. To leverage the intermediate representations learned by a teacher model, a feature-based KD method named FitNets was proposed in~\cite{Romero_ICLR15}, which reduces $10\times$ of parameters compared to a larger teacher DNN while with improved inference accuracy. Moreover, considering knowledge can be better represented in the embedding space, relation-based KD exploits the mutual relations of data samples in the teacher model's outputs~\cite{ParkW_CVPR19}. It can be applied with response-based or feature-based KD, while the combination of relation-based and response-based KD achieves higher accuracy on classifying CIFAR-100 and Tiny ImageNet as observed in \cite{ParkW_CVPR19}. Recently, progressive KD was proposed in~\cite{Dennis_arXiv23} that distills a large teacher model into a collection of small student models, which can be selectively activated in runtime for ensemble inference to balance the energy cost and inference accuracy.

\emph{2) Neural Architecture Search:} Model compression techniques are bundled with existing DNN model architectures. Typically, DNN models are handcrafted by compositing different basic operations. For example, MobileNet~\cite{Howard_MobileNet17}, a streamlined architecture for mobile vision, adopts the depth-wise separable convolutions. However, handcrafted approaches are too laborious to enumerate many possibilities. Neural architecture search (NAS), which leverages optimization methods, such as reinforcement learning and evolutionary algorithms~\cite{RenP_ACMCOMPSurvey2021}, efficiently traverses the search space for target performance metrics. Despite primitive NAS frameworks were concerned more on model accuracy and computation complexity~\cite{RenP_ACMCOMPSurvey2021}, they have been later generalized with energy awareness~\cite{WangD_EMC2_19,Benmeziane_arXiv210109336}. To incorporate the energy consumption, a multi-object NAS framework was developed in~\cite{Hsu_abs-1806-10332}, where the energy cost of a candidate model is measured via the NVIDIA profiling tool ``nvprof''. For other platforms, energy consumption can be estimated via alternative profiling methods (e.g., by measuring the current and voltage of RPi)~\cite{Srinivas_abs-1906_07214} or prediction models (e.g., random forest)~\cite{Speckhard_NCA23}. More discussions on platform-aware NAS are deferred to Section \ref{Sec:Future:HW-SW}. Besides, mixed-precision NAS, which jointly considers quantization and NAS, was investigated in~\cite{GongC_ICCAD19}. It reduces the inference energy consumption on the CIFAR-100 classification task by $20.9\%$ and $76.2\%$, compared with HAQ~\cite{WangK_CVPR19} and MobileNetV2~\cite{Sandler_MobileNetV2}, respectively, while maintaining similar inference accuracy.

\begin{figure*}[ht!]
	\begin{subfigure}{.5\textwidth}
		\centering
		\includegraphics[width=1\linewidth]{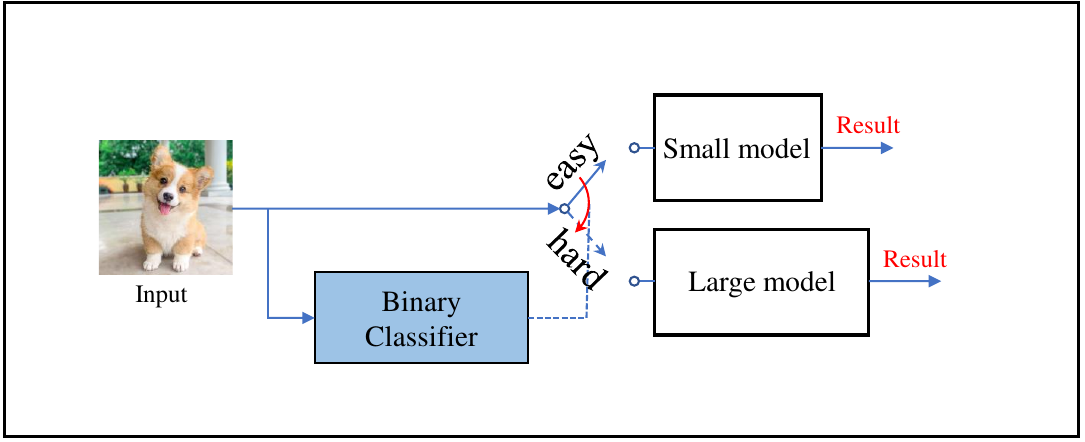}  
		\caption{Model selection.}
		\label{fig:DDNN_a}
	\end{subfigure}
	\begin{subfigure}{.5\textwidth}
		\centering
		\includegraphics[width=1\linewidth]{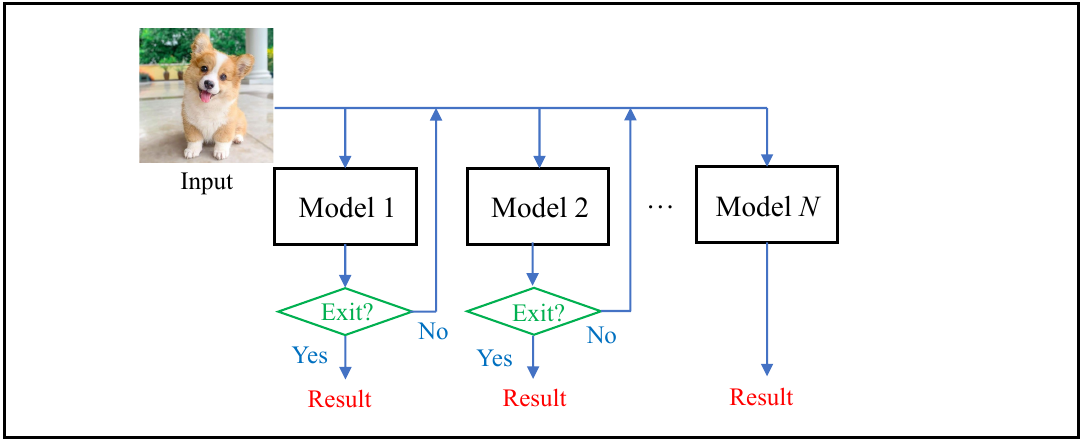}  
		\caption{Cascaded inference.}
		\label{fig:DDNN_b}
	\end{subfigure}
	\begin{subfigure}{.5\textwidth}
		\centering
		\includegraphics[width=1\linewidth]{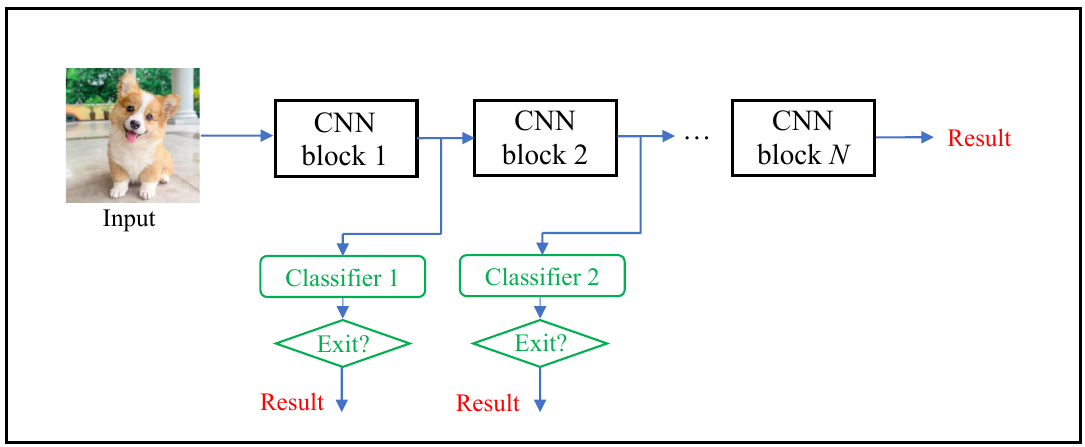}  
		\caption{Branchy network.}
		\label{fig:DDNN_c}
	\end{subfigure}
	\begin{subfigure}{.5\textwidth}
		\centering
		\includegraphics[width=1\linewidth]{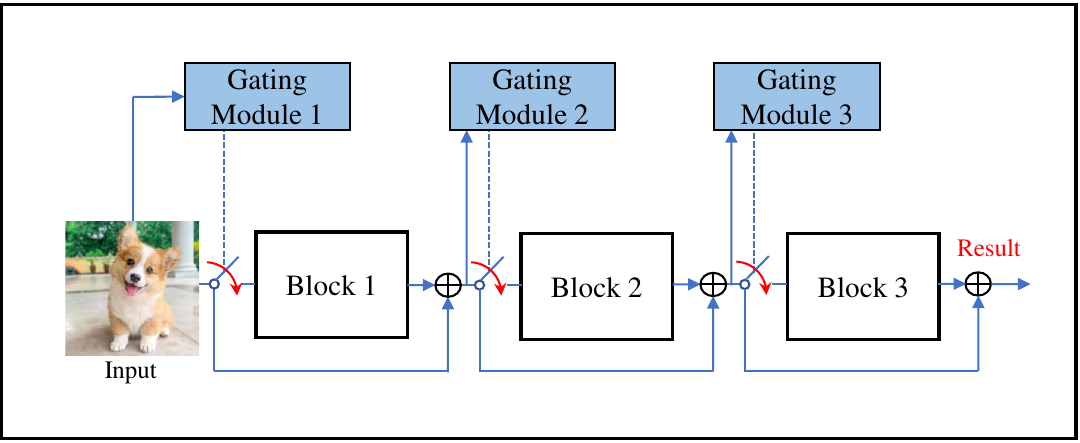}  
		\caption{Layer skipping.}
		\label{fig:DDNN_d}
	\end{subfigure}
	\caption{Input-adaptive deep neural network architectures for on-device inference, including (a) model selection~\cite{Maleki_TDAES21}, (b) cascaded inference~\cite{ZhouYH_IJCAI18}, (c) branchy network~\cite{Teerapittayanon_ICPR16}, and (d) layer skipping~\cite{WangX_ECCV18}.}
	\label{fig:DDNN}
\end{figure*}

\emph{3) Input-Adaptive Inference:} The computation graphs of conventional DNN models are fixed. Such one-for-all designs ignore the uneven difficulty of inference data, which are suboptimal in terms of EE. Hence, various input-adaptive inference strategies~\cite{Panda_DATE16,Laskaridis_EMDL21,HanY_TPAMI21} with examples depicted in Fig.~\ref{fig:DDNN}, including model selection, early termination, and layer/channel skipping, have been proposed.

\textbf{Model Selection:} As shown in Fig.~\ref{fig:DDNN_a}, model selection for input-adaptive inference deploys two DNN models with different computation overhead and inference accuracy. To decide which DNN to use, a lightweight binary classifier can be adopted to evaluate the input difficulty. For image classification, the two DNNs can be different pruned versions of a base DNN model, and a relabelled dataset can be obtained to train the binary classifier~\cite{Maleki_TDAES21}. MobiSR~\cite{LeeRoyson_MobiCom19} follows a similar idea for image super resolution, where the upscaling difficulty evaluation unit was designed with a total variation metric. It is also possible to select from more than two DNN models~\cite{MarcoSanz_TECS2020}. 

\textbf{Early Termination:} Early termination employs multiple intermediate exits to decide whether to terminate the inference process at an early stage. Fig.~\ref{fig:DDNN_b} and Fig.~\ref{fig:DDNN_c} show two early termination DNN architectures, which are respectively called \emph{cascaded inference}~\cite{ZhouYH_IJCAI18} and \emph{branchy network}~\cite{Teerapittayanon_ICPR16}. In cascaded inference, multiple fully functional DNNs with different computation overhead and inference accuracy are concatenated, where each input data is first processed by the small models, followed by the larger ones. The inference result of each DNN is used to compute a metric (e.g., confidence score) as an indicator for early termination. For branchy networks, early exits, which are lightweight classifiers, are inserted along the depth of a base DNN model. Since the early exits incur extra computation overhead and may even degrade the base model accuracy~\cite{HanY_TPAMI21}, the number and insertion locations are crucial to the EE. To address this issue, FlexDNN was developed to find an optimal insertion plan in~\cite{FangBiyi_SEC20} .

\textbf{Layer/Channel Skipping:} Layer/channel skipping allows more flexible adaptation by determining which network layers/channels in a network layer should be executed according to the inference data. It can be realized by training a gating module for each candidate layer/channel~\cite{HanY_TPAMI21}. For example, the layer skipping architecture SkipNet \cite{WangX_ECCV18} shown in Fig.~\ref{fig:DDNN_d} suits well for ResNets due to its robustness to layer dropping. Alternatively, BlockDrop~\cite{WuZuxuan_CVPR18} trains a policy network that outputs probabilities of binary dropping decisions for all layers under a single-step MDP formulation, which addresses both the prediction correctness and number of utilized residual blocks. To avoid overwhelming the energy savings brought by layer skipping, the gating modules/policy network should have negligible computation cost compared to the base model. Hence, it was further suggested in~\cite{WuZuxuan_CVPR18} to use a fraction of the depth in the base ResNet as the policy network architecture. Channel skipping such as CGNet~\cite{HuaWeizhe_NeurSIP19} can be viewed as a fine-grained variant of layer skipping since the skipping policies for individual channels need to be learned. Moreover, a multi-grained approach named IADI was proposed in~\cite{WangYue_JSTSP20}, which supports simultaneous layer and channel skipping.
 
Table~\ref{table:DDNN_case_studies} summarizes the computation and energy savings, as well as accuracy gain of selected input-adaptive on-device inference methods. Interested readers may refer to~\cite{HanY_TPAMI21} for more discussions on the taxonomy and architectures of input-adaptive inference.

\subsection{MEC Server-based Inference}

In many scenarios, computational resources on EUDs are too limited to handle complex inference tasks, and the MEC server provides a resource pool for EUDs to offload their heavy computations. In MEC server-based inference, DNN models are fully deployed at the MEC server and the energy consumption for transmitting the inference data dominates. To improve the EE, uplink transmission schemes need appropriate designs. Also, optimized inference task offloading policies are critical to exploit all available resources~\cite{Pavel_COMST17}. Besides, the DNN fault-tolerance~\cite{HUITZIL_ACCESS17} presents a new dimension of trading the inference accuracy for better EE via data preprocessing. Following this thread, we next introduce the EE-oriented design methodologies for MEC server-based inference.

\begin{table*}[ht!]
	\caption{Case Studies on Input-adaptive Inference} 
	\centering 
	\begin{tabular}{c c c c c c c} 
		\hline\hline 
		Name & Type & Task & Baseline (Dataset) & Acc. Gain & Comput. Saving & Energy Saving \\ [0.5ex] 
		\hline 
		\thead{ICNN~\cite{Maleki_TDAES21}} & \thead{Model \\ selection}  & \thead{Image \\ classification}  & \thead{ResNet-50 (CIFAR-100) \\  MobileNetV2 (CIFAR-100)}  & \thead{$-11.11\%$\\ $-13.95$\%}  & \thead{$22\%$ ($64\%$){$^{\text{a}}$}\\N.A.}  & \thead{$69\%${$^{\text{a}}$}\\ $66\%$} \\ 
		MobiSR~\cite{LeeRoyson_MobiCom19} & \thead{Model \\ selection} &  \thead{Super \\ resolution}  & RCAN (Urban100) & $-0.2$$\sim$$0$ dB$^\text{b}$ & $1$$\sim$$2\times$$^\text{b}$  & \thead{N.A.\\N.A.} \\
		\thead{Cascaded\\ 
		inference~\cite{ZhouYH_IJCAI18}} & \thead{Early\\termination} &  \thead{Image \\ classification}   & \thead{ResNet-32 (CIFAR-100)\\ResNet-110 (CIFAR-100)} & \thead{$+3.8\%$ \\$+0.87\%$} & \thead{$-75.05\%$$^\text{c}$ \\$51.02\%$} & \thead{N.A.\\N.A.} \\
		BranchyNet~\cite{Teerapittayanon_ICPR16} & \thead{Early\\termination}   & \thead{Image \\ classification} & \thead{AlexNet (CIFAR-10)\\ResNet-110 (CIFAR-10)} & \thead{$+0.81\%$ \\$-1.53\%$} & \thead{$33.9\%\ (58.7\%)$$^\text{d}$ \\ $46.4\%\ (47.5\%)$} & \thead{N.A.\\N.A.} \\
		FlexDNN~\cite{FangBiyi_SEC20} & \thead{Early\\termination}   & \thead{Activity \\ recognition} & \thead{Inception-V3 (UCF-15)\\VGG-16 (UCF-15)} & \thead{$\sim$$0\%$ \\$\sim$$0\%$} & \thead{$3.9\times$$\ (91.8\%)^\text{e}$ \\ $2\times$$\ (79.2\%)$} & \thead{$73.0\%^\text{e}$ \\ $50.9\%$ } \\
		SkipNet~\cite{WangX_ECCV18}  & \thead{Layer/channel\\ skipping}  &  \thead{Image \\ classification} & \thead{ResNet-110 (CIFAR-10)\\ResNet-110 (CIFAR-100)} &\thead{$-0.5$$\sim$$0.5\%$ \\$-0.5$$\sim$$0.5\%$} & \thead{$23\%$$\sim$$50\%^\text{c}$ \\$17\%$$\sim$$37\%$} & \thead{N.A. \\ N.A.}  \\
		BlockDrop~\cite{WuZuxuan_CVPR18}  & \thead{Layer/channel\\ skipping}  &  \thead{Image \\ classification} & \thead{ResNet-110 (CIFAR-10)\\ResNet-101 (ImageNet)} &\thead{$-1.3\%$$\sim$$+0.4\%$ \\$-1.2\%$$\sim$$+0.4\%$} & \thead{$64.3\%$$\sim$$76.7\%^\text{c}$ \\$5.7\%$$\sim$$36.9\%$} & \thead{N.A. \\ N.A.}  \\
		CGNet~\cite{HuaWeizhe_NeurSIP19}  & \thead{Layer/channel\\ skipping}  &  \thead{Image \\ classification} & \thead{ResNet-18 (CIFAR-10)\\VGG-16 (CIFAR-10)} &\thead{$-0.56$$\sim$$-0.04\%$ \\$-0.39$$\sim$$+0.08$\%} & \thead{$5.49$$\times$$\sim$$7.95$$\times$$^\text{c}$ \\$3.41$$\times$$\sim$$5.10$$\times$} & \thead{N.A. \\ N.A.}  \\
		IADI~\cite{WangYue_JSTSP20} &  \thead{Layer/channel\\ skipping} & \thead{Image \\ classification} & ResNet-74 (CIFAR-10) & $-1.9$$\sim$$-0.15\%$     & $40$$\sim$$80\%^\text{c}$  & $36$$\sim$$78.2\%${$^{\text{f}}$} \\ [1ex] 
		\hline 
	\end{tabular}
	
	\begin{tablenotes}
		\item a: The computation saving is measured by the speedup on a CPU (Eyeriss) platform. When with restrictions on the accuracy loss, the amount of energy saving generally reduces~\cite{Maleki_TDAES21}.
		\item b: The accuracy is measured by the peak signal-to-noise ratio (PSNR), and the computation saving is measured by the speedup when MobiSR is implemented on the Qualcomm SDM845 platform~\cite{LeeRoyson_MobiCom19}.
		\item c: The computation saving is measured by FLOP reduction.
		\item d: The computation saving is measured by the speedup on a platform with a 3.0 GHz CPU (NVIDIA GeForce
		GTX TITAN X (Maxwell) GPU)~\cite{Teerapittayanon_ICPR16}.
		\item e: The computation saving is measured by the speedup (frame drop rate reduction) on a Samsung Galaxy S8 smartphone~\cite{FangBiyi_SEC20}. The energy saving is calculated for processing each frame.
		\item f: The energy saving is measured on a Digilent ZedBoard Zynq-7000
		development board~\cite{WangYue_JSTSP20}.
	\end{tablenotes}
	\label{table:DDNN_case_studies}
\end{table*}

\emph{1) Communication-Computational Resource Management:} Given the inference data to be offloaded, green communication techniques can be adopted to design energy-efficient uplink transmission schemes~\cite{Hasan_COMST11,ZhangSQ_COMST17}. For EUDs that are able to compute some inference tasks, inference task offloading policies should be jointly designed. Meanwhile, efficient computational resource management at the MEC server is crucial to accommodate more inference requests to achieve maximum on-device energy savings. Notably, these considerations align with those in conventional MEC systems~\cite{MaoY_COMST17}.

Nevertheless, task offloading in MEC systems mainly aims at reducing the execution latency and energy consumption, while inference accuracy becomes a new performance metric in MEC server-based inference~\cite{ChenY_MCOM11,SunWen_IEEENetw19}. To narrow the research gap, an inference task distribution scheme among EUDs and the MEC server was presented in~\cite{YangBo_IoTJ20}, which considers both inference accuracy regarding data quality and computation demand of DNN models using statistical information estimated on a validation dataset. In addition to computation overhead, the placement overhead of caching DNN models at the MEC server is also non-negligible. To maximize the average inference accuracy while meeting the latency requirements, a multi-dimensional optimization problem was formulated and efficiently solved by DRL in~\cite{ZhangWT_TVT21}, which determines the DNN placement and communication-computational resource allocation. A recent investigation in \cite{MaH_IoTJ23} further considers the purchase of carbon emission rights when designing inference task offloading policies. 

\emph{2) Data Preprocessing:} Many DNN models accept input data with different configurations. For example, the real-time object detector YOLO~\cite{Redmon_YOLO15,Redmon_Yolov2} can process video frames in different resolutions. Some data configurations, such as frame rate, resolution, and encoding bitrate, are tightly coupled with the communication overhead~\cite{RanXukan_INFOCOM18}. Thus, preprocessing the raw inference data to proper configurations has great potential for improving the EE of MEC server-based inference~\cite{XuD_CL19}.

The impact of data preprocessing on inference accuracy can be estimated via offline profiling. For a bearing vibration signal dataset, the facility fault diagnosis accuracy of AlexNet was found to increase with the sampling rate~\cite{WuWen_TII21}. Since a higher sampling rate leads to an increased data volume, it needs to be dynamically configured according to the wireless channel condition. Lossy compression was exploited to reduce the communication overhead on the MNIST and CIFAR-10 classification tasks in~\cite{HuangXiufeng_IOTJ20}, where the inference accuracy was modeled as a function of compression ratio via a lookup table. In~\cite{RanXukan_INFOCOM18}, the frame quality for video analytics, including frame rate, resolution, and bit rate, was controlled considering both the energy cost and inference accuracy under a data-driven framework. Similarly, joint configuration adaptation and bandwidth allocation for multi-user video analytics systems was investigated in~\cite{WangCan_INFOCOM20}, where the inference accuracy was modeled as a bi-concave function with respect to resolution and frame rate. The data configuration in MEC server-based inference can also be adapted without finding an explicit function of the inference accuracy, which was achieved via DRL and Bayesian online learning in~\cite{HeZhaoliang_TMCCA21} and~\cite{Galanopoulos_INFOCOM21}, respectively. In~\cite{FanW_TII23}, the DNN placement, inference task offloading, data preprocessing, and communication-computational resource allocation are jointly optimized for industrial IoT applications.

\subsection{Device-edge Co-inference}
\label{Device-Edge Co-Inference}

For on-device inference, DNN models are fully executed at EUDs that may drain battery energy quickly. For MEC server-based inference, uploading the possibly high-dimensional data is both time- and energy-consuming. To maximize the benefits of both paradigms, \emph{device-edge co-inference}, which partitions a backbone DNN model between an EUD and the MEC server~\cite{ZhouZ_PIEEE20}, appears to be a more favorable solution. It can be envisaged that device-edge co-inference can improve both time and energy efficiency by harnessing computational resources at different nodes, despite extra communication and computation overhead compared to on-device and MEC server-based inference, respectively.

As shown in Fig.~\ref{fig:inferenceParadigms}, the on-device DNN partition (i.e., the first few layers of the backbone DNN) extracts a feature map from the input for transmission to the MEC server. The MEC server executes the server DNN partition (i.e., the remaining layers of the backbone DNN) using the received feature map to derive the final inference result. In the device-edge co-inference paradigm, there exists a general trade-off between the communication and computation energy consumption for a target inference accuracy, which can be adjusted via model partition point selection~\cite{ShaoJ_MCOM20}. Nevertheless, realizing energy-efficient device-edge co-inference is beyond model partitioning and requires additional optimization as discussed in the sequel.

\emph{1) Adaptive Model Partitioning:} A study in~\cite{KangY_Neurosurgeon17} measures the computation and data overhead of various DNN models at all candidate model partition points. Based on such information, a lightweight scheduler named Neurosurgeon was developed to select the best partition point in runtime. Compared to MEC server-based inference, Neurosurgeon reduces $59.5\%$ of the energy consumption on an EUD. The backbone DNN model in device-edge co-inference can also be a branchy network~\cite{LiEn_TWC20,Laskaridis_MOBICOM20,ZhaoZ_IoTDI21}. In particular, Edgent~\cite{LiEn_TWC20} exploits both model partitioning and early termination, where the optimal model partition point and early exit are determined via a configuration map for dynamic network status. While the early exit of Edgent is fixed for all input once selected, SPINN~\cite{Laskaridis_MOBICOM20} terminates the inference process at different exits in adaptation to the input data. Nevertheless, the large search space of candidate model partition points and early-exit confidence score thresholds necessitates low-complexity optimization. Therefore, EdgeML~\cite{ZhaoZ_IoTDI21} trains a light DRL agent that determines the optimal configurations of a branchy network according to system dynamics, which takes accuracy, latency, and energy consumption into account.

\emph{2) Feature Compression:} Model partitioning amortizes the computation workload between EUDs and the MEC server. However, the communication overhead may be increased for certain DNN models at some partition points because of the \emph{in-layer data amplification phenomenon}~\cite{KangY_Neurosurgeon17,LiHS_ICPADS18}. To achieve energy-efficient device-edge co-inference, it is thus necessary to reduce the feature size by exploiting statistical characteristics~\cite{Eshratifar_ISLPED19}. Empirical results show that lossless compression can achieve a $1.5$$\sim$$3\times$ compression ratio, while lossy compression eliminates the redundancy more aggressively with some accuracy loss~\cite{ChenZhuo_TIP19}. In~\cite{ShaoJW_ICCW19}, feature compression in device-edge co-inference was performed by an autoencoder, which is adapted to the channel condition to strength the robustness. Besides, the EE-oriented design methods for on-device and MEC server-based inference, e.g, model compression~\cite{ShaoJ_MCOM20,Krouka_PIMRC21,ShiW_INFOCOMWKSHOP,ZhangX_GLOBECOM21}, early termination~\cite{ShaoJ_ICASSP20,DongR_JCIN22}, and adaptive communication~\cite{HuDiyi_IoTDI20,LanQ_TWC23}, can be jointly utilized with feature compression. In addition, a theoretical characterization of the communication overhead-inference accuracy trade-off is also critical. This was achieved in~\cite{ShaoJ_JSAC22} via the information bottleneck principle~\cite{NTishby_Allerton99}, and based on which, a task-oriented communication scheme that only transmits sufficient but minimal information relevant to the inference task was developed.

\emph{3) Multi-device Cooperation:} Cooperation among multiple EUDs is a critical augmentation of device-edge co-inference. To realize this, CoEdge~\cite{LieK_TON21} partitions an input image into patches and distributes the processing of different patches to multiple cooperating EUDs. By exploiting the computational resources at proximal EUDs with awareness of their power-characteristics, a $25$$\sim$$67\%$ energy consumption reduction was observed for four DNN models on a realistic prototype, compared to a baseline scheme called Musical Chair~\cite{Hadidi_RAL18} that equally partitions the inference workloads. Several times of inference speedup compared to on-device inference was also achieved. The spatially-correlated views of multiple EUDs can also be harnessed to improve the inference accuracy with a given energy budget, which requires to extract and fuse useful features across EUDs. In~\cite{Teerapittayanon_ICDCS17}, a distributed DNN architecture was proposed, which is scalable across EUDs and achieves a $20\times$ communication cost reduction on a multi-view image classification task compared with offloading the raw images. To further reduce the feature redundancy, various approaches such as view elimination~\cite{ChoiJinhang_DAC19,Singhal_DATE20}, selective feature retransmission~\cite{JShao_TWC23}, and spatial-temporal feature fusion~\cite{ShaoJ_TWC23b}, have been developed. 

When EUDs make independent and concurrent inference requests, model partitioning and computational resource management should be jointly optimized~\cite{TangX_IOTJ21,XuZ_TPDS21,LiuZ_JSAC23,XiaoY_TCOM2302,HaoZ_TMC23}. For example, with a data-driven multi-core CPU model, the optimal model partitioning and computational resource allocation plan was obtained via an alternating optimization algorithm in~\cite{TangX_IOTJ21}. For applications where the inference requests arrive sequentially at EUDs, an online inference request admission policy was developed based on reinforcement learning in~\cite{XuZ_TPDS21} to minimize the energy consumption of both EUDs and MEC servers. In~\cite{LiuZ_JSAC23}, multiple efficient inference techniques, including batching and early termination, were optimized jointly with the allocation of communication and computational resources. Decentralized multi-agent reinforcement learning-based energy-efficient collaborative inference schemes were developed in~\cite{XiaoY_TCOM2302}, which exploits the learning experience of neighboring EUDs to accelerate policy optimization in large-scale edge AI systems. A prototype with two EUDs and two MEC servers showed that more than $40\%$ of the mobile energy consumption and inference latency can be saved compared with a baseline scheme. Assuming an autoencoder-based feature compression scheme, the model partitioning, and channel and power allocation of multiple EUDs were jointly optimized in \cite{HaoZ_TMC23}, which saves up to $56\%$ of inference latency and $72\%$ of energy consumption.

\subsection{Takeaways}
This section reviews the EE-oriented design approaches for three typical edge inference paradigms, including on-device inference, MEC server-based inference, and device-edge co-inference. For on-device inference, the essence is to reduce the computation energy consumption by removing the less important processing of DNNs using model compression, energy-aware NAS, and input adaptive inference techniques. Although efficient processing of DNNs has been investigated for years (e.g., \cite{SzeV_PIEEE17} and references therein), there is still a genuine need to modify the existing solutions for diverse edge AI platforms with strict computational resource and energy constraints. For MEC server-based inference, it is vital to develop energy-efficient inference task offloading strategies, which reduces both the on-device computation energy consumption and transmission energy consumption. These strategies are largely inherited from those developed for conventional MEC systems, yet the distinctive fault-tolerance property of DNN workloads should also be considered via data preprocessing. The device-edge co-inference paradigm provides an agile means to deploy DNN models across multiple computing nodes in edge AI systems. Similar to MEC server-based inference, removing the redundant communication overhead in intermediate feature maps is critical to save the transmission energy. It is also mandatory to strike a good balance among the on-device computation workload, communication overhead, and inference accuracy. In addition, the potential collaboration among EUDs can not only harvest the distributed computational resources, but also benefit from the task dependency and spatial data correlation. We foresee in future edge AI systems, the three edge inference paradigms coexist for heterogeneous applications. Therefore, unifying their respective EE-oriented design methodologies and unraveling scaling laws of energy consumption regarding the EUD and MEC infrastructure densities would be necessary. Moreover, in applications where sampling the inference data incurs significant overhead~\cite{Likamwa_MPRV21}, understanding the impact of sensing on edge inference becomes crucial.

\section{Future Research Directions}\label{FutureSec}

In this section, we highlight several prospective future research directions of green edge AI. Specially, we first introduce the low-power sensing techniques via integrated sensing and communication (ISAC). Hardware/software co-designs for edge AI are then discussed, which complement the existing algorithmic innovations to further boost the EE. Besides, we elaborate two critical aspects of neuromorphic computing for ultra-energy-efficient edge AI, including the spiking neural networks (SNNs) and compute-in-memory (CIM) techniques. In addition, green energy-powered edge AI that can potentially achieve carbon-neutral operations, is also investigated. Finally, we elaborate several possible approaches for delivering generative AI (GenAI) services at mobile edge networks in an energy-efficient manner.

\subsection{Low-Power Sensing via ISAC}

With the seamless integration of sensing and communication, 6G wireless networks are on the verge of revolutionizing edge AI systems, enabling them to perceive and understand the physical world in unprecedented ways~\cite{LiuF_JSAC22}. These advancements can benefit a plethora of emerging applications, including high-accuracy tracking, simultaneous localization, imaging, and mapping, augmented human sensing, as well as autonomous vehicles and the Metaverse.

The key success of ISAC lies in empowering edge AI systems to ``see'' the environment by exploiting radio wave propagation. In this context, MEC servers play a prominent role by utilizing communication signals for device-free sensing~\cite{ShiQ_JSAC22}. By doing so, EUDs are liberated from the power-intensive data acquisition tasks that would otherwise burden the centralized edge learning process. To support ISAC, full-duplex MEC servers need to be deployed. However, this setup introduces new challenges due to the simultaneous arrival of echo sensing signals from sensing targets and uplink transmission signals from EUDs before the downlink transmission concludes. The resulting severe self-interference necessitates the developments of dedicated schemes to effectively manage mutual interference between sensing and communication signals~\cite{XuD_TCOM22}. 

On the other hand, EUDs in edge AI systems can leverage the uplink/sidelink signals for low-power sensing, utilizing either multiple access or D2D channels \cite{LiuA_COMST22}. The sensed data can be offloaded to an MEC server for downstream inference following the novel integrated sensing, computation, and communication (ISCC) paradigm~\cite{WenD_TWC24}. In each ISAC frame, EUDs transmit the sensed data from the previous frame while simultaneously sensing the environment. Nevertheless, a critical issue arises as the conventional uplink transmission beams are typically directed toward the receiver (i.e., the MEC server in edge AI systems), which may not be optimal for sensing the surroundings. Therefore, effective waveform and beamforming designs specifically tailored for edge AI applications are essential to ensure accurate sensing and efficient data offloading~\cite{XiaoZ_JSAC22}. Another important aspect in ISAC-enabled low-power sensing for edge AI is the trade-off between sensing accuracy, communication rate, and intelligence performance~\cite{hua2022mimo}. Proper understanding and management of this triple trade-off are necessary. By optimizing the related parameters, ISAC can empower a broad range of applications with energy-efficient and accurate perception capability to achieve green edge AI.

\subsection{Hardware/software Co-design} \label{Sec:Future:HW-SW}
The existing methods for energy-efficient edge AI, including those reviewed in previous sections, mainly focus on software designs, and their benefits are greatly affected by the underlying hardware implementations~\cite{SzeV_PIEEE17}. For example, implementing DNN inference applications on field programmable gate arrays (FPGAs) and application specific integrated circuits (ASICs) are more energy-efficient than on general-purpose CPUs and GPUs~\cite{Nechi_ACMTRTS23}. Also, a recent study in~\cite{TuX_SEC23} found that even for smartphones of the same brand, the energy consumption and AI performance can be very different for certain DNN inference tasks. It is hence critical to investigate the hardware/software interplay for improving EE of edge AI. 

A precursory study in~\cite{Flavio_MICRO22} suggests to deploy multiple pruned CNNs instead of an unpruned one for on-device inference, which not only increases the robustness and accuracy, but also offers the opportunities to use ultra-low-power processing cores and scale-down the memory access voltage for energy savings. Inspired by NAS, hardware-aware NAS (HW-NAS) was proposed by factoring the hardware-related metrics (e.g., latency, energy consumption, and memory space) in the search objectives and constraints~\cite{Hadjer_IJCAI21}. However, it normally confines the search space to a pool of candidate neural network architectures, and the search results are only suitable for fixed hardware configurations. Given the popularity of reconfigurable AI accelerators such as FPGAs, expanding the search space of HW-aware NAS by incorporating different configurations of hardware platforms is a natural extension. For example,~\cite{JiangW_TCAD20} models the hardware design space as an FPGA pool and develops a two-level exploration framework. The experimental result shows that with hardware/software co-design, the EE achieves a remarkable $54.05\%$ increase on ImageNet classification with Xilinx FPGAs. In addition, cross-layer optimization from hardware, software, to compiler, can be conducted to further improve EE~\cite{HaoC_MDAT21}. The significant challenges of hardware/software co-design include the huge search space and cost of searching algorithms, given diverse DNN architectures and edge AI hardware platforms~\cite{Bringmann_CODESISSS21,ShiY_Slides2022}. To resolve these challenges, it is critical for hardware and software researchers of edge AI to collaborate and establish clear understanding on design objectives and constraints outside their own expertise as suggested in \cite{Bringmann_CODESISSS21,Fasfous_Thesis2022}. Moreover, as DNN training has different computation and memory access patterns compared to DNN inference~\cite{ChenY_ENG20,Kudithipudi_NatElec23}, training-specialized hardware/software co-design strategies are needed to maximize the EE of cooperative edge learning.

\subsection{Neuromorphic Computing}

Human brains are among the most energy-efficient computing devices, which can provide $10^{18}$ floating point operations per second with merely $20$ Watts of power consumption~\cite{Madhavan_23}. Neuromorphic computing endeavors to imitate operations in human brains, where biological neurons communicate with electrochemical signals through action potentials \cite{Yamazaki_BrainSci22}. SNNs are a family of neuromorphic algorithms. They differ from the conventional DNNs by encoding input as time-series and using discrete spikes to transfer information between spiking neurons, each of which can fire a spike with sufficient input strength called membrane potential~\cite{JangH_SPM19}. Thus, SNNs are orders of magnitude more energy-efficient than DNNs especially with the support of neuromorphic chips, such as the IBM's TrueNorth and Intel's Loihi chips. As a proof-of-concept, Accenture Labs demonstrated via an automotive voice control application that, an SNN deployed at Intel's Kapoho Bay, which is a computing stick empowered by Intel's Loihi chips, outperforms a CNN running at a GPU by $1000\times$ and $200$ ms in terms of EE and latency, respectively, while maintaining comparable inference accuracy~\cite{Accenture_20}. Notably, there are recent interests of training SNNs under the framework of FEEL~\cite{Venkatesha_TSP21,YangH_NatCom22} and SNN-based cooperative edge inference~\cite{LiuY_TWC23}. However, SNNs bring new challenges in training~\cite{Eshraghian_PIEEE23} and hardware implementation~\cite{BOUVIER_JETCS19}. Latest studies have also found that EE improvements achieved by SNNs over DNNs vary with inference task types~\cite{PLAGWITZ_arXiv23,Ottati_JETCAS23}. Hence, properly selecting between SNNs and DNNs based on application needs is crucial.

Neuromorphic computing also deviates from the traditional von Neumann architectures with separate units for processing and memory, with which, the energy spent on fetching data from off-chip memory for DNN computations is significant. Hence, it inspires to adopt the CIM techniques for edge AI, which can perform MAC operations in the analog domain within memory subarray~\cite{YuS_MCAS21}. The history of CIM starts from the 1990s, and it receives great attention since the last decade with the developments of transistor and memory technologies~\cite{ChangL_SciCHINA21}. In 2021, an analog matrix processor for high-performance AI inference based on non-volatile flash memory was announced by Mythic$^\circledR$, which delivers 25 tera operations per second (TOPS) but just needs 3 Watts of power, i.e., only $1/10$ of the desktop GPU power consumption \cite{MYTHIC_2021}. However, limitations of current CIM technologies, e.g., sensitive to analog errors~\cite{WanZ_ACMJETCS22} and immature development software~\cite{KHAN_arXiv24}, await to be addressed.

\subsection{Green Energy-powered Edge AI}

Although methods to reduce the non-renewable energy consumption are paramount in no doubt, green energy sources such as solar, wind, and ambient radio frequency (RF) energy that can be scavenged by energy harvesting technologies, are new drivers for energy-efficient IoT applications~\cite{MaD_COMST20}. Performing DNN inference on green energy-powered EUDs thus becomes a perfect candidate to achieve energy-efficient edge inference. However, given the intensive computation workloads, the intermittent green energy availability may frequently interrupt the inference processes, which need to be restarted when sufficient energy is collected~\cite{LvM_ACCESS22}. To tackle this fundamental obstacle, the task-based execution model from the area of intermittent computing was adapted for DNN inference in~\cite{Gobieski_ASPLOS19} to build a full-scale RF energy-powered prototype. Since the available green energy is time-varying, adaptive inference is essential to maximize the long-term average accuracy. This idea was first grounded by deploying a multi-exit DNN in~\cite{WuY_DAC21}, where the best exit is selected via Q-learning in runtime according to the level of available energy. Green energy-aware adaptive inference was also achieved by selecting models with different energy overhead and accuracy performance~\cite{ParkG_ISLPED23}. Nevertheless, these works concentrate on on-device inference, and methods to efficiently exploit the computational resources at the MEC server via device-edge co-inference remains uncharted.

Compared with green energy-powered edge inference, green energy-powered edge learning is much less investigated. Recently, FEEL systems where the EUDs are supported by green energy sources were proposed in~\cite{Basak_ISIT21,Hamdi_IoTJ22}. Likewise, due to the unstable energy availability, EUDs might drop out of the training process unexpectedly, rendering model convergence hard to guarantee with conventional FEEL algorithms such as FedAvg. Fortunately, it was showed in~\cite{Basak_ISIT21} that such an issue can be overcome by re-scaling the learning rate with the green energy arrival probability when the training loss function is strongly convex. Besides, green energy management is also a crucial aspect that was investigated in~\cite{Hamdi_IoTJ22}, where the number of participating green energy-powered EUDs was maximized via joint user scheduling and power allocation. However, a comprehensive understanding on the general convergence behavior of green energy-powered FEEL is still absent. Also, joint learning- and green energy-aware resource management strategies are pending development. Beyond vanilla FEEL, it is interesting to integrate green energy sources with other emerging distributed edge learning paradigms, such as DEEL and H-FEEL discussed in Section~\ref{otherCoLearnFrameworks}.

\subsection{Green Generative Edge AI}
GenAI techniques, which can create contents such as text, images, speech, and videos~\cite{CaoY_arXiv23}, are unleashing the next wave of AI innovations, especially after the release of ChatGPT by OpenAI in late 2022. At the moment, many impressive GenAI applications are underpinned by foundation models (FMs) that have a few to hundreds of billions of parameters and need to be trained on massive datasets~\cite{ZhouJ_arXiv24}. Apparently, training such giant models consumes huge amount of computational and energy resources, and thus it can merely take place at large datacenters. Although with less resource demand, making inference via FMs for GenAI services remains mostly in the Cloud, with prompts being uploaded by EUDs. Similar to other intelligent mobile applications, migrating GenAI services to mobile edge networks, namely generative edge AI, can benefit EUDs with low latency and immersive experience~\cite{LaiB_arXiv23}. However, by the large-size nature of FMs, it is critical to fit the compute-intensive GenAI workloads to the resource-limited MEC environments and improve the EE.

Although it is unlikely to train FMs from scratch within mobile edge networks, fine-tuning pre-trained models with the data and computational resources at EUDs is plausible for personalization and task adaptation. FEEL is promising for this purpose~\cite{Woisetschlager_arXiv23}. However, updating billions of parameters at once may encounter the \emph{memory wall}, e.g., fine-tuning the LLaMA model~\cite{Touvron_LLaMA23} even with a small batch size of 8 incurs more than 20 GB of peak memory usage~\cite{XuM_arXiv23}, not to mention the significant on-device energy consumption and processing latency. Hence, parameter-efficient fune-tuning methods are necessarily, which freeze most model parameters to minimize the on-device computation and communication overhead. To further reduce the on-device computation energy consumption or with extremely restrictive on-device computational resources, the split learning (SL) framework can be applied~\cite{LinZ_arXiv23}. For FM-based inference, existing model compression methods can be adapted for energy-efficient deployments at EUDs or MEC servers~\cite{ZhuX_arXiv23}, which should preserve the versatility and generalizability of FMs. Another approach to reduce the inference energy cost is to promote device cooperation motivated by the fact that requested contents from nearby EUDs are usually correlated. This approach was demonstrated via diffusion model-based image generation in~\cite{DuH_IEEENetw23}, where the shared processing steps are computed on the MEC server while EUDs just need to handle the prompt-specific computations. Moreover, energy-efficient SNNs have recently been found capable of language generation and understanding in~\cite{ZhuRJ_arXiv23} for edge applications. Although still in the  conceptual stage, green generative edge AI is surely an area worthy of attention given the intense conflict between significant computation overhead of GenAI workloads and limited energy availability at EUDs.

\section{Conclusions}\label{ConclusionsSec}

In an era with the proliferation of edge devices and surging demand for real-time intelligent processing, the significant energy consumption associated with edge AI applications poses not only grand operational challenges but also steadily increasing environmental concerns. Understanding and mitigating the energy footprint of the related technologies thus become imperative for building energy-efficient and eco-friendly edge AI systems. This paper delves into the critical realm of green edge AI, concentrating on the intersection of energy efficiency, AI, and edge computing. Our exploration has provided a comprehensive overview of strategies to mitigate the formidable energy challenge of the emerging edge AI landscape. From dissecting the energy consumption in basic tasks of edge AI such as data acquisition, edge learning, and edge inference, to unraveling the design principles that emphasize energy efficiency, adaptability, and the judicious trade-off between intelligence and greenness, this paper navigates through the multifaceted dimensions of green edge AI. Promising future research directions, including integrated sensing and communication, hardware/software co-design, neuromorphic computing, green energy-powered edge AI, and green generative edge AI, have also been highlighted to attract early attention from both academia and industry. In conclusion, it is evident that the synthesis of technological advancements, algorithmic innovations, and strategic energy management is crucial toward realizing the 6G vision of sustainable and pervasive intelligence. We anticipate that this survey paper will function as a timely and valuable reference for researchers, practitioners, and policymakers engaged in shaping the trajectory of green edge AI.


\bibliographystyle{IEEEtran}  
\bibliography{greenEdgeAI}  

\begin{thebibliography}{100}
\providecommand{\url}[1]{#1}
\csname url@samestyle\endcsname
\providecommand{\newblock}{\relax}
\providecommand{\bibinfo}[2]{#2}
\providecommand{\BIBentrySTDinterwordspacing}{\spaceskip=0pt\relax}
\providecommand{\BIBentryALTinterwordstretchfactor}{4}
\providecommand{\BIBentryALTinterwordspacing}{\spaceskip=\fontdimen2\font plus
\BIBentryALTinterwordstretchfactor\fontdimen3\font minus
  \fontdimen4\font\relax}
\providecommand{\BIBforeignlanguage}[2]{{%
\expandafter\ifx\csname l@#1\endcsname\relax
\typeout{** WARNING: IEEEtran.bst: No hyphenation pattern has been}%
\typeout{** loaded for the language `#1'. Using the pattern for}%
\typeout{** the default language instead.}%
\else
\language=\csname l@#1\endcsname
\fi
#2}}
\providecommand{\BIBdecl}{\relax}
\BIBdecl

\bibitem{MoorJ_AIMAG06}
J.~Moor, ``The dartmouth college artificial intelligence conference: {T}he next
  fifty years,'' \emph{AI Mag.}, vol.~27, no.~4, pp. 87--91, Winter 2006.

\bibitem{SilverD_Nature16}
D.~Silver\emph{ et al.}, ``Mastering the game of {G}o with deep neural networks
  and tree search,'' \emph{Nat.}, vol. 529, pp. 484--489, Jan. 2016.

\bibitem{GuoY_NeuCom16}
Y.~Guo, Y.~Liu, A.~Oerlemans, S.~Lao, S.~Wu, and M.~S. Lew, ``Deep learning for
  visual understanding: {A} review,'' \emph{Neurocomput.}, vol. 187, pp.
  27--48, Apr. 2016.

\bibitem{Otter_TNNLS21}
D.~W. Otter, J.~R. Medina, and J.~K. Kalita, ``A survey of the usages of deep
  learning for natural language processing,'' \emph{IEEE Trans. Neural Netw.
  Learn. Syst.}, vol.~32, no.~2, pp. 604--624, Feb. 2021.

\bibitem{Esteva_NatureMed19}
A.~Esteva\emph{ et al.}, ``A guide to deep learning in healthcare,'' \emph{Nat.
  Med.}, vol.~25, pp. 24--29, Jan. 2019.

\bibitem{Khalil_IoTJ21}
R.~A. Khalil\emph{ et al.}, ``Deep learning in the industrial {I}nternet of
  {T}hings: {P}otentials, challenges, and emerging applications,'' \emph{IEEE
  Internet Things J.}, vol.~8, no.~14, pp. 11\,016--11\,040, Jul. 2021.

\bibitem{Ozbayoglu_ASC20}
A.~M. Ozbayoglu, M.~U. Gudelek, and O.~B. Sezer, ``Deep learning for financial
  applications: {A} survey,'' \emph{Appl. Soft Comput.}, vol.~93, p. 106384,
  Aug. 2020.

\bibitem{SzeV_PIEEE17}
V.~Sze, Y.-H. Chen, T.-J. Yang, and J.~S. Emer, ``Efficient processing of deep
  neural networks: {A} tutorial and survey,'' \emph{Proc. IEEE}, vol. 105,
  no.~12, pp. 2295--2329, Dec. 2017.

\bibitem{BIANCHINI_ACMCOMM20}
R.~Bianchini\emph{ et al.}, ``Toward {ML}-centric cloud platforms,''
  \emph{Commun. ACM}, vol.~63, no.~2, pp. 2295--2329, Feb. 2020.

\bibitem{MaoY_COMST17}
Y.~Mao, C.~You, J.~Zhang, K.~Huang, and K.~B. Letaief, ``A survey on mobile
  edge computing: The communication perspective,'' \emph{{IEEE} Commun. Surveys
  Tuts.}, vol.~19, no.~4, pp. 2322--2358, Fourth Quart. 2017.

\bibitem{TongW_Book21}
W.~Tong and P.~Zhu, \emph{6{G}: {T}he {N}ext {H}orizon: {F}rom {C}onnected
  {P}eople and {T}hings to {C}onnected {I}ntelligence}.\hskip 1em plus 0.5em
  minus 0.4em\relax Cambridge, UK: Cambridge University Press, 2021.

\bibitem{Tari_MCC15}
Z.~Tari, X.~Yi, U.~S. Premarathne, P.~Bertok, and I.~Khalil, ``Security and
  privacy in cloud computing: {V}ision, trends, and challenges,'' \emph{IEEE
  Cloud Comput.}, vol.~2, no.~2, pp. 30--38, Mar.-Apr. 2015.

\bibitem{ETSI_MEC_15}
Y.~C. Hu, M.~Patel, D.~Sabella, N.~Sprecher, and V.~Young, ``Mobile edge
  computing - {A} key technology towards 5{G},'' \emph{ETSI White Paper},
  vol.~11, Sep. 2015.

\bibitem{ZhouZ_PIEEE20}
Z.~Zhou, X.~Chen, E.~Li, L.~Zeng, K.~Luo, and J.~Zhang, ``Edge intelligence:
  Paving the last mile of artificial intelligence with edge computing,''
  \emph{Proc. {IEEE}}, vol. 107, no.~8, pp. 1738--1762, Aug. 2019.

\bibitem{KBL_JSAC22}
K.~B. Letaief, Y.~Shi, J.~Lu, and J.~Lu, ``Edge artificial intelligence for
  6{G}: {V}ision, enabling technologies, and applications,'' \emph{{IEEE} J.
  Sel. Areas Commun.}, vol.~40, no.~1, pp. 5--36, Jan 2022.

\bibitem{ChenJ_PIEEE19}
J.~Chen and X.~Ran, ``Deep learning with edge computing: {A} review,''
  \emph{Proc. {IEEE}}, vol. 107, no.~8, pp. 1655--1674, Aug. 2019.

\bibitem{ITU_R_2306}
ITU-R, ``Framework and overall objectives of the future development of {IMT}
  for 2030 and beyond,'' \emph{DRAFT NEW RECOMMENDATION}, Jun. 2023.

\bibitem{Gartner_2022}
J.~Wiles, ``What’s new in artificial intelligence from the 2022 {G}artner
  hype cycle,''
  \url{https://www.gartner.com/en/articles/what-s-new-in-artificial-intelligence-from-the-2022-gartner-hype-cycle},
  Sep. 2022.

\bibitem{EdgeAI_WhitePaper_20}
\BIBentryALTinterwordspacing
E.~Peltonen\emph{ et al.}, ``6{G} white paper on edge intelligence,'' Jun.
  2020. [Online]. Available:
  \url{https://arxiv.org/ftp/arxiv/papers/2004/2004.14850.pdf}
\BIBentrySTDinterwordspacing

\bibitem{DingY_ACMCOMRev22}
A.~Y. Ding\emph{ et al.}, ``Roadmap for edge {AI}: {A} dagstuhl perspective,''
  \emph{ACM SIGCOMM Comput. Commun. Rev.}, vol.~52, no.~1, pp. 28--33, Jan.
  2022.

\bibitem{Desislavov_SusComput23}
R.~Desislavov, F.~M.-Plumed, and J.~H.-Orallo, ``Trends in {AI} inference
  energy consumption: {B}eyond the performance-vs-parameter laws of deep
  learning,'' \emph{{Sustain. Comput.: Inf. Syst.}}, vol.~38, p. 100857, Feb.
  2023.

\bibitem{HornikK_NN89}
K.~Hornik, ``Multilayer feedforward networks are universal approximators,''
  \emph{Neural Netw.}, vol.~2, pp. 359--366, 1989.

\bibitem{DengJ_CVPR09}
J.~Deng, W.~Dong, R.~Socher, L.-J. Li, K.~Li, and F.-F. Li, ``Image{N}et: {A}
  large-scale hierarchical image database,'' in \emph{Proc. IEEE Conf. Comput.
  Vision Pattern Recogn. (CVPR)}, Miami, FL, USA, Jun. 2009.

\bibitem{Alex_NeurIPS12}
A.~Krizhevsky, I.~Sutskever, and G.~E. Hinton, ``{ImageNet} classification with
  deep convolutional neural networks,'' in \emph{Proc. 26th Int. Conf. Neural
  Inf. Process. Syst. (NeurIPS)}, Lake Tahoe, NV, USA, Dec. 2012.

\bibitem{YuJ_CoCa22}
\BIBentryALTinterwordspacing
J.~Yu, Z.~Wang, V.~Vasudevan, L.~Yeung, M.~Seyedhosseini, and Y.~Wu, ``Co{C}a:
  {C}ontrastive captioners are image-text foundation models.'' [Online].
  Available: \url{https://arxiv.org/pdf/2205.01917.pdf}
\BIBentrySTDinterwordspacing

\bibitem{Stanford_AI19}
\BIBentryALTinterwordspacing
R.~Perrault\emph{ el al.}, ``The artificial intelligence index report 2019,''
  Dec. 2019. [Online]. Available:
  \url{https://hai.stanford.edu/sites/default/files/ai_index_2019_report.pdf}
\BIBentrySTDinterwordspacing

\bibitem{Strubell_ACL19}
E.~Strubell, A.~Ganesh, and A.~McCallum, ``Energy and policy considerations for
  deep learning in {NLP},'' in \emph{Proc. Annu. Meeting Assoc. Comput.
  Linguistics (ACL)}, Florence, Italy, Jul. 2019.

\bibitem{WangH_GLOBECOM19}
H.~Wang, B.~Kim, J.~Xie, and Z.~Han, ``How is energy consumed in smartphone
  deep learning apps? {E}xecuting locally vs. remotely,'' in \emph{Proc. IEEE
  Global Commun. Conf. (GLOBECOM)}, Waikoloa, HI, USA, Dec. 2019.

\bibitem{ChatGPTenergy_2023}
Zodhya, ``{How much energy does ChatGPT consume?}''
  \url{https://medium.com/@zodhyatech/how-much-energy-does-chatgpt-consume-4cba1a7aef85},
  May 2023.

\bibitem{Ligozat_Sustainability22}
A.-L. Ligozat, J.~Lefevre, A.~Bugeau, and J.~Combaz, ``Unraveling the hidden
  environmental impacts of {AI} solutions for environment life cycle assessment
  of {AI} solutions,'' \emph{Sustain.}, vol.~14, no.~9, pp. 1--14, Apr. 2022.

\bibitem{Schwartz_ACMCOM19}
R.~Schwartz, J.~Dodge, N.~A. Smith, and O.~Etzioni, ``Green {AI},''
  \emph{Commun. ACM}, vol.~63, no.~12, pp. 54--63, Dec. 2020.

\bibitem{Verdecchia_WILEY23}
J.~S. R.~Verdecchia and L.~Cruz, ``A systematic review of green {AI},''
  \emph{WILEY Data Mining Knowl. Discov.}, vol.~13, no.~4, p. e1507, 2023.

\bibitem{UN_Green-economy}
\BIBentryALTinterwordspacing
``Green economy,'' {U}nited Nation Environment Programme. [Online]. Available:
  \url{https://www.unep.org/regions/asia-and-pacific/regional-initiatives/supporting-resource-efficiency/green-economy}
\BIBentrySTDinterwordspacing

\bibitem{ChenY_MCOM11}
Y.~Chen, S.~Zhang, S.~Xu, and G.~Y. Li, ``Fundamental trade-offs on green
  wireless networks,'' \emph{IEEE Commun. Mag.}, vol.~49, no.~6, pp. 30--37,
  Jun. 2011.

\bibitem{Sharma_Internet17}
P.~Sharma, P.~Pegus-II, D.~Irwin, P.~Shenoy, J.~Goodhue, and J.~Culbert,
  ``Design and operational analysis of a green data center,'' \emph{IEEE
  Internet Comput.}, vol.~21, no.~4, pp. 16--24, Jul./Aug. 2017.

\bibitem{Albreem_Access21}
M.~A. Albreem, A.~M. Sheikh, M.~H. Alsharif, M.~Jusoh, and M.~N.~M. Yasin,
  ``{Green Internet of Things (GIoT)}: {A}pplications, practices, awareness,
  and challenge,'' \emph{IEEE Access}, vol.~9, pp. 38\,833--38\,858, Mar. 2021.

\bibitem{MastelicT_CC15}
T.~Mastelic and I.~Brandic, ``Recent trends in energy-efficient cloud
  computing,'' \emph{IEEE Cloud Comput.}, vol.~2, no.~1, pp. 40--47, Jan.-Feb.
  2015.

\bibitem{Tractia_18}
Tractica, ``Artificial intelligence edge device shipments to reach 2.6 billion
  units annually by 2025,''
  \url{https://www.edge-ai-vision.com/2018/09/artificial-intelligence-edge-device-shipments-to-reach-2-6-billion-units-annually-by-2025/},
  Sep. 2018.

\bibitem{ShiY_COMST20}
Y.~Shi, K.~Yang, T.~Jiang, J.~Zhang, and K.~B. Letaief,
  ``Communication-efficient edge {AI:} {A}lgorithms and systems,'' \emph{{IEEE}
  Commun. Surveys Tuts.}, vol.~22, no.~4, pp. 2167--2191, Fourth Quart. 2020.

\bibitem{XuW_JSTSP23}
W.~Xu, Z.~Yang, D.~W.-K. Ng, M.~Levorato, Y.~C. Eldar, and M.~Debbah, ``Edge
  learning for {B5G} networks with distributed signal processing: {S}emantic
  communication, edge computing, and wireless sensing,'' \emph{IEEE J. Sel.
  Topics Signal Process.}, vol.~17, no.~1, pp. 9--39, Jan. 2023.

\bibitem{Baccour_COMST22}
E.~Baccour\emph{ et al.}, ``Pervasive {AI} for {I}o{T} applications:
  {R}esource-efficient distributed artificial intelligence,'' \emph{IEEE
  Commun. Surveys Tuts.}, vol.~24, no.~4, pp. 2366--2418, Fourth Quart. 2022.

\bibitem{XuDianlei_PIEEE21}
D.~Xu\emph{ et al.}, ``Edge intelligence: {E}mpowering intelligence to the edge
  of network,'' \emph{Proc. IEEE}, vol. 109, no.~11, pp. 1778--1837, Nov. 2021.

\bibitem{LimW_COMST20}
W.~Y.~B. Lim\emph{ et al.}, ``Federated learning in mobile edge networks: {A}
  comprehensive survey,'' \emph{{IEEE} Commun. Surveys Tuts.}, vol.~22, no.~3,
  pp. 2031--2063, Third Quart. 2020.

\bibitem{Mazumder_JESTCS21}
A.~N. Mazumder\emph{ et al.}, ``A survey on the optimization of neural network
  accelerators for micro-{AI} on-device inference,'' \emph{IEEE J. Emerg. Sel.
  Topics Circuits Syst.}, vol.~11, no.~4, pp. 532--547, Dec. 2021.

\bibitem{LiuD_Neurocomputing22}
D.~Liu, H.~Kong, X.~Luo, W.~Liu, and R.~Subramaniam, ``Bringing {AI} to edge:
  {F}rom deep learning’s perspective,'' \emph{Neurocomput.}, vol. 485, pp.
  297--320, May 2022.

\bibitem{RenQ_MIR23}
W.-Q. Ren\emph{ et al.}, ``A survey on collaborative {DNN} inference for edge
  intelligence,'' \emph{Mach. Intell. Res.}, vol.~20, no.~3, pp. 370--395, Jun.
  2023.

\bibitem{Shuvo_PIEEE23}
M.~M.~H. Shuvo, S.~K. Islam, J.~Cheng, and B.~I. Morshed, ``Efficient
  acceleration of deep learning inference on resource-constrained edge devices:
  {A} review,'' \emph{Proc. IEEE}, vol. 111, no.~1, pp. 42--91, Jan. 2023.

\bibitem{JJXu_arXiv21}
\BIBentryALTinterwordspacing
J.~Xu, W.~Zhou, Z.~Fu, and H.~Zhou, ``A survey on green deep learning.''
  [Online]. Available: \url{https://arxiv.org/pdf/2111.05193.pdf}
\BIBentrySTDinterwordspacing

\bibitem{LanQ_JCIN21}
Q.~Lan, D.~Wen, Z.~Zhang, Q.~Zeng, X.~Chen, P.~Popovski, and K.~Huang, ``What
  is semantic communication? {A} view on conveying meaning in the era of
  machine intelligence,'' \emph{J. Commun. Inf. Netw.}, vol.~6, no.~4, pp.
  336--371, Dec. 2021.

\bibitem{ZhuG_SciChina23}
G.~Zhu\emph{ et al.}, ``Pushing {AI} to wireless network edge: an overview on
  integrated sensing, communication, and computation towards {6G},''
  \emph{SCIENCE CHINA Inf. Sci.}, p. 130301:1–130301:19, Mar. 2023.

\bibitem{WangAlex_ICLR19}
A.~Wang, A.~Singh, J.~Michael, F.~Hill, O.~Levy, and S.~R. Bowman, ``{GLUE:}
  {A} multi-task benchmark and analysis platform for natural language
  understanding,'' in \emph{Proc. Int. Conf. Learn. Repr. (ICML)}, New Orleans,
  LA, USA, May 2019.

\bibitem{LiuD_TCCN21}
D.~Liu, G.~Zhu, J.~Zhang, and K.~Huang, ``Data-importance aware user scheduling
  for communication-efficient edge machine learning,'' \emph{{IEEE} Trans.
  Cogn. Commun. Netw.}, vol.~7, no.~1, pp. 265--278, Mar. 2021.

\bibitem{ZhouZ_ICTAI19}
Z.~Zhou, V.~Tam, K.~S. Lui, E.~Y. Lam, A.~Yuen, X.~Hu, and N.~Law, ``Applying
  deep learning and wearable devices for educational data analytics,'' in
  \emph{Proc. IEEE Int. Conf. Tools Artif. Intell. (ICTAI)}, Portland, OR, USA,
  Nov. 2019.

\bibitem{JiaL_IOTJ23}
L.~Jia, Z.~Zhou, F.~Xu, and H.~Jin, ``Cost-efficient continuous edge learning
  for artificial intelligence of things,'' \emph{IEEE Internet Things J.},
  vol.~9, no.~10, pp. 7325--7337, May 2022.

\bibitem{LiT_MSP20}
T.~Li, A.~K. Sahu, A.~Talwalkar, and V.~Smith, ``Federated learning:
  Challenges, methods, and future directions,'' \emph{IEEE Signal Process.
  Mag.}, vol.~37, no.~3, pp. 50--60, May 2020.

\bibitem{ShaoJ_MCOM20}
J.~Shao and J.~Zhang, ``Communication-computation trade-off in
  resource-constrained edge inference,'' \emph{{IEEE} Commun. Mag.}, vol.~58,
  no.~12, pp. 20--26, Dec. 2020.

\bibitem{OV5675}
OMNIVISION, ``{OV5675 - 5-megapixel product brief},''
  \url{https://www.ovt.com/wp-content/uploads/2023/08/OV5675-PB-v1.3-WEB.pdf}.

\bibitem{Intel_CameraL515}
INTEL, ``{Intel® RealSense$^{\text{TM}}$ LiDAR Camera L515 - Datasheet},''
  \url{https://www.intelrealsense.com/download/7691/}.

\bibitem{TI_AWRL1432}
TI, ``{Single-chip low-power 76-GHz to 81-GHz automotive mmWave radar sensor -
  AWRL1432},'' \url{https://www.ti.com/product/AWRL1432}.

\bibitem{ROHM_BH1790GLC}
ROHM, ``{Optical sensor for heart rate monitor IC - BH1790GLC},''
  \url{https://fscdn.rohm.com/en/products/databook/datasheet/ic/sensor/pulse_wave/bh1790glc-e.pdf}.

\bibitem{Wifienergy}
P.~Riihikallio, ``8 reasons to turn down the transmit power of your {Wi-Fi},''
  \url{https://metis.fi/en/2017/10/txpower/}, Oct. 2017.

\bibitem{3GPP_TS38101}
3GPP, ``User equipment {(UE)} radio transmission and reception,'' \emph{3GPP TS
  38.101-1 V15.23.0}, Sep. 2023.

\bibitem{HeKaiMing_CVPR16}
K.~He, X.~Zhang, S.~Ren, and J.~Sun, ``Deep residual learning for image
  recognition,'' in \emph{Proc. IEEE Conf. Comput. Vision Pattern Recogn.
  (CVPR)}, Las Vegas, NV, USA, Jun. 2016.

\bibitem{SimonyanK_ICLR15}
K.~Simonyan and A.~Zisserman, ``Very deep convolutional networks for
  large-scale image recognition,'' in \emph{Proc. Int. Conf. Learn. Repr.
  ({ICLR})}, San Diego, CA, USA, May 2015.

\bibitem{WuY_TCAD20}
Y.~Wu, Z.~Wang, Y.~Shi, and J.~Hu, ``Enabling on-device {CNN} training by
  self-supervised instance filtering and error map pruning,'' \emph{IEEE Trans.
  Comput.-Aided Design Integr. Circuits Syst.}, vol.~39, no.~11, pp.
  3445--3457, Nov. 2020.

\bibitem{MWelling_ICML18}
M.~Welling, ``{Intelligence per kilowatts-hour},'' Jul. 2018, {K}eynote Speech
  at the 2018 Int. Conf. Mach. Learn. (ICML).

\bibitem{Hasan_COMST11}
Z.~Hasan, H.~Boostanimehr, and V.~K. Bhargava, ``Green cellular networks: {A}
  survey, some research issues and challenges,'' \emph{IEEE Commun. Surveys
  Tuts.}, vol.~13, no.~4, pp. 524--540, Fourth Quart. 2011.

\bibitem{Sevilla_IJCNN22}
J.~Sevilla, L.~Heim, A.~Ho, T.~Besiroglu, M.~Hobbhahn, and P.~Villalobos,
  ``Compute trends across three eras of machine learning,'' in \emph{Proc. Int.
  Joint Conf. Neural Netw. (IJCNN)}, Padua, Italy, Jul. 2022.

\bibitem{HUITZIL_ACCESS17}
C.~T.-Huitzil and B.~Girau, ``Fault and error tolerance in neural networks: {A}
  review,'' \emph{{IEEE} Access}, vol.~5, pp. 17\,322--17\,341, Sep. 2017.

\bibitem{RuospoA_Comput23}
A.~Ruospo, E.~Sanchez, L.~M. L.~L. Dilillo, M.~Traiola, and A.~Bosio, ``A
  survey on deep learning resilience assessment methodologies,''
  \emph{Comput.}, vol.~56, no.~2, pp. 57--66, Feb. 2023.

\bibitem{Rosenfeld_ICLR20}
J.~S. Rosenfeld, A.~Rosenfeld, Y.~Belinkov, and N.~Shavit, ``A constructive
  prediction of the generalization error across scales,'' in \emph{Proc. Int.
  Conf. Learn. Repr. ({ICLR})}, Virtual Conf., Apr.-May 2020.

\bibitem{Johnson_ACL20}
M.~Johnson, P.~Anderson, M.~Dras, and M.~Steedman, ``Predicting accuracy on
  large datasets from smaller pilot data,'' in \emph{Proc. 56th Annu. Meeting
  Assoc. Comput. Linguistics (ACL)}, Melbourne, VIC, Australia, Jul. 2018.

\bibitem{ZhuG_MCOM20}
G.~Zhu, D.~Liu, Y.~Du, C.~You, J.~Zhang, and K.~Huang, ``Toward an intelligent
  edge: {W}ireless communication meets machine learning,'' \emph{IEEE Commun.
  Mag.}, vol.~58, no.~1, pp. 19--25, Jan. 2020.

\bibitem{CarlosJ_TCSVT17}
J.~C. SanMiguel and A.~Cavallaro, ``Energy consumption models for smart camera
  networks,'' \emph{IEEE Trans. Circuits Syst. Video Techn.}, vol.~27, no.~12,
  pp. 2661--2674, Dec. 2017.

\bibitem{Likamwa_MPRV21}
R.~Likamwa, J.~Hu, V.~Kodukula, and Y.~Liu, ``Adaptive resolution-based
  tradeoffs for energy-efficient visual computing systems,'' \emph{IEEE
  Pervasive Comput.}, vol.~20, no.~2, pp. 18--26, Apr.-Jun. 2021.

\bibitem{Giouroukis_DEBS20}
D.~Giouroukis, A.~Dadiani, J.~Traub, S.~Zeuch, and V.~Markl, ``A survey of
  adaptive sampling and filtering algorithms for the {Internet of Things},'' in
  \emph{Proc. 14th ACM Int. Conf. Distrib. Event-based Syst. (DEBS)}, Montreal,
  QC, Canada, Jul. 2020.

\bibitem{UTKUDEMIRELB_ACMJCOMPHEALTH22}
B.~Utkudemirel, L.~Chen, and M.~A.~A. Faruque, ``Data-driven energy-efficient
  adaptive sampling using deep reinforcement learning,'' \emph{ACM Trans.
  Comput. Healthcare}, vol.~4, no.~3, p. Article 19, Sep. 2023.

\bibitem{KarakiA_IWCMC19}
A.~Karaki, A.~Nasser, C.~A. Jaoude, and H.~Harb, ``An adaptive sampling
  technique for massive data collection in distributed sensor networks,'' in
  \emph{Proc. IEEE Int. Wireless Commun. Mob. Comput. Conf. (IWCMC)}, Tangier,
  Morocco, Jun. 2019.

\bibitem{LouP_Sensors20}
P.~Lou, L.~Shi, X.~Zhang, Z.~Xiao, and J.~Yan, ``A data-driven adaptive
  sampling method based on edge computing,'' \emph{Sensors}, vol.~20, no.~8,
  Apr. 2020.

\bibitem{Sushmita_TGCN22}
S.~Ghosh, S.~De, S.~Chatterjee, and M.~Portmann, ``Edge intelligence framework
  for data-driven dynamic priority sensing and transmission,'' \emph{IEEE
  Trans. Green Commun. Netw.}, vol.~6, no.~1, pp. 376--390, Mar. 2022.

\bibitem{ChengW_AAAI18}
W.~Cheng, S.~Erfani, R.~Zhang, and R.~Kotagiri, ``Learning datum-wise sampling
  frequency for energy-efficient human activity recognition,'' in \emph{Proc.
  AAAI Conf. Artif. Intell.}, New Orleans, LA, USA, Apr. 2018.

\bibitem{Siddique_JTRC19}
C.~Siddique and X.~Ban, ``State-dependent self-adaptive sampling {(SAS)} method
  for vehicle trajectory data,'' \emph{Transp. Res. Part C: Emerg. Techn.},
  vol. 100, pp. 224--237, 2019.

\bibitem{LiuD_TWC21}
D.~Liu, G.~Zhu, Q.~Zeng, J.~Zhang, and K.~Huang, ``Wireless data acquisition
  for edge learning: Data-importance aware retransmission,'' \emph{{IEEE}
  Trans. Wireless Commun.}, vol.~20, no.~1, pp. 406--420, Mar. 2021.

\bibitem{ZhouS_ICC20}
X.~Huang and S.~Zhou, ``Adaptive transmission for edge learning via training
  loss estimation,'' in \emph{Proc. IEEE Int. Conf. Commun. (ICC)}, Dublin,
  Ireland, Jun. 2020.

\bibitem{SettlesB_2012}
B.~Settles, \emph{Active Learning}, ser. Synthesis Lectures on Artificial
  Intelligence and Machine Learning.\hskip 1em plus 0.5em minus 0.4em\relax
  Morgan {\&} Claypool Publishers, 2012.

\bibitem{ZengZ_WCL21}
Z.~Zeng, Y.~Liu, W.~Tang, and F.~Chen, ``Noise is useful: Exploiting data
  diversity for edge intelligence,'' \emph{IEEE Wireless Commun. Lett.},
  vol.~7, no.~1, pp. 957--961, May 2021.

\bibitem{ChinchaliS_arXiv20}
S.~Chinchali\emph{ et al.}, ``Sampling training data for continual learning
  between robots and the cloud,'' in \emph{Proc. Int. Symp. Experimental
  Robot.}, Malta, Nov. 2021.

\bibitem{Khong_IECON20}
C.-K. Tham and R.~Rajagopalan, ``Active learning for {IoT} data prioritization
  in edge nodes over wireless networks,'' in \emph{Proc. 46th Annu. Conf. IEEE
  Ind. Electron. Soc. (IECON)}, Singapore, Oct. 2020.

\bibitem{AzarJ_FGCS19}
J.~Azar, A.~Makhoul, M.~Barhamgi, and R.~Couturier, ``An energy efficient {IoT}
  data compression approach for edge machine learning,'' \emph{Future Gen.
  Comput. Syst.}, vol.~96, pp. 168--175, Jul. 2019.

\bibitem{WangShuai_TWC20}
S.~Wang, Y.-C. Wu, M.~Xia, R.~Wang, and H.~V. Poor, ``Machine intelligence at
  the edge with learning centric power allocation,'' \emph{IEEE Trans. Wireless
  Commun.}, vol.~19, no.~11, pp. 7293--7308, Jul. 2020.

\bibitem{ZhouLiangKai_TVT21}
L.~Zhou\emph{ et al.}, ``Learning centric wireless resource allocation for edge
  computing: {A}lgorithm and experiment,'' \emph{IEEE Trans. Veh. Techn.},
  vol.~70, no.~1, pp. 1035--1040, Jan. 2021.

\bibitem{XieH_TWC23}
H.~Xie, M.~Xia, P.~Wu, S.~Wang, and H.~V. Poor, ``Edge learning for large-scale
  {Internet of Things} with task-oriented efficient communication,'' \emph{IEEE
  Trans. Wireless Commun.}, vol.~22, no.~12, pp. 9517--9532, Dec. 2023.

\bibitem{LiXY_TWC22}
X.~Li, S.~Wang, G.~Zhu, Z.~Zhou, K.~Huang, and Y.~Gong, ``Data partition and
  rate control for learning and energy efficient edge intelligence,''
  \emph{IEEE Trans. Wireless Commun.}, vol.~21, no.~11, pp. 9127--9142, Nov.
  2022.

\bibitem{Merluzzi_ACCESS21}
M.~Merluzzi, P.~D. Lorenzo, and S.~Barbarossa, ``Wireless edge machine
  learning: {R}esource allocation and trade-offs,'' \emph{IEEE Access}, vol.~9,
  pp. 45\,377--45\,398, Mar. 2021.

\bibitem{Shorten_IOTJ21}
C.~Shorten and T.~M. Khoshgoftaar, ``A survey on image data augmentation for
  deep learning,'' \emph{J. Big Data}, vol.~6, pp. 1--48, 2019.

\bibitem{HuangH_arXiv18}
\BIBentryALTinterwordspacing
H.~Huang, P.~S. Yu, and C.~Wang, ``An introduction to image synthesis with
  generative adversarial nets.'' [Online]. Available:
  \url{http://arxiv.org/abs/1803.04469}
\BIBentrySTDinterwordspacing

\bibitem{Maayan_ISBI18}
M.~F.-Adar, E.~Klang, M.~Amitai, J.~Goldberger, and H.~Greenspan, ``Synthetic
  data augmentation using {GAN} for improved liver lesion classification,'' in
  \emph{Proc. IEEE Int. Symp. Biomed. Imag. (ISBI)}, Washington, DC, USA, Apr.
  2018.

\bibitem{Sandfort_SciReport19}
V.~Sandfort, K.~Yan, P.~J. Pickhardt, and R.~M. Summers, ``Data augmentation
  using generative adversarial networks {(CycleGAN)} to improve
  generalizability in {CT} segmentation tasks,'' \emph{Sci. Report}, vol.~9, p.
  16884, Nov. 2019.

\bibitem{ParkD_INTERSPEECH19}
D.~S. Park\emph{ et al.}, ``Spec{A}ugment: {A} simple data augmentation method
  for automatic speech recognition,'' in \emph{Proc. 20th Annu. Conf. Int.
  Speech Commun. Assoc. (INTERSPEECH)}, Graz, Austria, Sep. 2019.

\bibitem{YunS_arXiv20}
\BIBentryALTinterwordspacing
S.~Yun, S.~J. Oh, B.~Heo, D.~Han, and J.~Kim, ``Videomix: Rethinking data
  augmentation for video classification.'' [Online]. Available:
  \url{https://arxiv.org/pdf/2012.03457.pdf}
\BIBentrySTDinterwordspacing

\bibitem{FengY_IJCNLP21}
S.~Y. Feng\emph{ et al.}, ``A survey of data augmentation approaches for
  {NLP},'' in \emph{Findings Assoc. Comput. Linguistics: ACL-IJCNLP 2021}, Aug.
  2021.

\bibitem{MaximilianE_ICASSP20}
M.~E. Tschuchnig, C.~Ferner, and S.~Wegenkittl, ``Sequential {IoT} data
  augmentation using generative adversarial networks,'' in \emph{Proc. IEEE
  Int. Conf. Acoust., Speech Signal Process. (ICASSP)}, Barcelona, Spain, May
  2020.

\bibitem{HeY_ApplSci20}
Y.~He, B.~Fu, J.~Yu, R.~Li, and R.~Jiang, ``Efficient learning of healthcare
  data from {IoT} devices by edge convolution neural networks,'' \emph{MDPI
  Appl. Sci.}, vol.~10, no.~24, p. 8934, Dec. 2020.

\bibitem{ZhangJ_IoTJ21}
J.~Zhang\emph{ et al.}, ``Data augmentation and {Dense-LSTM} for human activity
  recognition using {WiFi} signal,'' \emph{IEEE Internet Things J.}, vol.~8,
  no.~6, pp. 4628--4641, Mar. 2021.

\bibitem{PanJ_TKDE10}
S.~J. Pan and Q.~Yang, ``A survey on transfer learning,'' \emph{IEEE Trans.
  Knowl. Data Eng.}, vol.~22, no.~10, pp. 1345--1359, Oct. 2010.

\bibitem{ZhuangF_ProcIEEE21}
F.~Zhuang\emph{ et al.}, ``A comprehensive survey on transfer learning,''
  \emph{Proc. IEEE}, vol. 109, no.~1, pp. 43--76, Jan. 2021.

\bibitem{SufianA_CISS21}
A.~Sufian, C.~You, and M.~Dong, ``A deep transfer learning-based edge computing
  method for home health monitoring,'' in \emph{Proc. Annu. Conf. Inf. Sci.
  Syst. (CISS)}, Baltimore, MD, USA, Mar. 2021.

\bibitem{YangJ_IoTJ20}
J.~Yang, H.~Zou, S.~Cao, Z.~Chen, and L.~Xie, ``{MobileDA}: {T}oward
  edge-domain adaptation,'' \emph{IEEE Internet Things J.}, vol.~7, no.~8, pp.
  6909--6918, Aug. 2020.

\bibitem{LiuX_IoTJ21}
X.~Liu, W.~Yu, F.~Liang, D.~Griffith, and N.~Golmie, ``Toward deep transfer
  learning in industrial {I}nternet of {T}hings,'' \emph{IEEE Internet Things
  J.}, vol.~8, no.~15, pp. 12\,163--12\,175, Aug. 2021.

\bibitem{LuCH_IOTJ21}
C.-H. Lu and X.-Z. Lin, ``Toward direct edge-to-edge transfer learning for
  {IoT}-enabled edge cameras,'' \emph{IEEE Internet Things J.}, vol.~8, no.~6,
  pp. 4931--4943, Mar. 2021.

\bibitem{Khan_IoTJ22}
D.~Khan and I.~W.-H. Ho, ``{CrossCount:} efficient device-free crowd counting
  by leveraging transfer learning,'' \emph{IEEE Internet Things J.}, vol.~10,
  no.~5, pp. 4049--4058, Mar. 2023.

\bibitem{WangY_ACMComputSurveys20}
Y.~Wang, Q.~Yao, J.~T. Kwok, and L.~M. Ni, ``Generalizing from a few examples:
  {A} survey on few-shot learning,'' \emph{ACM Comput. Surveys}, vol.~53,
  no.~3, pp. 1--34, Jun. 2020.

\bibitem{ZengY_TSUSC20}
Y.~Zeng, B.~Song, C.~Yuwen, X.~Du, and M.~Guizani, ``Few-shot scale-insensitive
  object detection for edge computing platform,'' \emph{IEEE Trans. Sustain.
  Comput.}, pp. 726--735, Oct.-Dec. 2022.

\bibitem{YangL_IOTJ21}
L.~Yang, Y.~Li, J.~Wang, and N.~N. Xiong, ``{FSLM}: {A}n intelligent few-shot
  learning model based on siamese networks for {IoT} technology,'' \emph{IEEE
  Internet Things J.}, vol.~8, no.~12, pp. 9717--9729, Jun. 2021.

\bibitem{ZhaoZ_CL22}
Z.~Zhao, Y.~Lai, Y.~Wang, W.~Jia, and H.~He, ``A few-shot learning based
  approach to {IoT} traffic classification,'' \emph{IEEE Commmun. Lett.},
  vol.~26, no.~3, pp. 537--541, Mar. 2022.

\bibitem{TilmonB_CVPR23}
B.~Tilmon\emph{ et al.}, ``Energy-efficient adaptive {3D} sensing,'' in
  \emph{Proc. {IEEE} Conf. Comput. Vision Pattern Recogn. ({CVPR})}, Vancouver,
  BC, Canada, Jun. 2023.

\bibitem{Rubaie_IEEESP19}
M.~A.-Rubaie and J.~M. Chang, ``Privacy-preserving machine learning: {T}hreats
  and solutions,'' \emph{IEEE Security Privacy}, vol.~17, no.~2, pp. 49--58,
  Mar.-Apr. 2019.

\bibitem{TensorFlow_onDeviceTraining}
\BIBentryALTinterwordspacing
TensorFlow, ``On-device training with tensorflow lite.'' [Online]. Available:
  \url{https://www.tensorflow.org/lite/examples/on_device_training/overview}
\BIBentrySTDinterwordspacing

\bibitem{DasA_arXiv22}
A.~Das, Y.~D. Kwon, J.~Chauhan, and C.~Mascolo, ``Enabling on-device smartphone
  {GPU} based training: {L}essons learned,'' in \emph{Proc. IEEE Int. Conf.
  Pervasive Comput. Commun. (PerCom) Wkshops.}, Pisa, Italy, Mar. 2022.

\bibitem{YangQ_ACMTIST19}
Q.~Yang, Y.~Liu, T.~Chen, and Y.~Tong, ``Federated machine learning: Concept
  and applications,'' \emph{{ACM} Trans. Intell. Syst. Techn.}, vol.~10, no.~2,
  p.~12, Mar. 2019.

\bibitem{LiuW_TSIPN22}
W.~Liu, L.~Chen, and W.~Zhang, ``Decentralized federated learning: Balancing
  communication and computing costs,'' \emph{IEEE Trans. Signal Inf. Process.
  Netw.}, vol.~8, pp. 131--143, Feb. 2023.

\bibitem{LiuL_TWC23}
L.~Liu, J.~Zhang, S.~Song, and K.~B. Letaief, ``Hierarchical federated learning
  with quantization: {C}onvergence analysis and system design,'' \emph{IEEE
  Trans. Wireless Commun.}, vol.~22, no.~1, pp. 2--18, Jan. 2023.

\bibitem{McMahanB_AISTATS17}
B.~McMahan, E.~Moore, D.~Ramage, S.~Hampson, and B.~A. y~Arcas,
  ``Communication-efficient learning of deep networks from decentralized
  data,'' in \emph{Proc. Int. Conf. Artif. Intell. Statistics (AISTATS)}, Fort
  Lauderdale, FL, USA, Apr. 2017.

\bibitem{ChenM_JSAC21}
M.~Chen, D.~Gündüz, K.~Huang, W.~Saad, M.~Bennis, A.~V. Feljan, and H.~V.
  Poor, ``Distributed learning in wireless networks: {R}ecent progress and
  future challenges,'' \emph{{IEEE J. Sel. Areas Commun.}}, vol.~39, no.~12,
  pp. 3579--3605, Dec. 2021.

\bibitem{Justus_BigData18}
D.~Justus, J.~Brennan, S.~Bonner, and A.~S. McGough, ``Predicting the
  computational cost of deep learning models,'' in \emph{Proc. IEEE Int. Conf.
  Big Data}, Seattle, WA, USA, Dec. 2018.

\bibitem{Kim_TWC22}
M.~Kim, W.~Saad, M.~Mozaffari, and M.~Debbah, ``Green, quantized federated
  learning over wireless networks: An energy-efficient design,'' \emph{IEEE
  Trans. Wireless Commun.}, vol.~23, no.~2, pp. 1386--1402, Feb. 2024.

\bibitem{Hubara_NIPS2016}
I.~Hubara, M.~Courbariaux, D.~Soudry, R.~El{-}Yaniv, and Y.~Bengio, ``Binarized
  neural networks,'' in \emph{Proc. 30th Conf. Neural Inf. Proc. Syst.
  (NeurIPS)}, Barcelona, Spain, Dec. 2016.

\bibitem{YangY_JSAC21}
Y.~Yang, Z.~Zhang, and Q.~Yang, ``Communication-efficient federated learning
  with binary neural networks,'' \emph{IEEE J. Sel. Areas Commun.}, vol.~39,
  no.~12, pp. 3836--3850, Dec. 2021.

\bibitem{AlistarhD_NeurIPS17}
D.~Alistarh, D.~Grubic, J.~Li, R.~Tomioka, and M.~Vojnovic, ``{QSGD}:
  Communication-efficient {SGD} via gradient quantization and encoding,'' in
  \emph{Proc. 31st Conf. Neural Inf. Process. Syst. (NeurIPS)}, Long Beach, CA,
  USA, Dec. 2017.

\bibitem{Reisizadeh_AISTATS20}
A.~Reisizadeh, A.~Mokhtari, H.~Hassani, A.~Jadbabaie, and R.~Pedarsani,
  ``{FedPAQ}: {A} communication-efficient federated learning method with
  periodic averaging and quantization,'' in \emph{Proc. Int. Conf. Aritif.
  Intell. Stats. (AISTATS)}, Palermo, Italy, Aug. 2020.

\bibitem{Bernstein_ICML18}
J.~Bernstein, Y.-X. Wang, K.~Azizzadenesheli, and A.~Anandkumar, ``sign{SGD}:
  {C}ompressed optimisation for non-convex problems,'' in \emph{Proc. Int.
  Conf. Mach. Learn. (ICML)}, Stockholm, Sweden, Jun. 2018.

\bibitem{JinR_TWC22}
R.~Jin, X.~He, and H.~Dai, ``Communication efficient federated learning with
  energy awareness over wireless networks,'' \emph{IEEE Trans. Wireless
  Commun.}, vol.~21, no.~7, pp. 5204--5219, Jul. 2022.

\bibitem{Jhunjhunwala_ICASSP21}
D.~Jhunjhunwala, A.~Gadhikar, G.~Joshi, and Y.~C. Eldar, ``Adaptive
  quantization of model updates for communication-efficient federated
  learning,'' in \emph{Proc. IEEE Int. Conf. Acoust., Speech Signal Process.
  (ICASSP)}, Toronto, ON, Canada, Jun. 2021.

\bibitem{Shlezinger_TSP21}
N.~Shlezinger, M.~Chen, Y.~C. Eldar, H.~V. Poor, and S.~Cui, ``{UVeQFed}:
  Universal vector quantization for federated learning,'' \emph{IEEE Trans.
  Signal Process.}, vol.~69, pp. 500--514, 2021.

\bibitem{LinY_ICLR18}
Y.~Lin, S.~Han, H.~Mao, Y.~Wang, and W.~Dally, ``Deep gradient compression:
  Reducing the communication bandwidth for distributed training,'' in
  \emph{Proc. Int. Conf. Learn. Repr. (ICLR)}, Vancouver, BC, Canada, Jun.
  2018.

\bibitem{HanP_ICDCS20}
P.~Han, S.~Wang, and K.~K. Leung, ``Adaptive gradient sparsification for
  efficient federated learning: {A}n online learning approach,'' in \emph{Proc.
  IEEE Int. Conf. Distr. Comput. Syst. (ICDCS)}, Singapore, Dec. 2020.

\bibitem{LiP_TWC23}
P.~Li\emph{ et al.}, ``Snowball: {E}nergy efficient and accurate federated
  learning with coarse-to-fine compression over heterogeneous wireless edge
  devices,'' \emph{IEEE Trans. Wireless Commun.}, vol.~22, no.~10, pp.
  6778--6792, Oct. 2023.

\bibitem{LiuS_WCL21}
S.~Liu, G.~Yu, R.~Yin, and J.~Yuan, ``Adaptive network pruning for wireless
  federated learning,'' \emph{IEEE Wireless Commun. Lett.}, vol.~10, no.~7, pp.
  1572--1576, Jul. 2021.

\bibitem{WenD_WCL19}
D.~Wen, K.-J. Jeon, and K.~Huang, ``Federated dropout – a simple approach for
  enabling federated learning on resource constrained devices,'' \emph{{IEEE
  Wireless Commun. Lett.}}, vol.~11, no.~5, pp. 923 -- 927, May 2022.

\bibitem{JiangY_TNNLS22}
Y.~Jiang, S.~Wang, V.~Valls, B.~J. Ko, W.-H. Lee, K.~K. Leung, and
  L.~Tassiulas, ``Model pruning enables efficient federated learning on edge
  devices,'' \emph{IEEE Trans. Neural Netw. Learn. Syst.}, vol.~34, no.~12, pp.
  10\,374--10\,386, Dec. 2023.

\bibitem{LiLiang_INFOCOM21}
L.~Li, D.~Shi, R.~Hou, H.~Li, M.~Pan, and Z.~Han, ``To talk or to work:
  {F}lexible communication compression for energy efficient federated learning
  over heterogeneous mobile edge devices,'' in \emph{Proc. IEEE Int. Conf.
  Comput. Commun. (INFOCOM)}, Vancouver, BC, Canada, May 2021.

\bibitem{PrakashP_IoTJ22}
P.~Prakash\emph{ et al.}, ``Io{T} device friendly and communication-efficient
  federated learning via joint model pruning and quantization,'' \emph{IEEE
  Internet Things J.}, vol.~9, no.~15, pp. 13\,638--13\,650, Aug. 2022.

\bibitem{Hinton_arXiv15}
\BIBentryALTinterwordspacing
G.~E. Hinton, O.~Vinyals, and J.~Dean, ``Distilling the knowledge in a neural
  network.'' [Online]. Available: \url{http://arxiv.org/abs/1503.02531}
\BIBentrySTDinterwordspacing

\bibitem{LiD_NISPwkshop19}
\BIBentryALTinterwordspacing
D.~Li and J.~Wang, ``{FedMD}: {H}eterogenous federated learning via model
  distillation.'' [Online]. Available:
  \url{https://arxiv.org/pdf/1910.03581.pdf}
\BIBentrySTDinterwordspacing

\bibitem{LiuL_ICC22}
L.~Liu, J.~Zhang, S.~H. Song, and K.~B. Letaief, ``Communication-efficient
  federated distillation with active data sampling,'' in \emph{Proc. IEEE Int.
  Conf. Commun. (ICC)}, Seoul, South Korea, May 2022.

\bibitem{wu2021fedkd}
C.~Wu, F.~Wu, L.~Lyu, Y.~Huang, and X.~Xie, ``Communication-efficient federated
  learning via knowledge distillation,'' \emph{Nat. Commun.}, vol.~13, no.~1,
  p. 2032, Apr. 2022.

\bibitem{ShaoJ_NatCom24}
J.~Shao, F.~Wu, and J.~Zhang, ``Selective knowledge sharing for
  privacy-preserving federated distillation without a good teacher,''
  \emph{Nat. Commun.}, vol.~15, pp. 1--11, Jan. 2024.

\bibitem{OhS_CL20}
S.~Oh, J.~Park, E.~Jeong, H.~Kim, M.~Bennis, and S.-L. Kim, ``{Mix2FLD}:
  Downlink federated learning after uplink federated distillation with two-way
  mixup,'' \emph{IEEE Commun. Lett.}, vol.~24, no.~10, pp. 2211--2215, Oct.
  2020.

\bibitem{MillsJ_arXiv23}
G.~M. Jed~Mills, Jia~Hu, ``Faster federated learning with decaying number of
  local {SGD} steps,'' \emph{IEEE Trans. Parallel Distr. Syst.}, vol.~34,
  no.~7, pp. 2198--2207, Jul. 2023.

\bibitem{Dinh_TON21}
C.~T. Dinh\emph{ et al.}, ``Federated learning over wireless networks:
  {C}onvergence analysis and resource allocation,'' \emph{{IEEE/ACM} Trans.
  Netw.}, vol.~29, no.~1, pp. 398--409, Feb. 2021.

\bibitem{WangS_JSAC19}
S.~Wang\emph{ et al.}, ``Adaptive federated learning in resource constrained
  edge computing systems,'' \emph{{IEEE J. Sel. Areas Commun.}}, vol.~37,
  no.~6, pp. 1205--1221, Jun. 2019.

\bibitem{Balles_UAI17}
L.~Balles, J.~Romero, and P.~Hennig, ``Coupling adaptive batch sizes with
  learning rates,'' in \emph{Proc. 33rd Conf. Uncertainty Artif. Intell.
  (UAI)}, Sydney, NSW, Australia, Aug. 2017.

\bibitem{MaZ_TMC22}
Z.~Ma, Y.~Xu, H.~Xu, Z.~Meng, L.~Huang, and Y.~Xue, ``Adaptive batch size for
  federated learning in resource-constrained edge computing,'' \emph{{IEEE
  Trans. Mob. Comput.}}, vol.~22, no.~1, pp. 37--53, Jan. 2023.

\bibitem{ZhanY_IPDPS20}
Y.~Zhan, P.~Li, and S.~Guo, ``Experience-driven computational resource
  allocation of federated learning by deep reinforcement learning,'' in
  \emph{Proc. {IEEE} Int. Parallel Distrib. Process. Symp. {(IPDPS)}}, New
  Orleans, LA, USA, May 2020.

\bibitem{YangZ_TWC21}
Z.~Yang, M.~Chen, W.~Saad, C.~S. Hong, and M.~S.-Bahaei, ``Energy efficient
  federated learning over wireless communication networks,'' \emph{{IEEE Trans.
  Wireless Commun.}}, vol.~20, no.~3, pp. 3579--3605, Mar. 2021.

\bibitem{MoX_JCIN21}
X.~Mo and J.~Xu, ``Energy-efficient federated edge learning with joint
  communication and computation design,'' \emph{{J. Commun. Inf. Netw.}},
  vol.~6, no.~2, pp. 110--124, Jun. 2021.

\bibitem{ZengQ_TWC21}
Q.~Zeng, Y.~Du, K.~Huang, and K.~K. Leung, ``Energy-efficient resource
  management for federated edge learning with {CPU-GPU} heterogeneous
  computing,'' \emph{IEEE Trans. Wireless Commun.}, vol.~20, no.~12, pp.
  7947--7962, Dec. 2021.

\bibitem{ZhangT_TGCN22}
T.~Zhang and S.~Mao, ``Energy-efficient federated learning with intelligent
  reflecting surface,'' \emph{IEEE Trans. Green Commun. Netw.}, vol.~6, no.~2,
  pp. 845--858, Jun. 2022.

\bibitem{Nishio_ICC19}
T.~Nishio and R.~Yonetani, ``Client selection for federated learning with
  heterogeneous resources in mobile edge,'' in \emph{Proc. IEEE Int. Conf.
  Commun. (ICC)}, Shanghai, China, May 2019.

\bibitem{LiL_RTSS19}
L.~Li, H.~Xiong, Z.~Guo, J.~Wang, and C.-Z. Xu, ``{SmartPC}: Hierarchical pace
  control in real-time federated learning system,'' in \emph{Proc. IEEE
  Real-Time Syst. Symp. (RTSS)}, Hong Kong, China, Dec. 2019.

\bibitem{HuY_WCSP20}
Y.~Hu, H.~Huang, and N.~Yu, ``Device scheduling for energy-efficient federated
  learning over wireless network based on {TDMA} mode,'' in \emph{Proc. Int.
  Conf. Wireless Commun. Signal Process. (WCSP)}, Nanjing, China, Oct. 2020.

\bibitem{YuL_IoTJ22}
L.~Yu, R.~Albelaihi, X.~Sun, N.~Ansari, and M.~Devetsikiotis, ``Jointly
  optimizing client selection and resource management in wireless federated
  learning for {I}nternet of {T}hings,'' \emph{IEEE Internet Things J.},
  vol.~9, no.~6, pp. 4385--4395, Mar. 2022.

\bibitem{RenJ_TWC20}
J.~Ren, Y.~He, D.~Wen, G.~Yu, K.~Huang, and D.~Guo, ``Scheduling for cellular
  federated edge learning with importance and channel awareness,'' \emph{IEEE
  Trans. Wireless Commun.}, vol.~19, no.~11, pp. 7690–--7703, Nov. 2020.

\bibitem{SunY_TWC23}
Y.~Sun, Z.~Lin, Y.~Mao, S.~Jin, and J.~Zhang, ``Channel and gradient-importance
  aware device scheduling for over-the-air federated learning,'' \emph{IEEE
  Trans. Wireless Commun.}, vol.~23, no.~7, pp. 6905--6920, Jul. 2024.

\bibitem{XuJ_TWC2021}
J.~Xu and H.~Wang, ``Client selection and bandwidth allocation in wireless
  federated learning networks: A long-term perspective,'' \emph{IEEE Trans.
  Wireless Commun.}, vol.~20, no.~2, pp. 1188--1200, Feb. 2021.

\bibitem{GuoK_JSTSP22}
K.~Guo, Z.~Chen, H.~H. Yang, and T.~Q.~S. Quek, ``Dynamic scheduling for
  heterogeneous federated learning in private {5G} edge networks,'' \emph{IEEE
  J. Sel. Topics Signal Process.}, vol.~16, no.~1, pp. 26--40, Jan. 2022.

\bibitem{BattiloroC_TGCN22}
C.~Battiloro, P.~D. Lorenzo, M.~Merluzzi, and S.~Barbarossa, ``Lyapunov-based
  optimization of edge resources for energy-efficient adaptive federated
  learning,'' \emph{IEEE Trans. Green Commun. Netw.}, vol.~7, no.~1, pp.
  265--280, Mar. 2023.

\bibitem{KimYG_MICRO21}
Y.~G. Kim and C.-J. Wu, ``Auto{FL}: {E}nabling heterogeneity-aware energy
  efficient federated learning,'' in \emph{Proc. IEEE/ACM Int. Symp.
  Microarchit. (MICRO)}, Athens, Greece, Oct. 2021.

\bibitem{Katharopoulos_ICML18}
A.~Katharopoulos and F.~Fleuret, ``Not all samples are created equal: {D}eep
  learning with importance sampling,'' in \emph{Proc. Int. Conf. Mach. Learn.
  (ICML)}, Stockholm, Sweden, Jul. 2018.

\bibitem{XiaoY_WCSP20}
Y.~Xiao, Y.~Li, G.~Shi, and H.~V. Poor, ``Optimizing resource-efficiency for
  federated edge intelligence in {IoT} networks,'' in \emph{Proc. Int. Conf.
  Wireless Commun. Signal Process. (WCSP)}, Nanjing, China, Oct. 2020.

\bibitem{HeY_TVT20}
Y.~He, J.~Ren, G.~Yu, and J.~Yuan, ``Importance-aware data selection and
  resource allocation in federated edge learning system,'' \emph{IEEE Trans.
  Veh. Techn.}, vol.~69, no.~11, pp. 13\,593--13\,605, Nov. 2020.

\bibitem{Albaseer_TNSE21}
A.~M. Albaseer, M.~Abdallah, A.~A.-Fuqaha, and A.~Erbad, ``Fine-grained data
  selection for improved energy efficiency of federated edge learning,''
  \emph{{IEEE Trans. Netw. Sci. Eng.}}, vol.~9, no.~5, pp. 3258--3271,
  Sep.-Oct. 2022.

\bibitem{LiA_INFOCOM21}
A.~Li, L.~Zhang, J.~Tan, Y.~Qin, J.~Wang, and X.-Y. Li, ``Sample-level data
  selection for federated learning,'' in \emph{Proc. IEEE Int. Conf. Comput.
  Commun. (INFOCOM)}, Virtual Event, May 2021.

\bibitem{YangW_WCSP20}
W.~Yang, Y.~Zhang, W.~Y.~B.~Lim, Z.~Xiong, Y.~Jiao, and J.~Jin, ``Privacy is
  not free: Energy-aware federated learning for mobile and edge intelligence,''
  in \emph{Proc. Int. Conf. Wireless Commun. Signal Process. (WCSP)}, Nanjing,
  China, Oct. 2020.

\bibitem{JiZ_TVT21}
Z.~Ji, L.~Chen, N.~Zhao, Y.~Chen, G.~Wei, and F.~R. Yu, ``Computation
  offloading for edge-assisted federated learning,'' \emph{{IEEE} Trans. Veh.
  Techn.}, vol.~70, no.~9, pp. 9330--9334, Sep. 2021.

\bibitem{Wutyee_ACCESS21}
C.~W. Zaw, S.~R. Pandey, K.~Kim, and C.~S. Hong, ``Energy-aware resource
  management for federated learning in multi-access edge computing systems,''
  \emph{IEEE Access}, vol.~9, pp. 34\,938--34\,950, Jan. 2021.

\bibitem{CaiX_CJIoT19}
X.~Cai, X.~Mo, and J.~Xu, ``{D2D} computation task offloading for efficient
  federated learning,'' \emph{Chinese J. Internet Things}, vol.~3, no.~4, pp.
  82--90, Dec. 2019.

\bibitem{WangS_INFOCOM21}
S.~Wang, M.~Lee, S.~Hosseinalipour, R.~Morabito, M.~Chiang, and C.~G. Brinton,
  ``Device sampling for heterogeneous federated learning: Theory, algorithms,
  and implementation,'' in \emph{Proc. IEEE Conf. Comput. Commun. (INFOCOM)},
  Vancouver, BC, Canada, May 2021.

\bibitem{Shullary_MWSCAS22}
M.~H. Shullary, A.~A. Abdellatif, and Y.~Massoudn, ``Energy-efficient active
  federated learning on non-iid data,'' in \emph{Proc. IEEE Int. Midwest Symp.
  Circuits Syst. (MWSCAS)}, Fukuoka, Japan, Aug. 2022.

\bibitem{Prakash_JSAC21}
S.~Prakash\emph{ et al.}, ``Coded computing for low-latency federated learning
  over wireless edge networks,'' \emph{IEEE J. Sel. Areas Commun.}, vol.~39,
  no.~1, pp. 233--250, Jan. 2021.

\bibitem{SunY_ISIT22}
Y.~Sun, J.~Shao, S.~Li, Y.~Mao, and J.~Zhang, ``Stochastic coded federated
  learning with convergence and privacy guarantees,'' in \emph{Proc. IEEE Int.
  Symp. Inf. Theory (ISIT)}, Espoo, Finland, Jul. 2022.

\bibitem{ShaoJ_NeurIPS22}
J.~Shao, Y.~Sun, S.~Li, and J.~Zhang, ``{DReS-FL: D}ropout-resilient secure
  federated learning for non-iid clients via secret data sharing,'' in
  \emph{Proc. 36th Conf. Neural Inf. Process. Syst. (NeurIPS)}, New Orleans,
  LA, USA, Nov.-Dec. 2022.

\bibitem{MaT_IoTJ23}
T.~Ma, H.~Wang, and C.~Li, ``Quantized distributed federated learning for
  industrial {I}nternet of {T}hings,'' \emph{IEEE Internet Things J.}, vol.~10,
  no.~4, pp. 3027--3036, Feb. 2023.

\bibitem{ZhangX_ICASSP24}
X.~Zhang, C.-C. Chiu, and T.~He, ``Energy-efficient decentralized learning via
  graph sparsification,'' in \emph{Proc. IEEE Int. Conf. Acoustics, Speech
  Signal Process. (ICASSP)}, Seoul, Korea, Apr. 2024.

\bibitem{LiuS_TWC23}
S.~Liu, G.~Yu, D.~Wen, X.~Chen, M.~Bennis, and H.~Chen, ``Communication and
  energy efficient decentralized learning over {D2D} networks,'' \emph{IEEE
  Trans. Wireless Commun.}, vol.~22, no.~10, pp. 9549--9563, Dec. 2023.

\bibitem{JiangZ_TPDS23}
Z.~Jiang, Y.~Xu, H.~Xu, L.~Wang, C.~Qiao, and L.~Huang, ``Joint model pruning
  and topology construction for accelerating decentralized machine learning,''
  \emph{IEEE Trans. Parallel Distrib. Syst.}, vol.~34, no.~10, pp. 2827--2842,
  Oct. 2023.

\bibitem{SunH_ICASSP24}
H.~Sun, H.~Tian, W.~Ni, and J.~Zheng, ``On the convergence of hierarchical
  federated learning with gradient quantization and imperfect transmission,''
  in \emph{Proc. IEEE Int. Conf. Acoustics, Speech Signal Process. (ICASSP)},
  Seoul, Korea, Apr. 2024.

\bibitem{LiuX_TWC24}
X.~Liu, S.~Wang, and Y.~Deng, ``Adaptive federated pruning in hierarchical
  wireless networks,'' \emph{IEEE Trans. Wireless Commun.}, vol.~23, no.~6, pp.
  5985--5999, Oct. 2024.

\bibitem{LuoS_TWC20}
S.~Luo, X.~Chen, Q.~Wu, Z.~Zhou, and S.~Yu, ``{HFEL}: {J}oint edge association
  and resource allocation for cost-efficient hierarchical federated edge
  learning,'' \emph{IEEE Trans. Wireless Commun.}, vol.~19, no.~10, pp.
  6535--6548, Oct. 2020.

\bibitem{Mhaisen_TNSE22}
N.~Mhaisen, A.~A. Abdellatif, A.~Mohamed, A.~Erbad, and M.~Guizani, ``Optimal
  user-edge assignment in hierarchical federated learning based on statistical
  properties and network topology constraints,'' \emph{IEEE Trans. Netw. Sci.
  Eng.}, vol.~9, no.~1, pp. 55--66, Jan.-Feb. 2022.

\bibitem{LangN_TSP23}
N.~Lang, E.~Sofer, T.~Shaked, and N.~Shlezinger, ``Joint privacy enhancement
  and quantization in federated learning,'' \emph{IEEE Trans. Signal Process.},
  vol.~77, pp. 295--310, 2023.

\bibitem{SunY_TMC23}
Y.~Sun, Y.~Mao, and J.~Zhang, ``{MimiC}: {C}ombating client dropouts in
  federated learning by mimicking central updates,'' \emph{IEEE Trans. Mob.
  Comput.}, vol.~23, no.~7, pp. 7572--7584, Jul. 2024.

\bibitem{LinZhen_TWC22}
Z.~Lin, X.~Li, V.~K.~N. Lau, Y.~Gong, and K.~Huang, ``Deploying federated
  learning in large-scale cellular networks: {S}patial convergence analysis,''
  \emph{IEEE Trans. Wireless Commun.}, vol.~21, no.~3, pp. 1542--1556, Mar.
  2022.

\bibitem{LeeJ_abs-1907-01989}
\BIBentryALTinterwordspacing
J.~Lee\emph{ et al.}, ``On-device neural net inference with mobile {GPU}s,''
  2019. [Online]. Available: \url{http://arxiv.org/abs/1907.01989}
\BIBentrySTDinterwordspacing

\bibitem{Howard_MobileNet17}
\BIBentryALTinterwordspacing
A.~G. Howard\emph{ et al.}, ``{MobileNets}: Efficient convolutional neural
  networks for mobile vision applications.'' [Online]. Available:
  \url{http://arxiv.org/abs/1704.04861}
\BIBentrySTDinterwordspacing

\bibitem{NiuW_IJCAI20}
W.~Niu, P.~Zhao, Z.~Zhan, X.~Lin, Y.~Wang, and B.~Ren, ``Towards real-time
  {DNN} inference on mobile platforms with model pruning and compiler
  optimization,'' in \emph{Proc. 29th Int. Joint Conf. Artif. Intell.
  ({IJCAI})}, Yokohama, Japan, Jan. 2020.

\bibitem{GuoP_NSDI21}
P.~Guo, B.~Hu, and W.~Hu, ``Mistify: Automating {DNN} model porting for
  on-device inference at the edge,'' in \emph{Proc. 18th {USENIX} Symp. Netw.
  Syst. Design Implement. ({NSDI})}, Virtual Event, Apr. 2021.

\bibitem{Hashemi_DATE17}
S.~Hashemi, N.~Anthony, H.~Tann, R.~I. Bahar, and S.~Reda, ``Understanding the
  impact of precision quantization on the accuracy and energy of neural
  networks,'' in \emph{Proc. Design, Autom. Test Eur. Conf. Exhib. {(DATE)}},
  Lausanne, Switzerland, Mar. 2017.

\bibitem{Gholami_abs-2103-13630}
\BIBentryALTinterwordspacing
A.~Gholami, S.~Kim, Z.~Dong, Z.~Yao, M.~W. Mahoney, and K.~Keutzer, ``A survey
  of quantization methods for efficient neural network inference.'' [Online].
  Available: \url{https://arxiv.org/abs/2103.13630}
\BIBentrySTDinterwordspacing

\bibitem{ZhangD_ECCV18}
D.~Zhang, J.~Yang, D.~Ye, and G.~Hua, ``{LQ-N}ets: {L}earned quantization for
  highly accurate and compact deep neural networks,'' in \emph{{P}roc. Eur.
  Conf. Comput. Vision ({ECCV})}, Munich, Germany, Sep. 2018.

\bibitem{Banner_NIPS19}
R.~Banner, Y.~Nahshan, and D.~Soudry, ``Post training 4-bit quantization of
  convolutional networks for rapid-deployment,'' in \emph{Proc. 33rd Conf.
  Neural Inf. Proc. Syst. (NeurISP)}, Vancouver, BC, Canada, Dec. 2019.

\bibitem{WangK_CVPR19}
K.~Wang, Z.~Liu, Y.~Lin, J.~Lin, and S.~Han, ``{HAQ:} {H}ardware-aware
  automated quantization with mixed precision,'' in \emph{{P}roc. {IEEE} Conf.
  Comput. Vision Pattern Recogn. ({CVPR})}, Long Beach, CA, USA, Jun. 2019.

\bibitem{LiangT_NeuCom21}
T.~Liang, J.~Glossner, L.~Wang, S.~Shi, and X.~Zhang, ``Pruning and
  quantization for deep neural network acceleration: A survey,''
  \emph{Neurocomput.}, vol. 461, pp. 370--403, Oct. 2021.

\bibitem{YangT_CVPR17}
T.-J. Yang, Y.-H. Chen, and V.~Sze, ``Designing energy-efficient convolutional
  neural networks using energy-aware pruning,'' in \emph{Proc. {IEEE} Conf.
  Comput. Vision Pattern Recogn. ({CVPR})}, Honolulu, HI, USA, Jul. 2017.

\bibitem{Kung_Computer82}
H.~Kung, ``Why systolic architectures?'' \emph{Comput.}, vol.~15, no.~1, pp.
  37--46, Jan. 1982.

\bibitem{YangH_ICLR19}
H.~Yang, Y.~Zhu, and J.~Liu, ``Energy-constrained compression for deep neural
  networks via weighted sparse projection and layer input masking,'' in
  \emph{Proc. Int. Conf. Learn. Repr. ({ICLR})}, New Orleans, LA, USA, May
  2019.

\bibitem{ZhiX_arXiv23}
\BIBentryALTinterwordspacing
X.~Zhi, V.~Babbar, P.~Sun, F.~Silavong, R.~Shi, and S.~Moran, ``Lightweight
  parameter pruning for energy-efficient deep learning: A binarized gating
  module approach.'' [Online]. Available:
  \url{https://arxiv.org/pdf/2302.10798v2.pdf}
\BIBentrySTDinterwordspacing

\bibitem{GouJ_IJCV21}
J.~Gou, B.~Yu, S.~J. Maybank, and D.~Tao, ``Knowledge distillation: {A}
  survey,'' \emph{{S}pringer {I}nt. {J}. {C}omput. {V}ision}, vol. 129, pp.
  1789--1819, Mar. 2021.

\bibitem{Romero_ICLR15}
A.~Romero, N.~Ballas, S.~E. Kahou, A.~Chassang, C.~Gatta, and Y.~Bengio,
  ``Fitnets: {H}ints for thin deep nets,'' in \emph{Proc. {I}nt. {C}onf. Learn.
  Repr. ({ICLR})}, San Diego, CA, USA, May 2015.

\bibitem{ParkW_CVPR19}
W.~Park, D.~Kim, Y.~Lu, and M.~Cho, ``Relational knowledge distillation,'' in
  \emph{Proc. {IEEE} Conf. Comput. Vision Pattern Recogn. ({CVPR})}, Long
  Beach, CA, USA, Jun. 2019.

\bibitem{Dennis_arXiv23}
D.~K. Dennis, A.~Shetty, A.~Sevekari, K.~Koishida, and V.~Smith, ``Progressive
  ensemble distillation: {B}uilding ensembles for efficient inference,'' in
  \emph{Proc. 37th Conf. Neural Inf. Process. Syst. (NeurIPS)}, New Orleans,
  LA, USA, Dec. 2023.

\bibitem{RenP_ACMCOMPSurvey2021}
P.~Ren\emph{ et al.}, ``A comprehensive survey of neural architecture search:
  Challenges and solutions,'' \emph{{ACM} {C}omput. {S}urveys}, vol.~54, no.~4,
  pp. 1--34, May 2021.

\bibitem{WangD_EMC2_19}
D.~Wang, M.~Li, L.~Wu, V.~Chandra, and Q.~Liu, ``Energy-aware neural
  architecture optimization with splitting steepest descent,'' in \emph{Proc.
  5th Wkshop. Energy Efficient Mach. Learn. Cogn. Comput. ({EMC}$^{2}$)},
  Vancouver, BC, Canada, Dec. 2019.

\bibitem{Benmeziane_arXiv210109336}
\BIBentryALTinterwordspacing
H.~Benmeziane, K.~E. Maghraoui, H.~Ouarnoughi, S.~Niar, M.~Wistuba, and
  N.~Wang, ``A comprehensive survey on hardware-aware neural architecture
  search.'' [Online]. Available: \url{https://arxiv.org/pdf/2101.09336.pdf}
\BIBentrySTDinterwordspacing

\bibitem{Hsu_abs-1806-10332}
\BIBentryALTinterwordspacing
C.~Hsu\emph{ et al.}, ``{MONAS:} {M}ulti-objective neural architecture search
  using reinforcement learning.'' [Online]. Available:
  \url{http://arxiv.org/abs/1806.10332}
\BIBentrySTDinterwordspacing

\bibitem{Srinivas_abs-1906_07214}
\BIBentryALTinterwordspacing
S.~V.~K. Srinivas, H.~Nair, and V.~Vidyasagar, ``Hardware aware neural network
  architectures using {F}b{N}et.'' [Online]. Available:
  \url{https://arxiv.org/abs/1906.07214}
\BIBentrySTDinterwordspacing

\bibitem{Speckhard_NCA23}
D.~T. Speckhard, K.~Misiunas, S.~Perel, T.~Zhu, S.~Carlile, and M.~Slaney,
  ``Neural architecture search for energy efficient always-on audio machine
  learning,'' \emph{Neural Comput. Appl.}, vol.~35, no.~16, p. 12133–12144,
  Jun. 2023.

\bibitem{GongC_ICCAD19}
C.~Gong, Z.~Jiang, D.~Wang, Y.~Lin, Q.~Liu, and D.~Z. Pan, ``Mixed precision
  neural architecture search for energy efficient deep learning,'' in
  \emph{Proc. IEEE/ACM Int. Conf. Comput.-Aided Design (ICCAD)}, Westminster,
  CO, USA, Nov. 2019.

\bibitem{Sandler_MobileNetV2}
\BIBentryALTinterwordspacing
M.~Sandler, A.~Howard, M.~Zhu, A.~Zhmoginov, and L.-C. Chen, ``{MobileNetV2}:
  Inverted residuals and linear bottlenecks.'' [Online]. Available:
  \url{https://arxiv.org/pdf/1801.04381.pdf}
\BIBentrySTDinterwordspacing

\bibitem{Maleki_TDAES21}
M.-A. Maleki, A.~N.-Meybodi, M.~Kamal, A.~A.-Kusha, and M.~Pedram, ``An
  energy-efficient inference method in convolutional neural networks based on
  dynamic adjustment of the pruning level,'' \emph{{ACM} Trans. Design Autom.
  Electron. Syst.}, vol.~26, no.~6, pp. 1--20, Jul. 2021.

\bibitem{ZhouYH_IJCAI18}
J.~Guan, Y.~Liu, Q.~Liu, and J.~Peng, ``Energy-efficient amortized inference
  with cascaded deep classifiers,'' in \emph{Proc. 29th Int. Joint Conf. Artif.
  Intell. ({IJCAI})}, Stockholm, Sweden, Jul. 2018.

\bibitem{Teerapittayanon_ICPR16}
S.~Teerapittayanon, B.~McDanel, and H.~Kunge, ``Branchy{N}et: {F}ast inference
  via early exiting from deep neural networks,'' in \emph{Proc. Conf. Pattern
  Recogn. ({ICPR})}, Cancun, Mexico, Dec. 2016.

\bibitem{WangX_ECCV18}
X.~Wang, F.~Yu, Z.-Y. Dou, T.~Darrell, and J.~E. Gonzalez, ``{SkipNet}:
  {L}earning dynamic routing in convolutional networks,'' in \emph{{P}roc. Eur.
  Conf. Comput. Vision ({ECCV})}, Munich, Germany, Sep. 2018.

\bibitem{Panda_DATE16}
P.~Panda, A.~Sengupta, and K.~Roy, ``Conditional deep learning for
  energy-efficient and enhanced pattern recognition,'' in \emph{Proc. Design,
  Autom. Test Eur. Conf. Exhib. {(DATE)}}, Dresden, Germany, Mar. 2016.

\bibitem{Laskaridis_EMDL21}
S.~Laskaridis, A.~Kouris, and N.~D. Lane, ``Adaptive inference through
  early-exit networks: Design, challenges and directions,'' in \emph{Proc. Int.
  Wkshop. Embedded Mob. Deep Learn. {(EMDL)}}, Virtual Event, Jun. 2021.

\bibitem{HanY_TPAMI21}
Y.~Han, G.~Huang, S.~Song, L.~Yang, H.~Wang, and Y.~Wang, ``Dynamic neural
  networks: {A} survey,'' \emph{{IEEE} Trans. Pattern Anal. Mach. Intell.},
  vol.~44, no.~11, pp. 7436--7456, Nov. 2022.

\bibitem{LeeRoyson_MobiCom19}
J.~Guan, Y.~Liu, Q.~Liu, and J.~Peng, ``Mobi{SR}: {E}fficient on-device
  super-resolution through heterogeneous mobile processors,'' in \emph{Proc.
  Annu. Int. Conf. Mob. Comput. Netw. (MobiCom)}, Los Cabos, Mexico, Oct. 2019.

\bibitem{MarcoSanz_TECS2020}
V.~S. Marco, B.~Taylor, Z.~Wang, and Y.~Elkhatib, ``Optimizing deep learning
  inference on embedded systems through adaptive model selection,'' \emph{{ACM}
  {T}rans. {E}mbedded {C}omput. {S}yst.}, vol.~19, no.~1, pp. 1--28, Feb. 2020.

\bibitem{FangBiyi_SEC20}
B.~Fang, X.~Zeng, F.~Zhang, H.~Xu, and M.~Zhang, ``Flex{DNN}: {I}nput-adaptive
  on-device deep learning for efficient mobile vision,'' in \emph{ACM/IEEE
  Symp. Edge Comput. (SEC)}, San Jose, CA, USA, Nov. 2020.

\bibitem{WuZuxuan_CVPR18}
Z.~Wu, T.~Nagarajan, A.~Kumar, and S.~Rennie, ``Block{D}rop: {D}ynamic
  inference paths in residual networks,'' in \emph{Proc. {IEEE} Conf. Comput.
  Vision Pattern Recogn. ({CVPR})}, Salt Lake City, UT, USA, Jun. 2018.

\bibitem{HuaWeizhe_NeurSIP19}
W.~Hua, Y.~Zhou, C.~D. Sa, Z.~Zhang, and G.~Suh, ``Channel gating neural
  networks,'' in \emph{Proc. 33rd Conf. Neural Inf. Process. Syst. (NeurIPS)},
  Los Cabos, Mexico, Oct. 2019.

\bibitem{WangYue_JSTSP20}
Y.~Wang\emph{ et al.}, ``Dual dynamic inference: Enabling more efficient,
  adaptive, and controllable deep inference,'' \emph{IEEE J. Sel. Topics Signal
  Process.}, vol.~14, no.~4, pp. 623--663, May 2020.

\bibitem{Pavel_COMST17}
P.~Mach and Z.~Becvar, ``Mobile edge computing: {A} survey on architecture and
  computation offloading,'' \emph{{IEEE} Commun. Survey Tuts.}, vol.~19, no.~3,
  pp. 1628--1656, Third Quart. 2017.

\bibitem{ZhangSQ_COMST17}
S.~Zhang, Q.~Wu, S.~Xu, and G.~Y. Li, ``Fundamental green tradeoffs:
  {P}rogresses, challenges, and impacts on {5G} networks,'' \emph{IEEE Commun.
  Surveys Tuts.}, vol.~19, no.~1, pp. 33--56, First Quart. 2017.

\bibitem{SunWen_IEEENetw19}
W.~Sun, J.~Liu, and Y.~Yue, ``{AI}-enhanced offloading in edge computing: When
  machine learning meet industrial {I}o{T},'' \emph{IEEE Netw.}, vol.~33,
  no.~5, pp. 68--74, Sep./Oct. 2019.

\bibitem{YangBo_IoTJ20}
B.~Yang, X.~Cao, X.~Li, Q.~Zhang, and L.~Qian, ``Mobile-edge-computing-based
  hierarchical machine learning tasks distribution for {II}o{T},'' \emph{IEEE
  Internet Things J.}, vol.~7, no.~3, pp. 2169--2180, Mar. 2020.

\bibitem{ZhangWT_TVT21}
W.~Zhang\emph{ et al.}, ``Deep reinforcement learning based resource management
  for {DNN} inference in industrial {IoT},'' \emph{IEEE Trans. Veh. Techn.},
  vol.~70, no.~8, pp. 7605--7618, Aug. 2021.

\bibitem{MaH_IoTJ23}
H.~Ma, Z.~Zhou, X.~Zhang, and X.~Chen, ``Toward carbon-neutral edge computing:
  {G}reening edge {AI} by harnessing spot and future carbon markets,''
  \emph{IEEE Internet Things J.}, vol.~10, no.~18, pp. 16\,637--16\,649, Sep.
  2023.

\bibitem{Redmon_YOLO15}
\BIBentryALTinterwordspacing
J.~Redmon, S.~Divvala, and R.~G.~A. Farhadi, ``You only look once: {U}nified,
  real-time object detection.'' [Online]. Available:
  \url{https://arxiv.org/pdf/1506.02640.pdf}
\BIBentrySTDinterwordspacing

\bibitem{Redmon_Yolov2}
\BIBentryALTinterwordspacing
J.~Redmon and A.~Farhadi, ``{YOLO9000:} {B}etter, faster, stronger.'' [Online].
  Available: \url{https://arxiv.org/pdf/1612.08242}
\BIBentrySTDinterwordspacing

\bibitem{RanXukan_INFOCOM18}
X.~Ran, H.~Chen, X.~Zhu, Z.~Liu, and J.~Chen, ``Deep{D}ecision: {A} mobile deep
  learning framework for edge video analytics,'' in \emph{Proc. IEEE Int. Conf.
  Comput. Commun. (INFOCOM)}, Honolulu, HI, USA, Apr. 2018.

\bibitem{XuD_CL19}
D.~Xu, Q.~Li, and H.~Zhu, ``Energy-saving computation offloading by joint data
  compression and resource allocation for mobile-edge computing,'' \emph{{IEEE}
  Commun. {L}ett.}, vol.~23, no.~4, pp. 704--707, Apr. 2019.

\bibitem{WuWen_TII21}
W.~Wu, P.~Yang, W.~Zhang, C.~Zhou, and X.~Shen, ``Accuracy-guaranteed
  collaborative {DNN} inference in industrial {IoT} via deep reinforcement
  learning,'' \emph{IEEE Trans. Ind. Inform.}, vol.~17, no.~7, pp. 4988--4998,
  Jul. 2021.

\bibitem{HuangXiufeng_IOTJ20}
X.~Huang and S.~Zhou, ``Dynamic compression ratio selection for edge inference
  systems with hard deadlines,'' \emph{IEEE Internet Things J.}, vol.~17,
  no.~7, pp. 8800--8810, Sep. 2020.

\bibitem{WangCan_INFOCOM20}
C.~Wang, S.~Zhang, Y.~Chen, Z.~Qian, J.~Wu, and M.~Xiao, ``Joint configuration
  adaptation and bandwidth allocation for edge-based real-time video
  analytics,'' in \emph{Proc. IEEE Int. Conf. Comput. Commun. (INFOCOM)},
  Toronto, ON, Canada, Jul. 2020.

\bibitem{HeZhaoliang_TMCCA21}
Z.~He, H.~Li, Z.~Wang, S.~Xia, and W.~Zhu, ``Adaptive compression for online
  computer vision: {A}n edge reinforcement learning approach,'' \emph{ACM
  Trans. Multimedia Comput. Commun. Appl.}, vol.~17, no.~4, p. 118, Nov. 2021.

\bibitem{Galanopoulos_INFOCOM21}
A.~Galanopoulos, J.~A. A.-Romero, D.~J. Leith, and G.~Iosifidis, ``Auto{ML} for
  video analytics with edge computing,'' in \emph{Proc. IEEE Int. Conf. Comput.
  Commun. (INFOCOM)}, Vancouver, BC, Canada, May 2021.

\bibitem{FanW_TII23}
W.~Fan, Z.~Chen, Z.~Hao, Y.~Su, F.~Wu, B.~Tang, and Y.~Liu, ``{DNN} deployment,
  task offloading, and resource allocation for joint task inference in
  {IIoT},'' \emph{IEEE Trans. Ind. Informat.}, vol.~19, no.~2, pp. 1634--1646,
  Feb. 2023.

\bibitem{KangY_Neurosurgeon17}
Y.~Kang\emph{ et al.}, ``Neurosurgeon: {C}ollaborative intelligence between the
  cloud and mobile edge,'' \emph{{ACM SIGARCH} Comput. Archit. News}, vol.~45,
  no.~1, pp. 615--629, Dec. 2017.

\bibitem{LiEn_TWC20}
E.~Li, L.~Zeng, Z.~Zhou, and X.~Chen, ``Edge {AI}: {O}n-demand accelerating
  deep neural network inference via edge computing,'' \emph{{IEEE} Trans.
  Wireless Commun.}, vol.~19, no.~1, pp. 447--457, Jan. 2020.

\bibitem{Laskaridis_MOBICOM20}
S.~Laskaridis, S.~I. Venieris, M.~Almeida, I.~Leontiadis, and N.~D. Lane,
  ``{SPINN}: {S}ynergistic progressive inference of neural networks over device
  and cloud,'' in \emph{Proc. 26th Annu. Int. Conf. Mob. Comput. Netw.
  (MobiCom)}, London, UK, Sep. 2020.

\bibitem{ZhaoZ_IoTDI21}
Z.~Zhao, K.~Wang, N.~Ling, and G.~Xing, ``Edge{ML}: An {A}uto{ML} framework for
  real-time deep learning on the edge,'' in \emph{{Proc}. Int. Conf.
  Internet-of-Things Design Implement. ({I}o{TDI})}, Virtual Event, May 2021.

\bibitem{LiHS_ICPADS18}
H.~Li, C.~Hu, J.~Jiang, Z.~Wang, Y.~Wen, and W.~Zhu, ``{JALAD}: {J}oint
  accuracy-and latency-aware deep structure decoupling for edge-cloud
  execution,'' in \emph{{Proc}. {IEEE} Int. Conf. Parallel Distrib. Syst.
  (ICPADS)}, Singapore, Dec. 2018.

\bibitem{Eshratifar_ISLPED19}
A.~E. Eshratifar, A.~Esmaili, and M.~Pedram, ``Bottle{N}et: {A} deep learning
  architecture for intelligent mobile cloud computing services,'' in
  \emph{{Proc}. {IEEE/ACM} Int. Symp. Low Power Electron. Design (ISLPED)},
  Lausanne, Switzerland, Jul. 2019.

\bibitem{ChenZhuo_TIP19}
Z.~Chen, K.~Fan, S.~Wang, L.~Duan, W.~Lin, and A.~C. Kot, ``Toward intelligent
  sensing: {I}ntermediate deep feature compression,'' \emph{IEEE Trans. Image
  Process.}, vol.~29, pp. 2230--2243, 2019.

\bibitem{ShaoJW_ICCW19}
J.~Shao and J.~Zhang, ``Bottle{N}et++: {An} an end-to-end approach for feature
  compression in device-edge co-inference systems,'' in \emph{{Proc}. {IEEE}
  Int. Conf. Commun. Wkshop. (ICCW)}, Virtual Event, Jun. 2020.

\bibitem{Krouka_PIMRC21}
M.~Krouka, A.~Elgabli, C.~B. Issaid, and M.~Bennis, ``Energy-efficient model
  compression and splitting for collaborative inference over time-varying
  channels,'' in \emph{Proc. IEEE Annu. Symp. Personal, Indoor Mob. Radio
  Commun. (PIMRC)}, Virtual Event, Sep. 2021.

\bibitem{ShiW_INFOCOMWKSHOP}
W.~Shi, Y.~Hou, S.~Zhou, Z.~Niu, Y.~Zhang, and L.~Geng, ``Improving device-edge
  cooperative inference of deep learning via 2-step pruning,'' in \emph{{Proc}.
  {IEEE} Int. Conf. Comput. Commun. (INFOCOM) Wkshop.}, Paris, France, Sep.
  2019.

\bibitem{ZhangX_GLOBECOM21}
X.~Zhang, J.~Shao, Y.~Mao, and J.~Zhang, ``Communication-computation efficient
  device-edge co-inference via {AutoML},'' in \emph{Proc. Global Commun. Conf.
  ({GLOBECOM})}, Madrid, Spain, Dec. 2021.

\bibitem{ShaoJ_ICASSP20}
J.~Shao, H.~Zhang, Y.~Mao, and J.~Zhang, ``Branchy-{GNN}: {A} device-edge
  co-inference framework for efficient point cloud processing,'' in
  \emph{{Proc}. {IEEE} Conf. Acoust., Speech, Signal Process. ({ICASSP})},
  Toronto, ON, Canada, Apr. 2020.

\bibitem{DongR_JCIN22}
R.~Dong, Y.~Mao, and J.~Zhang, ``Resource-constrained edge {AI} with early exit
  prediction,'' \emph{J. Commun. Inf. Netw.}, vol.~7, no.~2, pp. 122--134, Jun.
  2022.

\bibitem{HuDiyi_IoTDI20}
D.~Hu and B.~Krishnamachari, ``Fast and accurate streaming {CNN} inference via
  communication compression on the edge,'' in \emph{{Proc}. {ACM/IEEE} Int.
  Conf. Internet-of-Things Design Implement. ({I}o{TDI})}, Sydney, NSW,
  Australia, Apr. 2020.

\bibitem{LanQ_TWC23}
Q.~Lan, Q.~Zeng, P.~Popovski, D.~Gündüz, and K.~Huang, ``Progressive feature
  transmission for split classification at the wireless edge,'' \emph{IEEE
  Trans. Wireless Commun.}, vol.~22, no.~6, pp. 3837--3852, Jun. 2023.

\bibitem{ShaoJ_JSAC22}
J.~Shao, Y.~Mao, and J.~Zhang, ``Learning task-oriented communication for edge
  inference: {A}n information bottleneck approach,'' \emph{IEEE J. Sel. Areas
  Commun.}, vol.~40, no.~1, pp. 197--211, Jan. 2022.

\bibitem{NTishby_Allerton99}
N.~Tishby, F.~C. Pereira, and W.~Bialek, ``The information bottleneck method,''
  in \emph{Proc. Annu. Allerton Conf. Commun. Control Comput.}, Monticello, IL,
  USA, Oct. 1999.

\bibitem{LieK_TON21}
L.~Zeng, X.~Chen, Z.~Zhou, L.~Yang, and J.~Zhang, ``Co{E}dge: {C}ooperative
  {DNN} inference with adaptive workload partitioning over heterogeneous edge
  devices,'' \emph{{IEEE/ACM} Trans. Netw.}, vol.~29, no.~2, pp. 595--608, Feb.
  2021.

\bibitem{Hadidi_RAL18}
R.~Hadidi, J.~Cao, M.~Woodward, M.~S. Ryoo, and H.~Kim, ``Distributed
  perception by collaborative robots,'' \emph{IEEE Robot. Autom. Lett.},
  vol.~3, no.~4, p. 3709–3716, Oct. 2018.

\bibitem{Teerapittayanon_ICDCS17}
S.~Teerapittayanon, B.~McDanel, and H.~Kung, ``Distributed deep neural networks
  over the cloud, the edge and end devices,'' in \emph{{Proc}. {IEEE} Int.
  Conf. Distrib. Comput. Syst. (ICDCS)}, Atlanta, GA, USA, Jun. 2017.

\bibitem{ChoiJinhang_DAC19}
J.~Choi, Z.~Hakimi, P.~W. Shin, J.~Sampson, and V.~Narayanan, ``Context-aware
  convolutional neural network over distributed system in collaborative
  computing,'' in \emph{Proc. ACM/IEEE Design Autom. Conf. (DAC)}, Las Vegas,
  NV, USA, Jun. 2019.

\bibitem{Singhal_DATE20}
M.~Singhal, V.~Raghunathan, and A.~Raghunathan, ``Communication-efficient
  view-pooling for distributed multi-view neural networks,'' in \emph{Proc.
  Design, Autom. Test Eur. Conf. Exhib. {(DATE)}}, Virtual Event, Jul. 2020.

\bibitem{JShao_TWC23}
J.~Shao, Y.~Mao, and J.~Zhang, ``Task-oriented communication for multidevice
  cooperative edge inference,'' \emph{IEEE Trans. Wireless Commun.}, vol.~22,
  no.~1, pp. 73--87, Jan. 2023.

\bibitem{ShaoJ_TWC23b}
J.~Shao, X.~Zhang, and J.~Zhang, ``Task-oriented communication for edge video
  analytics,'' \emph{IEEE Trans. Wireless Commun.}, vol.~34, no.~7, pp.
  2198--2207, 2024.

\bibitem{TangX_IOTJ21}
X.~Tang, X.~Chen, L.~Zeng, S.~Yu, and L.~Chen, ``Joint multi-user {DNN}
  partitioning and computational resource allocation for collaborative edge
  intelligence,'' \emph{{IEEE} Internet Things J.}, vol.~8, no.~12, pp.
  9511--9522, Dec. 2021.

\bibitem{XuZ_TPDS21}
Z.~Xu\emph{ et al.}, ``Energy-aware inference offloading for {DNN}-driven
  applications in mobile edge clouds,'' \emph{{IEEE} Trans. Parallel Distrib.
  Syst.}, vol.~32, no.~4, pp. 799--814, Apr. 2021.

\bibitem{LiuZ_JSAC23}
Z.~Liu, Q.~Lan, and K.~Huang, ``Resource allocation for multiuser edge
  inference with batching and early exiting,'' \emph{IEEE J. Sel. Areas
  Commun.}, vol.~41, no.~4, pp. 1186--1200, Apr. 2023.

\bibitem{XiaoY_TCOM2302}
Y.~Xiao\emph{ et al.}, ``Reinforcement learning based energy-efficient
  collaborative inference for mobile edge computing,'' \emph{IEEE Trans.
  Commun.}, vol.~71, no.~2, pp. 864--876, Feb. 2023.

\bibitem{HaoZ_TMC23}
Z.~Hao, G.~Xu, Y.~Luo, H.~Hu, J.~An, and S.~Mao, ``Multi-agent collaborative
  inference via {DNN} decoupling: {I}ntermediate feature compression and edge
  learning,'' \emph{IEEE Trans. Mob. Comput.}, vol.~22, no.~10, pp. 6041--6055,
  Oct. 2023.

\bibitem{LiuF_JSAC22}
F.~Liu, Y.~Cui, C.~Masouros, J.~Xu, T.~X. Han, Y.~C. Eldar, and S.~Buzzi,
  ``Integrated sensing and communications: Toward dual-functional wireless
  networks for 6{G} and beyond,'' \emph{IEEE J. Sel. Areas Commun.}, vol.~40,
  no.~6, pp. 1728--1767, Jun. 2022.

\bibitem{ShiQ_JSAC22}
Q.~Shi, L.~Liu, S.~Zhang, and S.~Cui, ``Device-free sensing in {OFDM} cellular
  network,'' \emph{IEEE J. Sel. Areas Commun.}, vol.~40, no.~6, pp. 1838--1853,
  Jun. 2022.

\bibitem{XuD_TCOM22}
D.~Xu, X.~Yu, D.~W.~K. Ng, A.~Schmeink, and R.~Schober, ``Robust and secure
  resource allocation for {ISAC} systems: A novel optimization framework for
  variable-length snapshots,'' \emph{IEEE Trans. Commun.}, vol.~70, no.~12, pp.
  8196--8214, Dec. 2022.

\bibitem{LiuA_COMST22}
A.~Liu\emph{ et al.}, ``A survey on fundamental limits of integrated sensing
  and communication,'' \emph{IEEE Commun. Surveys Tuts.}, vol.~24, no.~2, pp.
  994--1034, Second Quart. 2022.

\bibitem{WenD_TWC24}
D.~Wen, P.~Liu, G.~Zhu, Y.~Shi, J.~Xu, Y.~C. Eldar, and S.~Cui, ``Task-oriented
  sensing, computation, and communication integration for multi-device edge
  {AI},'' \emph{IEEE Trans. Wireless Commun.}, vol.~23, no.~3, pp. 2486--2502,
  Mar. 2024.

\bibitem{XiaoZ_JSAC22}
Z.~Xiao and Y.~Zeng, ``Waveform design and performance analysis for full-duplex
  integrated sensing and communication,'' \emph{IEEE J. Sel Areas Commun.},
  vol.~40, no.~6, pp. 1823--1837, Jun. 2022.

\bibitem{hua2022mimo}
H.~Hua, T.~X. Han, and J.~Xu, ``{MIMO} integrated sensing and communication:
  {CRB}-rate tradeoff,'' \emph{IEEE Trans. Wireless Commun.}, vol.~23, no.~4,
  pp. 2839--2854, Apr. 2024.

\bibitem{Nechi_ACMTRTS23}
A.~Nechi, L.~Groth, S.~Mulhem, F.~Merchant, R.~Buchty, and M.~Berekovic,
  ``{FPGA}-based deep learning inference accelerators: {W}here are we
  standing?'' \emph{ACM Trans. Reconfig. Techn. Systm.}, vol.~16, no.~4, p.~60,
  2023.

\bibitem{TuX_SEC23}
X.~Tu, A.~Mallik, D.~Chen, K.~Han, O.~Altintas, H.~Wang, and J.~Xie,
  ``Unveiling energy efficiency in deep learning: {M}easurement, prediction,
  and scoring across edge devices,'' in \emph{Proc. IEEE/ACM Symp. Edge Comput.
  (SEC)}, Wilmington, DE, USA, Dec. 2023.

\bibitem{Flavio_MICRO22}
F.~Ponzina\emph{ et al.}, ``A hardware/software co-design vision for deep
  learning at the edge,'' \emph{IEEE Micro}, vol.~42, no.~6, pp. 7--26,
  Nov.-Dec. 2022.

\bibitem{Hadjer_IJCAI21}
H.~Benmeziane, K.~E. Maghraoui, H.~Ouarnoughi, S.~Niar, M.~Wistuba, and
  N.~Wang, ``Hardware-aware neural architecture search: Survey and taxonomy,''
  in \emph{Proc. 30th Int. Joint Conf. Artif. Intell. (IJCAI)}, Virtual Event,
  Aug. 2021.

\bibitem{JiangW_TCAD20}
W.~Jiang\emph{ et al.}, ``Hardware/software co-exploration of neural
  architectures,'' \emph{IEEE Trans. Comput.-Aided Design Integr. Circuits
  Syst.}, vol.~39, no.~12, pp. 4805--4815, Dec. 2020.

\bibitem{HaoC_MDAT21}
C.~Hao, J.~Dotzel, J.~Xiong, L.~Benini, Z.~Zhang, and D.~Chen, ``Enabling
  design methodologies and future trends for edge {AI}: {S}pecialization and
  codesign,'' \emph{IEEE Design Test}, vol.~38, no.~4, pp. 7--26, Aug. 2021.

\bibitem{Bringmann_CODESISSS21}
O.~Bringmann\emph{ et al.}, ``Automated {HW/SW} co-design for edge {AI}:
  {S}tate, challenges and steps ahead,'' in \emph{Proc. Int. Conf.
  Hardware/Software Codesign Syst. Synth.}, Virtual Event, Oct. 2021.

\bibitem{ShiY_Slides2022}
Y.~Shi, ``Hardware/software co-design of deep learning accelerators,''
  \url{https://www3.nd.edu/~scl/slides/codesign-nas.pdf}.

\bibitem{Fasfous_Thesis2022}
N.~Y.~A. A.-Fasfous, ``Hardware-software co-design of deep neural networks:
  From handcrafted to automated design and deployment,'' \emph{Technische
  Universitat Munchen Dr.-Ing. Dissertation}, 2022.

\bibitem{ChenY_ENG20}
Y.~Chen, Y.~Xie, L.~Song, F.~Chen, and T.~Tang, ``A survey of accelerator
  architectures for deep neural networks,'' \emph{Eng.}, vol.~6, pp. 264--274,
  Jan. 2020.

\bibitem{Kudithipudi_NatElec23}
D.~Kudithipudi\emph{ et al.}, ``Design principles for lifelong learning {AI}
  accelerators,'' \emph{Nat. Electron.}, vol.~6, pp. 807--822, Nov. 2023.

\bibitem{Madhavan_23}
A.~Madhavan, ``{Brain-inspired computing can help us create faster, more
  energy-efficient devices — {I}f we win the race},''
  \url{https://www.nist.gov/blogs/taking-measure/brain-inspired-computing-can-help-us-create-faster-more-energy-efficient},
  Mar. 2023.

\bibitem{Yamazaki_BrainSci22}
K.~Yamazaki, V.-K. V.-Ho, D.~Bulsara, and N.~Le, ``Spiking neural networks and
  their applications: {A} review,'' \emph{Brain Sci.}, vol.~12, no.~7, p. 863,
  Jul. 2022.

\bibitem{JangH_SPM19}
H.~Jang, O.~Simeone, B.~Gardner, and A.~Gruning, ``An introduction to spiking
  neural networks: {P}robabilistic models, learning rules, and applications,''
  \emph{IEEE Signal Process. Mag.}, vol.~36, no.~6, pp. 64--77, Nov. 2019.

\bibitem{Accenture_20}
A.~Kass, ``How neuromorphic computing will help industries drive {AI} at the
  edge,''
  \url{https://medium.com/neuromorphic-computing-and-edge-ai/kass-on-neuromorphic-computing-7bfc5de81d5b},
  Dec. 2020.

\bibitem{Venkatesha_TSP21}
Y.~Venkatesha, Y.~Kim, L.~Tassiulas, and P.~Panda, ``Federated learning with
  spiking neural networks,'' \emph{IEEE Trans. Signal Process.}, vol.~9, pp.
  6183--6194, 2021.

\bibitem{YangH_NatCom22}
H.~Yang\emph{ et al.}, ``Lead federated neuromorphic learning for wireless edge
  artificial intelligence,'' \emph{Nat. Commun.}, vol.~13, p. 4269, 2022.

\bibitem{LiuY_TWC23}
Y.~Liu, Z.~Qin, and G.~Y. Li, ``Energy-efficient distributed spiking neural
  network for wireless edge intelligence,'' \emph{IEEE Trans. Wireless
  Commun.}, to appear.

\bibitem{Eshraghian_PIEEE23}
J.~K. Eshraghian\emph{ et al.}, ``Training spiking neural networks using
  lessons from deep learning,'' \emph{Proc. IEEE}, vol. 111, no.~9, pp.
  1016--1054, Sep. 2023.

\bibitem{BOUVIER_JETCS19}
M.~Bouvier, A.~Valentian, T.~Mesquida, F.~Rummens, M.~Reyboz, E.~Vianello, and
  E.~Beigné, ``Spiking neural networks hardware implementations and
  challenges: A survey,'' \emph{ACM J. Emerg. Techn. Comput. Syst.}, vol.~15,
  no.~2, pp. 10--19, Apr. 2019.

\bibitem{PLAGWITZ_arXiv23}
\BIBentryALTinterwordspacing
P.~Plagwitz, F.~Hannig, J.~Teich, and O.~Keszocze, ``To spike or not to spike?
  {A} quantitative comparison of {SNN} and {CNN} {FPGA} implementations.''
  [Online]. Available: \url{https://arxiv.org/pdf/2306.12742.pdf}
\BIBentrySTDinterwordspacing

\bibitem{Ottati_JETCAS23}
F.~Ottati, C.~Gao, Q.~Chen, G.~Brignone, M.~R. Casu, J.~K. Eshraghian, and
  L.~Lavagno, ``To spike or not to spike: {A} digital hardware perspective on
  deep learning acceleration,'' \emph{IEEE J. Emerg. Sel. Topics Circuits
  Syst.}, vol.~13, no.~4, pp. 1015--1025, Dec. 2023.

\bibitem{YuS_MCAS21}
S.~Yu, H.~Jiang, S.~Huang, X.~Peng, and A.~Lu, ``Compute-in-memory chips for
  deep learning: {R}ecent trends and prospects,'' \emph{IEEE Circuits Syst.
  Mag.}, vol.~21, no.~3, pp. 31--56, Third Quart. 2021.

\bibitem{ChangL_SciCHINA21}
L.~Chang\emph{ et al.}, ``Energy-efficient computing-in-memory architecture for
  {AI} processor: device, circuit, architecture perspective,'' \emph{SCIENCE
  CHINA Inf. Sci.}, vol.~64, pp. 1--15, Jun. 2021.

\bibitem{MYTHIC_2021}
MYTHIC, ``M1076 analog matrix processor - product brief,''
  \url{https://mythic.ai/wp-content/uploads/2022/03/M1076-AMP-Product-Brief-v1.0-1.pdf},
  Jun. 2021.

\bibitem{WanZ_ACMJETCS22}
Z.~Wan, T.~Wang, Y.~Zhou, S.~S. Iyer, and V.~P. Roychowdhury, ``Accuracy and
  resiliency of analog compute-in-memory inference engines,'' \emph{ACM J.
  Emerg. Techn. Comput. Syst.}, vol.~18, no.~2, p. Article 33, Mar. 2022.

\bibitem{KHAN_arXiv24}
\BIBentryALTinterwordspacing
A.~A. Khan, J.~Lima, and H.~Farzaneh, ``The landscape of compute-near-memory
  and compute-in-memory: {A} research and commercial overview.'' [Online].
  Available: \url{https://arxiv.org/pdf/2401.14428.pdf}
\BIBentrySTDinterwordspacing

\bibitem{MaD_COMST20}
D.~Ma, G.~Lan, M.~Hassan, W.~Hu, and S.~K. Das, ``Sensing, computing, and
  communications for energy harvesting {IoTs}: {A} survey,'' \emph{IEEE Commun.
  Surveys Tuts.}, vol.~22, no.~2, pp. 1222--1250, Second Quart. 2020.

\bibitem{LvM_ACCESS22}
M.~Lv and E.~Xu, ``Deep learning on energy harvesting {IoT} devices: Survey and
  future challenges,'' \emph{IEEE Access}, vol.~10, pp. 124\,999--125\,014,
  Nov. 2022.

\bibitem{Gobieski_ASPLOS19}
G.~Gobieski, B.~Lucia, and N.~Beckmann, ``Intelligence beyond the edge:
  {I}nference on intermittent embedded systems,'' in \emph{Proc. 24th Int.
  Conf. Archit. Support Programming Languages Operating Syst. (ASPLOS)},
  Providence, RI, USA, Apr. 2019.

\bibitem{WuY_DAC21}
Y.~Wu, Z.~Wang, Z.~Jia, Y.~Shi, and J.~Hu, ``Intermittent inference with
  nonuniformly compressed multi-exit neural network for energy harvesting
  powered devices,'' in \emph{Proc. 57th ACM/EDAC/IEEE Design Autom. Conf.},
  Virtual Event, Jun. 2020.

\bibitem{ParkG_ISLPED23}
G.~Park, O.~Khan, and E.~Seo, ``Energy-harvesting-aware adaptive inference of
  deep neural networks in embedded systems,'' in \emph{Proc. IEEE/ACM Int.
  Symp. Low Power Electron. Design (ISLPED)}, Vienna, Austria, Aug. 2023.

\bibitem{Basak_ISIT21}
B.~Güler and A.~Yener, ``Energy-harvesting distributed machine learning,'' in
  \emph{Proc. IEEE Int. Symp. Inf. Theory (ISIT)}, Melbourne, VIC, Australia,
  July 2021.

\bibitem{Hamdi_IoTJ22}
R.~Hamdi, M.~Chen, A.~B. S.~M. Qaraqe, and H.~V. Poor, ``Federated learning
  over energy harvesting wireless networks,'' \emph{IEEE Internet Things J.},
  vol.~9, no.~1, pp. 92--103, Jan. 2022.

\bibitem{CaoY_arXiv23}
\BIBentryALTinterwordspacing
Y.~Cao, S.~Li, Y.~Liu, Z.~Yan, Y.~Dai, P.~S. Yu, and L.~Sun, ``A comprehensive
  survey of {AI}-generated content ({AIGC}): {A} history of generative {AI}
  from {GAN} to {ChatGPT}.'' [Online]. Available:
  \url{https://arxiv.org/pdf/2303.04226.pdf}
\BIBentrySTDinterwordspacing

\bibitem{ZhouJ_arXiv24}
J.~Zhou\emph{ et al.}, ``Training and serving system of foundation models: {A}
  comprehensive survey,'' \emph{IEEE Open J. Comput. Soc.}, vol.~5, pp.
  107--119, Apr. 2024.

\bibitem{LaiB_arXiv23}
\BIBentryALTinterwordspacing
B.~Lai\emph{ et al.}, ``Resource-efficient generative mobile edge networks in
  {6G} era: {F}undamentals, framework and case study.'' [Online]. Available:
  \url{https://arxiv.org/pdf/2312.12063.pdf}
\BIBentrySTDinterwordspacing

\bibitem{Woisetschlager_arXiv23}
H.~Woisetschläger, A.~Isenko, S.~Wang, R.~Mayer, and H.-A. Jacobsen,
  ``Federated fine-tuning of {LLMs} on the very edge: {T}he good, the bad, the
  ugly,'' in \emph{Proc. IEEE 8th Wkshops. Data Manag. End-to-End Machine
  Learn. (DEEM)}, Santiago, Chile, Jun. 2024.

\bibitem{Touvron_LLaMA23}
\BIBentryALTinterwordspacing
H.~Touvron\emph{ et al.}, ``{LLaMA}: {O}pen and efficient foundation language
  models.'' [Online]. Available: \url{https://arxiv.org/pdf/2302.13971.pdf}
\BIBentrySTDinterwordspacing

\bibitem{XuM_arXiv23}
\BIBentryALTinterwordspacing
M.~Xu, D.~Cai, Y.~Wu, X.~Li, and S.~Wang, ``{FwdLLM}: {E}fficient {FedLLM}
  using forward gradient.'' [Online]. Available:
  \url{https://arxiv.org/pdf/2308.13894.pdf}
\BIBentrySTDinterwordspacing

\bibitem{LinZ_arXiv23}
\BIBentryALTinterwordspacing
Z.~Lin, G.~Qu, Q.~Chen, X.~Chen, Z.~Chen, and K.~Huang, ``Pushing large
  language models to the {6G} edge: {V}ision, challenges, and opportunities.''
  [Online]. Available: \url{https://arxiv.org/pdf/2309.16739.pdf}
\BIBentrySTDinterwordspacing

\bibitem{ZhuX_arXiv23}
\BIBentryALTinterwordspacing
X.~Zhu, J.~Li, Y.~Liu, C.~Ma, and W.~Wang, ``A survey on model compression for
  large language models.'' [Online]. Available:
  \url{https://arxiv.org/pdf/2308.07633.pdf}
\BIBentrySTDinterwordspacing

\bibitem{DuH_IEEENetw23}
H.~Du\emph{ et al.}, ``Exploring collaborative distributed diffusion-based
  {AI}-generated content ({AIGC}) in wireless networks,'' \emph{IEEE Netw.},
  vol.~38, no.~3, pp. 178--186, May 2024.

\bibitem{ZhuRJ_arXiv23}
\BIBentryALTinterwordspacing
R.~Zhu, Q.~Zhao, G.~Li, and J.~K. Eshraghian, ``{SpikeGPT}: {G}enerative
  pre-trained language model with spiking neural networks.'' [Online].
  Available: \url{https://arxiv.org/pdf/2302.13939.pdf}
\BIBentrySTDinterwordspacing

\end{thebibliography}

\end{document}